%% file: main.tex
\documentclass[runningheads]{llncs} 
\usepackage{eccv}
\usepackage{eccvabbrv} 
\usepackage{graphicx}
\usepackage{booktabs}
\usepackage[accsupp]{axessibility}  %
\usepackage{hyperref}
\usepackage{orcidlink}
\input{utils/preamble}
\input{figs/fig_commands}
\input{figs/addition_fig}

\AddToHook{cmd/appendix/before}{%
    \crefalias{section}{appendix}%
    \crefalias{subsection}{appendix}
}
\begin{document}
\title{Disentangling Generation and Regression in Stochastic Interpolants for Controllable\texorpdfstring{\\}{ }Image Restoration} 
\titlerunning{DiSI: Disentangled Stochastic Interpolant}

\author{
Yi Liu~\inst{1} \and
Jia Ma~\inst{1} \and
Wengen Li~\inst{1} \and
Jihong Guan~\inst{1} \and
Shuigeng Zhou~\inst{2} \and
Yichao Zhang~\inst{1} \\
}

\authorrunning{Yi Liu et al.}

\institute{ 
$^1$ Tongji University\quad
$^2$ Fudan University. \\
\email{\{liuyi61,2432018,lwengen,jhguan,yichaozhang\}@tongji.edu.cn, sgzhou@fudan.edu.cn}
}

\ifthenelse{\boolean{showseqatt}}{
\newcommand{\DiSIteaser}{
    \includegraphics[width=0.95\textwidth]{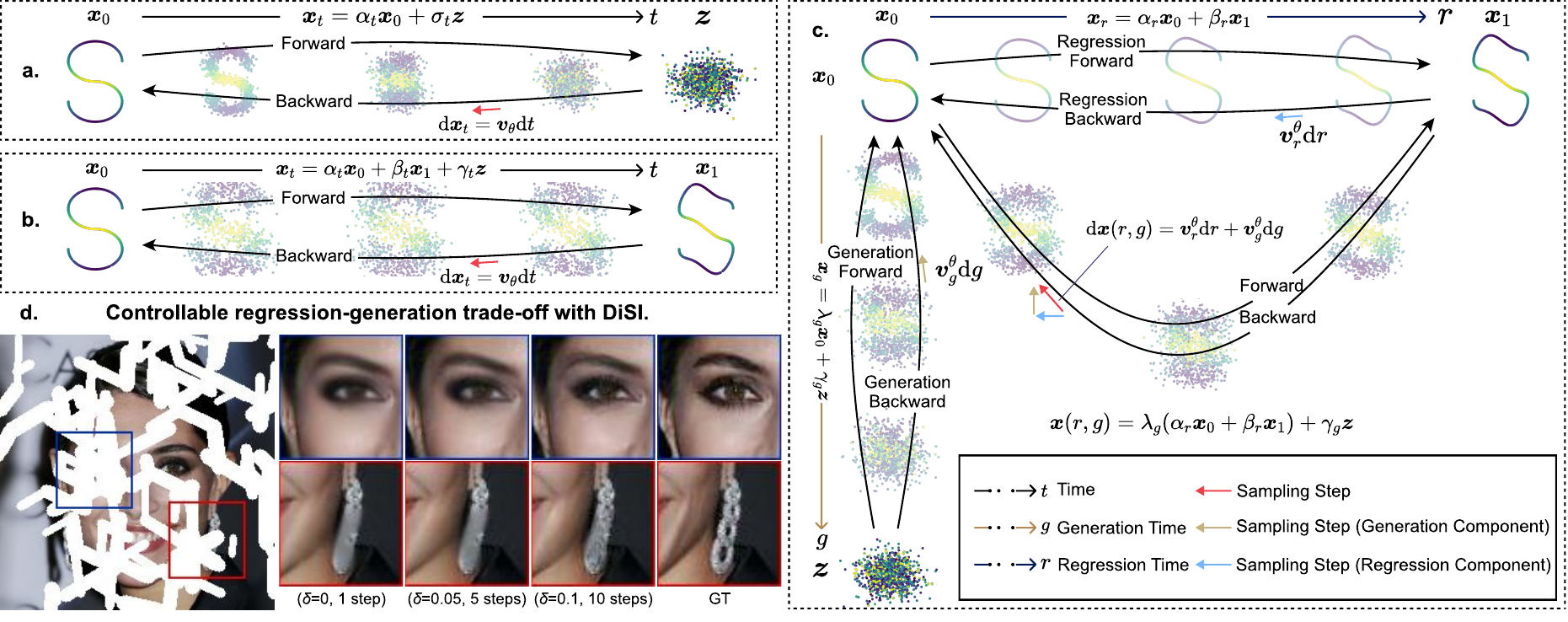}
}
}{
\newcommand{\DiSIteaser}{
    \includegraphics[width=0.95\textwidth]{figs/teaser.pdf}
}
}

\maketitle
\begin{figure}
\centering
    \DiSIteaser
    \caption{
        (a-c) A conceptual comparison between our method \textbf{DiSI} and major existing generative frameworks DMs, FMs and SIs, illustrated by restoring a distorted $\mathsf{S}$ curve.
        (a) DMs/FMs: a path between data $\bfx_0$ and noise $\bfz$.
        (b) SIs: a path between two data points ($\bfx_0$, $\bfx_1$) with intermediate noise. 
        (c) \textbf{DiSI}: a decoupled framework with an independent \textbf{Regression Time} $r$ for the data-to-data path and a \textbf{Generation Time} $g$ for the noise component. 
        Unlike (a-b) learning a single velocity $\bfv_\theta$, \textbf{DiSI} learns two decoupled velocities ($\bfv^\theta_r, \bfv^\theta_g$) for each time axis (see \cref{sec:framework}).
        (d) \textbf{DiSI} enables a smooth and \textbf{controllable} transition from one-step regression to multi-step generation by simply adjusting $\delta$ (see \cref{sec:inference}), using \textbf{the same trained model weights}.
    }
    \label{fig:compare}
\end{figure}
\input{sec/0_abstract}    
\input{sec/1_introduction}
\input{sec/2_related_work}
\input{sec/3_method}
\input{sec/4_experiments}
\input{sec/5_conclusion}

\bibliographystyle{splncs04}
\bibliography{main}
\input{sec/X_suppl}
%
\end{document}

%% file: utils/preamble.tex
\usepackage{setspace}
\usepackage{etoolbox}
\usepackage{multirow}
\usepackage{makecell}
\usepackage{booktabs} 
\usepackage{comment}
\usepackage{colortbl}
\usepackage{units}
\usepackage{ifthen}
\usepackage{float}
\usepackage{array}  
\usepackage{graphicx}  
\usepackage{amsmath} %
\usepackage{tabu}
\usepackage[export]{adjustbox}
\usepackage{arydshln} 
\usepackage{caption}
\usepackage[table]{xcolor} %
\usepackage{bm}
\usepackage{pifont}
\usepackage{wrapfig}
\usepackage{cuted}
\usepackage{tabularx}
\newcolumntype{Y}{>{\centering\arraybackslash}X}

\makeatletter
\newcommand{\thickhline}{%
    \noalign {\ifnum 0=`}\fi \hrule height 1pt
    \futurelet \reserved@a \@xhline
}

\definecolor{codegreen}{rgb}{0.0,0.6,0.0}
\definecolor{textgray}{HTML}{2B2B2B}
\definecolor{blurbox}{HTML}{002FA7}
\usepackage[ruled,vlined,linesnumbered]{algorithm2e}
\makeatletter
\newcommand{\algorithmfootnote}[2][\footnotesize]{%
  \let\old@algocf@finish\@algocf@finish%
  \def\@algocf@finish{\old@algocf@finish%
    \leavevmode\rlap{\begin{minipage}{\linewidth}
    #1#2
    \end{minipage}}%
  }%
}

\newcommand{\name}{DiSI }

\newcommand{\bfx}{\boldsymbol{x}}

\newcommand{\bfz}{\boldsymbol{z}}
\newcommand{\bfv}{\boldsymbol{v}}

\newcommand{\bfk}{\boldsymbol{k}}

\newcommand{\bfQ}{\boldsymbol{Q}}
\newcommand{\bfK}{\boldsymbol{K}}
\newcommand{\bfV}{\boldsymbol{V}}

\newcommand{\alphat}{\alpha_t}
\newcommand{\betat}{\beta_t}
\newcommand{\gammat}{\gamma_t}

\newcommand{\sigmad}{\sigma_{d}}
\newcommand{\matN}{\mathcal{N}}
\newcommand{\mL}{\mathcal{L}}
\newcommand{\sgaussian}{\sim \matN(0, \sigmad^2\bfI)}

\newcommand{\bfI}{\boldsymbol{I}}

\newcommand{\bbR}{\mathbb{R}}
\newcommand{\bbE}{\mathbb{E}}

\newcommand{\rrho}{{1+\rho}}

\newcommand{\rmrho}{{1-\rho}}
\newcommand{\rsqrhos}{\sqrt{1-\rho^2}}

\newcommand{\var}{\text{Var}}

\newcommand{\ud}{\mathrm{d}}

\newcommand{\udt}{\ud t}

\newcommand{\lb}{\left(}
\newcommand{\rb}{\right)}

\newcommand{\lsb}{\left[}
\newcommand{\rsb}{\right]}
\newcommand{\lL}{\left\|}
\newcommand{\rL}{\right\|}
\newcommand{\ldot}{\left.}
\newcommand{\rdot}{\right.}

\newcommand{\bfxrg}{\bfx\lb r,g \rb}
\newcommand{\bfvr}{\bfv_r^\theta}
\newcommand{\bfvg}{\bfv_g^\theta}
\newcommand{\net}{\mathcal{F}_{\theta}}
\newcommand{\netred}{\mathcal{F}_{\textcolor{red}{\theta}}}

\newcommand{\rqtwo}{\cfrac{1}{\sqrt{2}}}
\newcommand{\rqftwo}{\frac{1}{\sqrt{2}}}
\newcommand{\score}{
    \nabla_{\bfx}\log p_t(\bfx)
}
\newcommand{\ar}{
    \rqtwo\lb
        \frac{\cos{r}}{\sqrt{1+\rho}}-
        \frac{\sin{r}}{\sqrt{1-\rho}}
    \rb
}

\newcommand{\br}{
    \rqtwo\lb
        \frac{\cos{r}}{\sqrt{1+\rho}}+
        \frac{\sin{r}}{\sqrt{1-\rho}}
    \rb
}

\newcommand{\dotar}{
    -\rqtwo\lb
        \frac{\sin{r}}{\sqrt{1+\rho}}+
        \frac{\cos{r}}{\sqrt{1-\rho}}
    \rb
}

\newcommand{\dotbr}{
    -\rqtwo\lb
        \frac{\sin{r}}{\sqrt{1+\rho}}-
        \frac{\cos{r}}{\sqrt{1-\rho}}
    \rb
}

\newcommand{\rrange}{
    \arcsin{\sqrt{\frac{1-\rho}{2}}}
}
\newcommand{\hpi}{
    \frac{\pi}{2}
}
\newcommand{\condxrg}{
    | \bfx_{r,g} = \bfx
}

\newcommand{\bcondxrg}{
    \Big| \bfx_{r,g} = \bfx
}

\newcommand{\bbcondxrg}{
    \Bigg| \bfx_{r,g} = \bfx
}
\newcommand{\ptx}{
    p_{r,g}(\bfx)
}
\newcommand{\hptx}{
    \hat p_{r,g}(\bfx)
}
\newcommand{\hptk}{
    \hat p_{r,g}(\bfk)
}
\newcommand{\intr}{
    \int_{\bbR^d}
}
\newcommand{\bfvrf}{
     \bfv\lb\bfx_{r,g}, r\rb 
}
\newcommand{\bfvgf}{
     \bfv\lb\bfx_{r,g}, g\rb 
}
\newcommand{\bftvgf}{
     \tilde{\bfv}\lb\bfx_{r,g}, g\rb 
}
\newcommand{\retas}{
    \sqrt{1-\eta^2}
}

\newcommand{\rfgamma}[1]{
    \cfrac{\dot\gamma_{#1}}{\gamma_{#1}}
}

\newcommand{\linearfunc}[1]{
    \retas\rfgamma{#1}
}

\newcommand{\cov}{
\text{cov}
}
\newcommand{\dg}{
    \dot g(t)
}
\newcommand{\dr}{
    \dot r(t)
}

\newcommand{\dgt}{
    \dot g(\tau)
}
\newcommand{\drt}{
    \dot r(\tau)
}

\newcommand{\ovalr}{
    \varphi\sin t
}
\newcommand{\ovalg}{
    \delta \cos t
}
\newcommand{\linr}{
    2\varphi t - \varphi
}
\newcommand{\ling}{
    \delta t
}

\newcommand{\s}{\hphantom{0}}
\newcommand{\coloneqq}{\mathrel{\mathop:}\mathrel{\mkern-1.2mu}=}

\newcommand{\sinote}{\footnote{
    The general SI form~\cite{albergo2023stochastic} is defined as $\bfx_t = \bfI(t,\bfx_0,\bfx_1) + \gammat\bfz$, subject to $\bfI(0,\bfx_0,\bfx_1)=\bfx_0$ and $\bfI(1,\bfx_0,\bfx_1)=\bfx_1$. 
    In our work, we adopt a linear instance of this formulation.
}}

\newcolumntype{x}[1]{>{\centering\arraybackslash}p{#1pt}}
\newcolumntype{y}[1]{>{\raggedright\arraybackslash}p{#1pt}}
\newcolumntype{z}[1]{>{\raggedleft\arraybackslash}p{#1pt}}
\newcolumntype{I}{!{\vrule width 1pt}}
\definecolor{baselinecolor}{gray}{.9}

\newcommand{\best}[1]{\textbf{#1}}
\newcommand{\secondbest}[1]{\underline{#1}}
\newcommand{\retrain}[1]{}
\newcommand{\na}{N/A}

\newboolean{showseqatt}
\setbool{showseqatt}
{true}
\newboolean{showpred}
\setbool{showpred}{true}

%% file: figs/fig_commands.tex
\newcommand{\network}{
\begin{figure*}[t]
    \centering
    \includegraphics[width=0.99\linewidth]{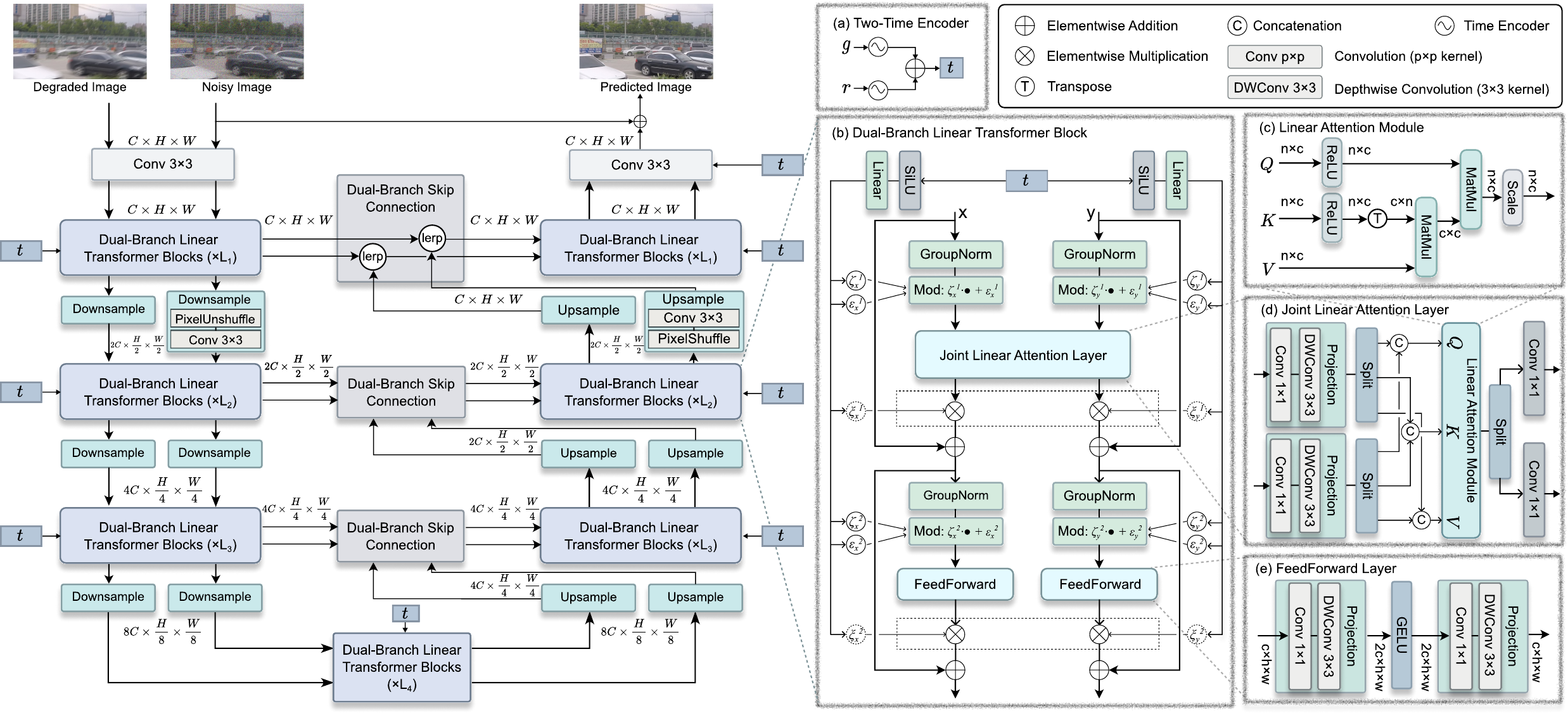}
    \caption{
        \textbf{DULiT architecture.} 
        \textbf{Left}: Backbone (feature resolutions at bottom).
        \textbf{Right}: Modules including (a) Two-Time Encoder and (b) DULiT Block, which comprises the (c) Linear Attention Module, (d) JLA layer, and (e) FFN.
        Dimensions $b, n, c$ denote batch size, sequence length, and channels. 
        Modules in dashed borders are optional.
    }
   \label{fig:network}
\end{figure*}
}

\newcommand{\trajectory}{
\begin{figure*}[t]
    \centering
    \includegraphics[width=\linewidth]{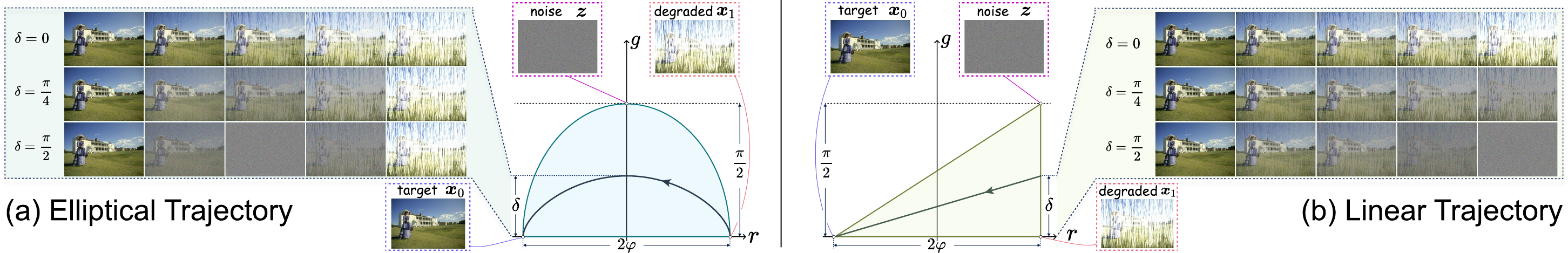}
    \caption{
        Two inference trajectories. %
        (a) The Elliptical Trajectory bridges noiseless $\bfx_1$ and $\bfx_0$ with noise $\bfz$ in intermediate states.
        (b) The Linear Trajectory starts from a noisy mix of $\bfx_1$ and $\bfz$, and ends at noiseless $\bfx_0$. 
        For both, $\delta$ controls the noise level. 
    }
    
   \label{fig:trajectory}
\end{figure*}
}

\newcommand{\vpath}{
\begin{wrapfigure}{R}{0.4\textwidth}
    \centering
    \vspace{-4mm} %
    \includegraphics[width=\linewidth]{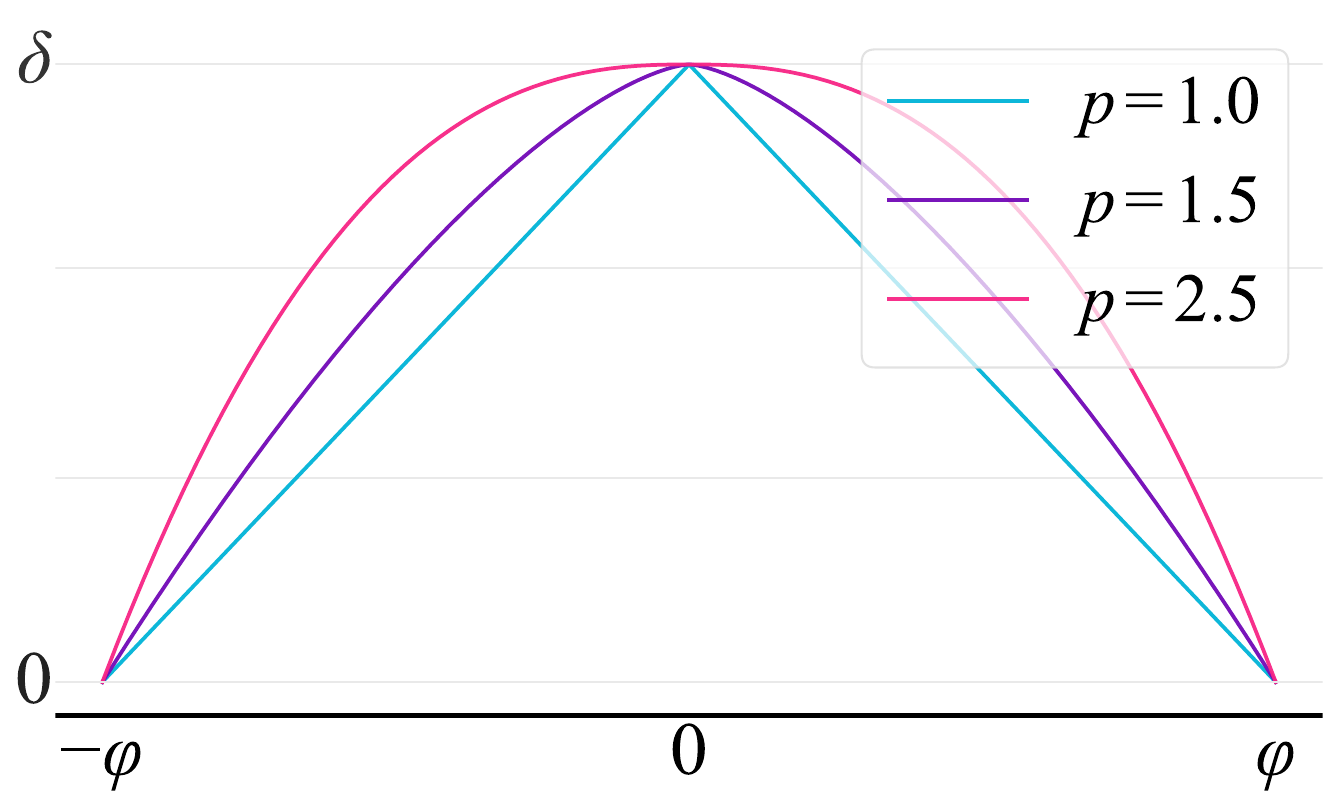}
    \caption{Illustration of V path.}
    \label{fig:v_path}
    \vspace{-4mm} %
\end{wrapfigure}
}

%% file: figs/addition_fig.tex
\ifthenelse{\boolean{showseqatt}}{
\newcommand{\aderaining}{
\begin{figure}[t]
    \centering
    \includegraphics[width=0.98\linewidth]{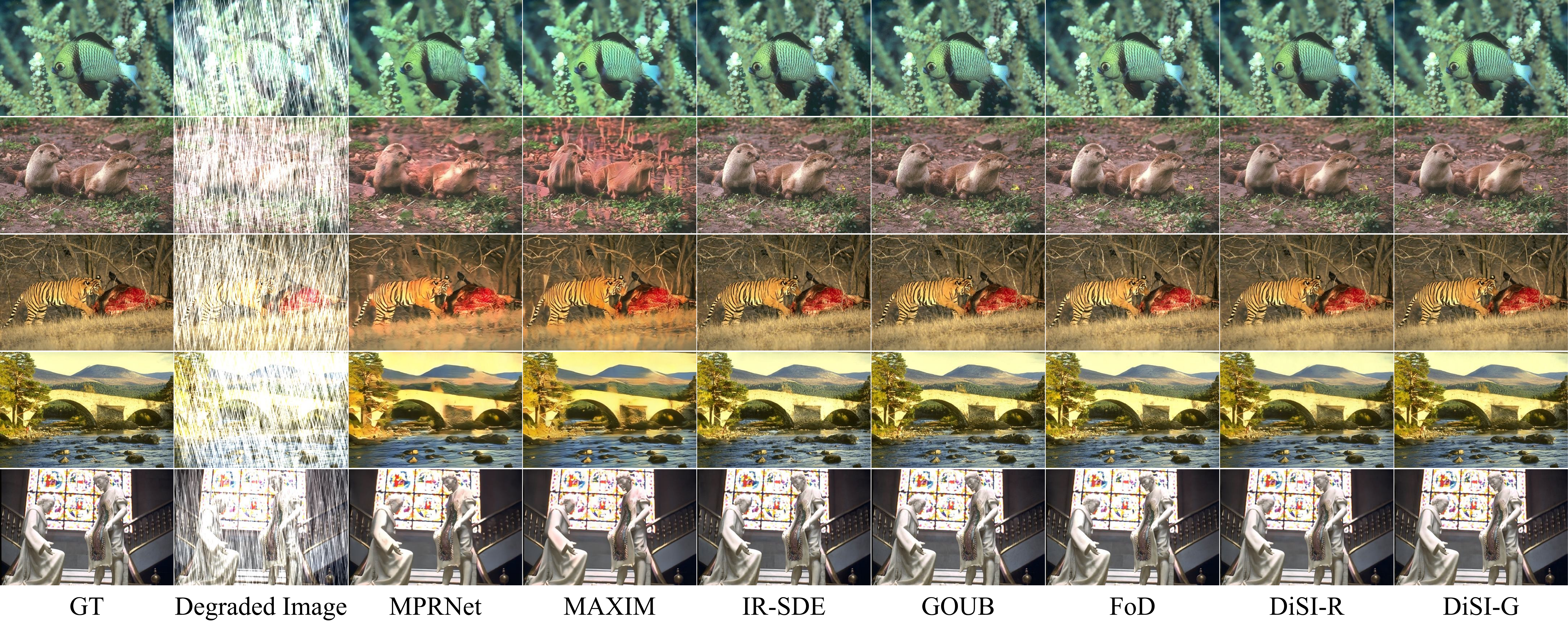}
    \caption{
        Visual comparisons for image deraining on the Rain100H benchmark.
        \label{fig:additional_deraining}
    }
\end{figure}
}
}{
\newcommand{\aderaining}{
\begin{figure}[t]
    \centering
    \includegraphics[width=0.98\linewidth]{figs/addition_deraining.pdf}
    \caption{
        Visual comparisons for image deraining on the Rain100H benchmark.
        \label{fig:additional_deraining}
    }
\end{figure}
}
}

\ifthenelse{\boolean{showseqatt}}{
\newcommand{\alowlight}{
\begin{figure}[t]
    \centering
    \includegraphics[width=0.98\linewidth]{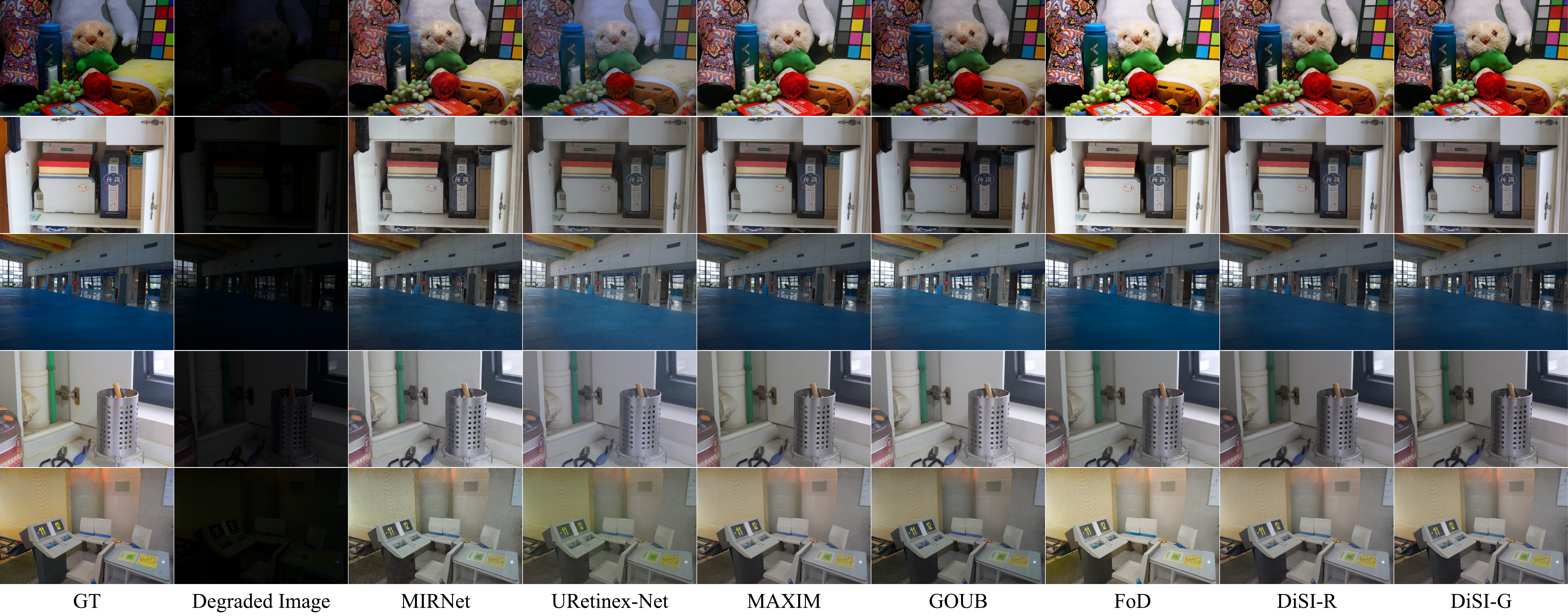}
    \caption{
        Visual comparisons for low-light enhancement on the LOL benchmark.
        \label{fig:additional_lowlight}
    }
\end{figure}
}
}{
\newcommand{\alowlight}{
\begin{figure}[t]
    \centering
    \includegraphics[width=0.98\linewidth]{figs/addition_lowlight.pdf}
    \caption{
        Visual comparisons for low-light enhancement on the LOL benchmark.
        \label{fig:additional_lowlight}
    }
\end{figure}
}
}

\ifthenelse{\boolean{showseqatt}}{
\newcommand{\ainpainting}{
\begin{figure}[t]
    \centering
    \includegraphics[width=0.98\linewidth]{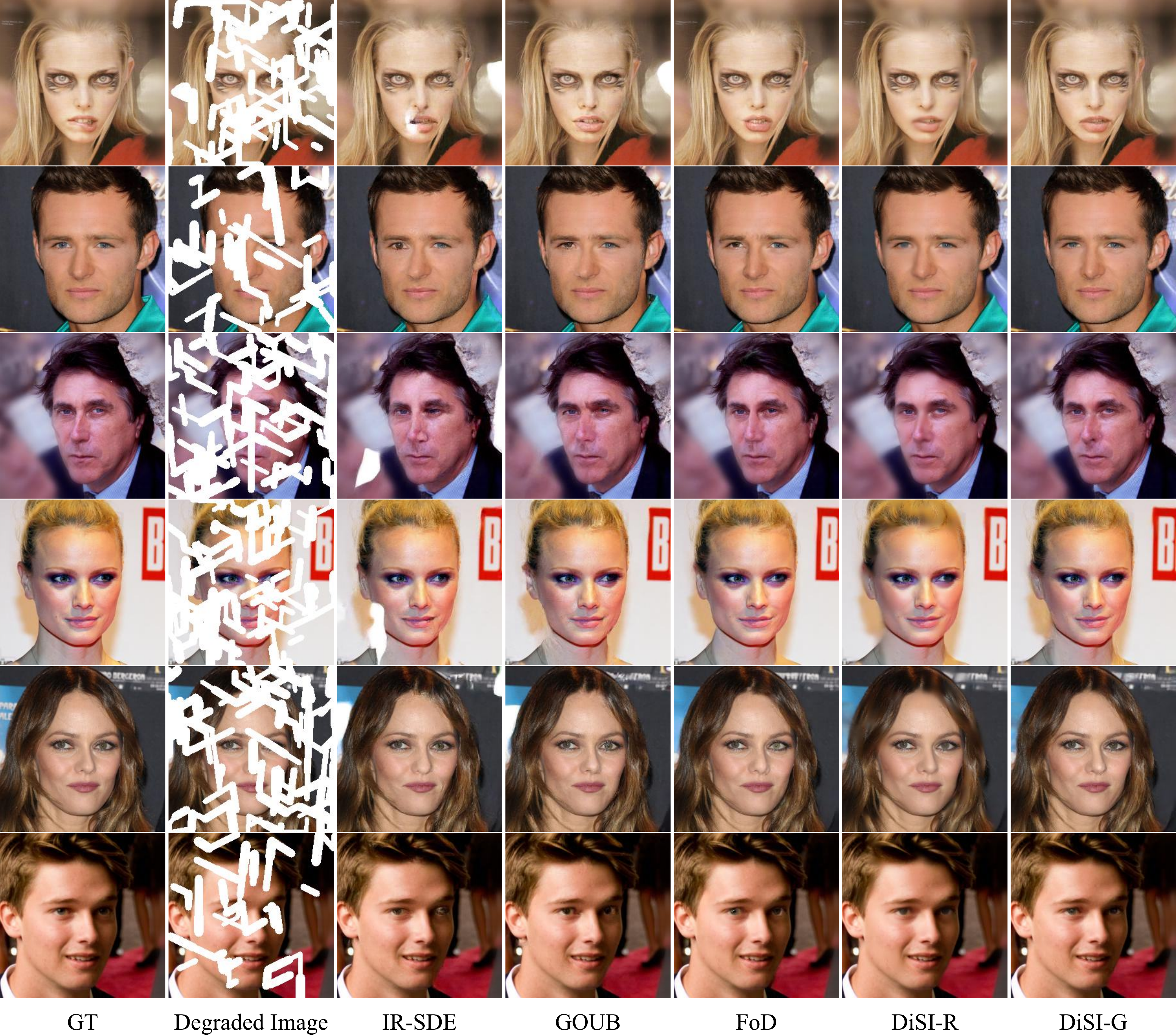} 
    \caption{
        Visual comparisons for image inpainting on the Celeba-HQ benchmark.
        \label{fig:additional_inpainting}
    }
\end{figure}
}
}{
\newcommand{\ainpainting}{
\begin{figure}[t]
    \centering
    \includegraphics[width=0.98\linewidth]{figs/additional_inpainting.pdf}
    \caption{
        Visual comparisons for image inpainting on the Celeba-HQ benchmark.
        \label{fig:additional_inpainting}
    }
\end{figure}
}
}

\ifthenelse{\boolean{showseqatt}}{
\newcommand{\adeblurring}{
\begin{figure}[t]
    \centering
    \includegraphics[width=0.98\linewidth]{figs/additional_deblur_new.pdf}
    \caption{
        Visual comparisons for image deblurring on the GoPro benchmark.
        \label{fig:additional_deblurring}
    }  
\end{figure}
}
}{
\newcommand{\adeblurring}{
\begin{figure}[t]
    \centering
    \includegraphics[width=0.98\linewidth]{figs/addition_deblurring.pdf}
    \caption{
        Visual comparisons for image deblurring on the GoPro benchmark.
        \label{fig:additional_deblurring}
    }
\end{figure}
}
}

\ifthenelse{\boolean{showseqatt}}{
\newcommand{\adderaining}{
\begin{figure}[t]
    \centering
    \includegraphics[width=0.98\linewidth]{figs/additional_deraining_detail_new.pdf}
    \caption{
        Zoomed-in image deraining results on the Rain100H test set.
        \label{fig:additional_deraining_details}
    }
    
\end{figure}
}
}{
\newcommand{\adderaining}{
\begin{figure}[t]
    \centering
    \includegraphics[width=0.98\linewidth]{figs/addition_deraining_detail.pdf}
    \caption{
        Zoomed-in image deraining results on the Rain100H test set.
        \label{fig:additional_deraining_details}
    }
\end{figure}
}
}

\ifthenelse{\boolean{showseqatt}}{
\newcommand{\addlowlight}{
\begin{figure}[t]
    \centering
    \includegraphics[width=0.98\linewidth]{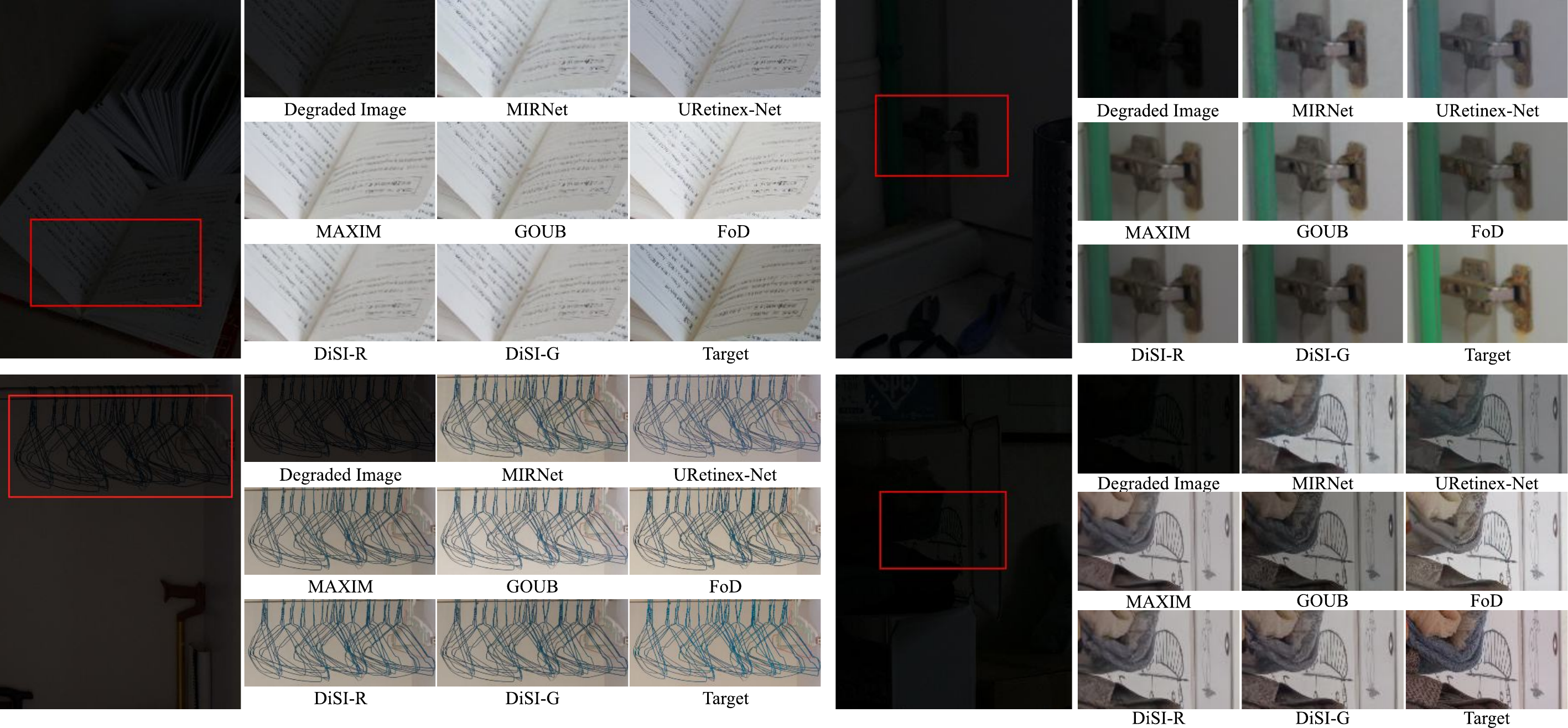}
    \caption{
        Zoomed-in low-light enhancement results on the LOL test set.
        \label{fig:additional_lowlight_details}
    }  
\end{figure}
}
}{
\newcommand{\addlowlight}{
\begin{figure}[t]
    \centering
    \includegraphics[width=0.98\linewidth]{figs/addition_lowlight_detail.pdf}
    \caption{
        Zoomed-in low-light enhancement results on the LOL test set.
        \label{fig:additional_lowlight_details}
    }
    
\end{figure}
}
}

\ifthenelse{\boolean{showseqatt}}{
\newcommand{\addinpainting}{
\begin{figure}[t]
    \centering
    \includegraphics[width=0.98\linewidth]{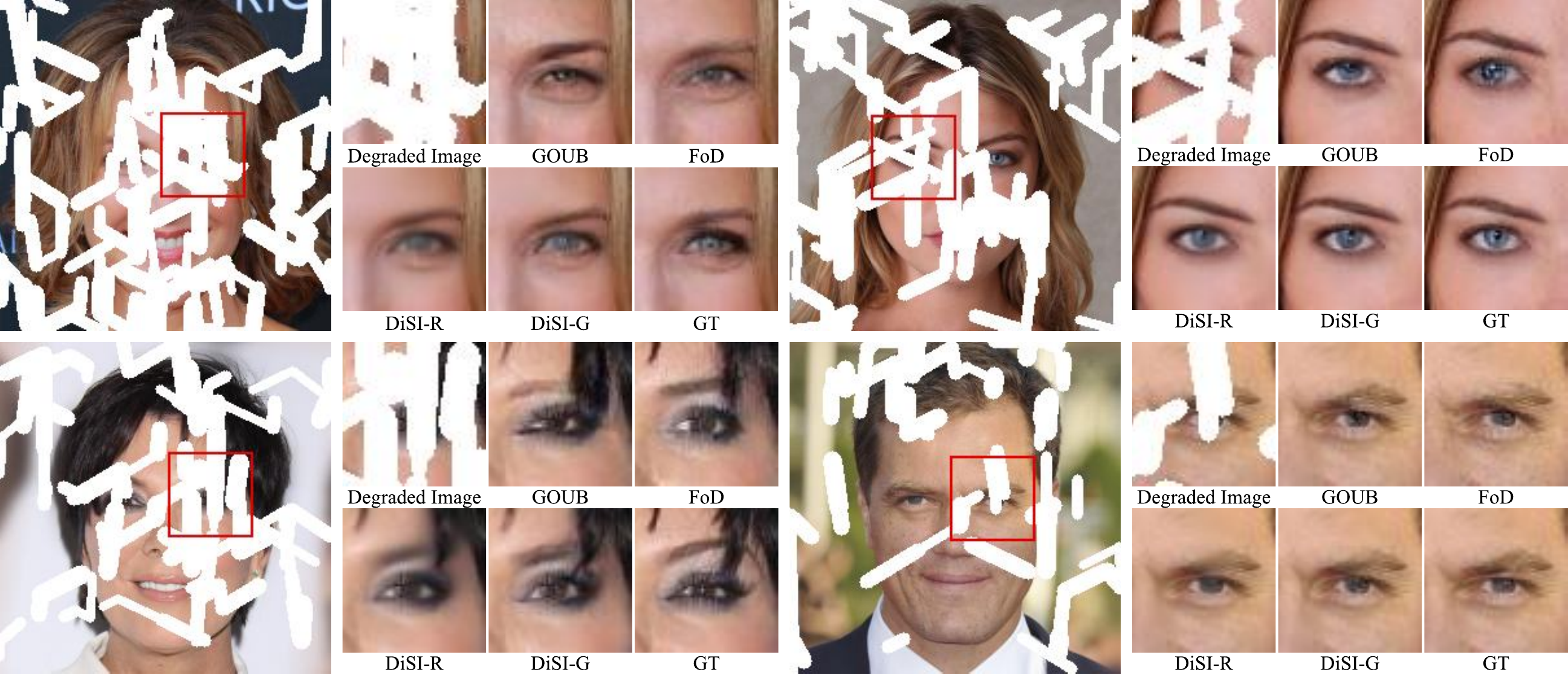}
    \caption{
        Zoomed-in image inpainting results on the CelebA-HQ test set.
        \label{fig:additional_inpainting_details}
    }
\end{figure}
}
}{
\newcommand{\addinpainting}{
\begin{figure}[t]
    \centering
    \includegraphics[width=0.98\linewidth]{figs/addition_inpainting_detail.pdf}
    \caption{
        Zoomed-in image inpainting results on the CelebA-HQ test set.
        \label{fig:additional_inpainting_details}
    }
\end{figure}
}
}

\ifthenelse{\boolean{showseqatt}}{
\newcommand{\addebluring}{
\begin{figure}[t]
    \centering
    \includegraphics[width=0.98\linewidth]{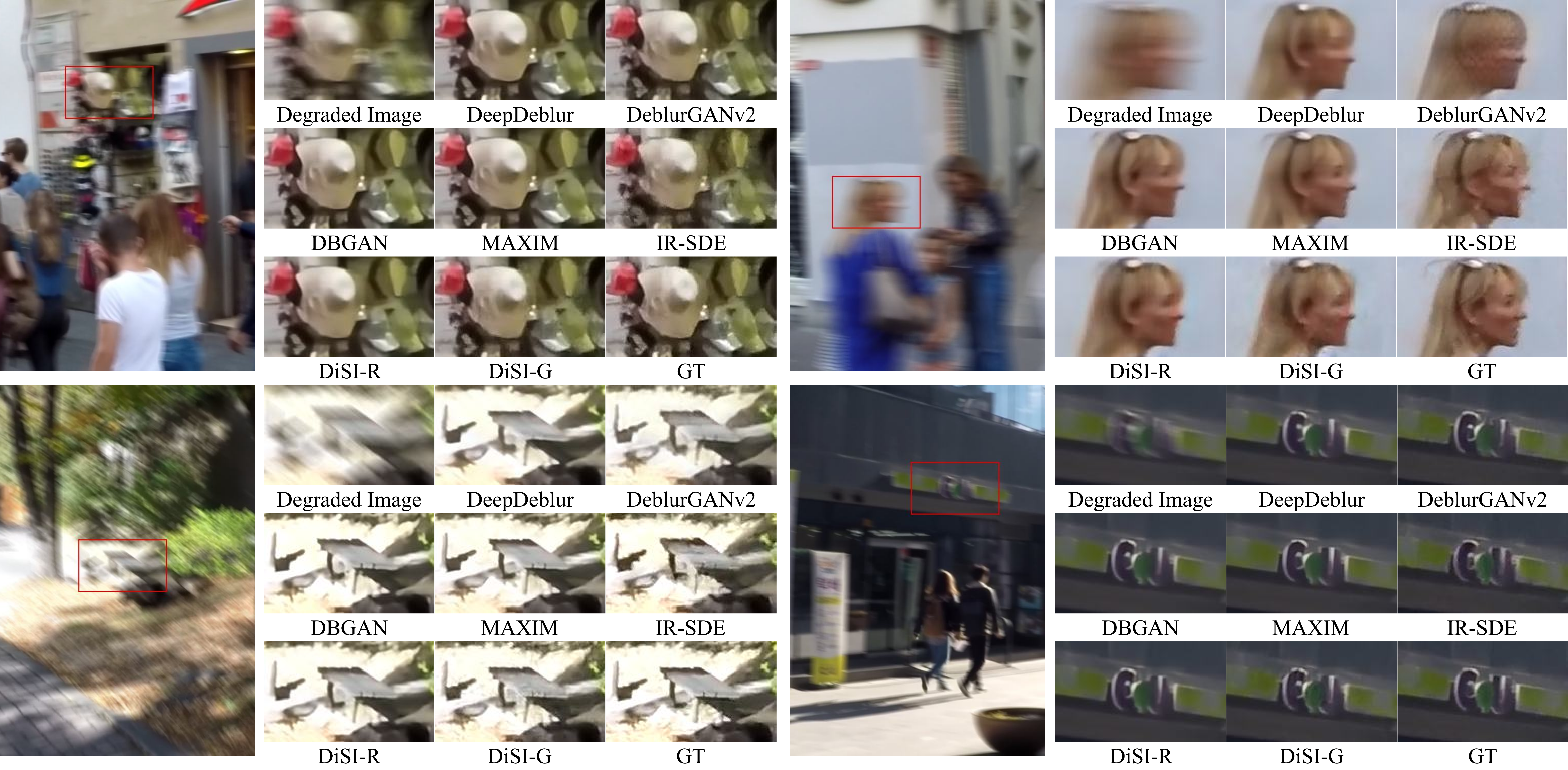}
    \caption{
        Zoomed-in image deblurring results on the GoPro test set.
        \label{fig:additional_deblurring_details}
    }
\end{figure}
}
}{
\newcommand{\addebluring}{
\begin{figure}[t]
    \centering
    \includegraphics[width=0.98\linewidth]{figs/addition_deblurring_detail.pdf}
    \caption{
        Zoomed-in image deblurring results on the GoPro test set.
        \label{fig:additional_deblurring_details}
    }
\end{figure}
}
}

%% file: sec/0_abstract.tex
\begin{abstract}
Recent advances in Image Restoration (IR) have been largely driven by generative methods such as Diffusion Models and Flow Matching, which excel in synthesizing realistic textures while suffering from slow multi-step inference and compromised pixel fidelity.
In contrast, classical regression-based IR methods excel precisely in these aspects, offering single-step efficiency and high pixel-level reconstruction fidelity. 
To bridge this gap, we propose \textbf{DiSI}, a unified framework that \textbf{Di}sentangles the underlying \textbf{S}tochastic \textbf{I}nterpolant process into independent generation and regression components.
This decoupling endows DiSI with remarkable versatility, enabling a continuous and controllable transition from a pure regression process to a fully generative one. 
Technically, we instantiate this framework with two specific sampling trajectories, accompanied by a unified sampler for high-quality, few-step inference on arbitrary trajectories. 
Furthermore, we design a dual-branch U-Net style transformer network in pixel space, using a dedicated branch to enhance conditional guidance while ensuring high throughput. 
Extensive experiments demonstrate that DiSI efficiently achieves competitive results on various IR tasks, while uniquely offering the inference-time flexibility to control the distortion-perception trade-off within a single model.
\keywords{Image Restoration \and Stochastic Interpolants \and Diffusion Models \and Disentanglement}
\end{abstract}

%% file: sec/1_introduction.tex
\section{Introduction}
\label{sec:introduction}
Image Restoration (IR) is a fundamental and long-standing problem in low-level vision that involves restoring high-quality images from their low-quality (\ie, degraded) counterparts~\cite{yip2007digital}. 
It encompasses multiple tasks, including image super-resolution~\cite{ledig2017photo,wang2018esrgan,saharia2022image,zhang2018residual,zhang2018image}, denoising~\cite{zhang2017beyond,dabov2007image,chen2016trainable}, and
inpainting~\cite{pathak2016context,li2022mat,lugmayr2022repaint}. 
Solving these problems is essential for enhancing image quality and usability.  %

Advances in deep learning have inspired a paradigm of regression-based IR methods, where networks~\cite{liang2021swinir,zamir2022restormer} are trained to map degraded images to clean ones by minimizing a pixel-wise loss (\eg, MSE)~\cite{su2022survey}. 
Despite their efficiency and high performance on various distortion metrics (\eg, PSNR), they often yield over-smoothed or unrealistic textures, compromising perceptual quality~\cite{blau2018perception}.

%
%
%
%
%

To synthesize realistic textures, IR has embraced generative models, with Diffusion Models (DMs)~\cite{ho2020denoising,song2019generative,song2020score} and Flow Matching (FM)~\cite{liu2023flow,lipman2022flow} now dominating this field. 
DMs and FMs involve a forward process adding noise, and a backward process reconstructing samples from Gaussian noise (see \cref{fig:compare}a), conditioned on degraded images~\cite{chung2023diffusion,saharia2022image}. 
However, this is unnecessary and inefficient for IR, as the degraded image itself offers a better starting point than pure noise.  
This insight inspires recent works, \eg, residual diffusion models~\cite{yue2023resshift,liu2024residual}, Mean-Reverting Diffusion Models (MRDMs)~\cite{luo2023image,luo2023refusion}, and Diffusion Bridges (DBs)~\cite{zhou2024denoising,liu2023i2sb,yue2024image}, to construct a direct degraded-to-clean diffusion path. 
A unifying framework is provided by Stochastic Interpolants (SIs)~\cite{albergo2023stochastic,ma2024sit} that generalize DMs/FMs by defining transitions between arbitrary distributions. 
This naturally supports direct degraded-to-clean paths (see \cref{fig:compare}b), particularly suitable for IR~\cite{albergo2024stochastic}.

%
%
%
%
%
%
%
%
%
%
%
%
Despite their powerful ability to synthesize realistic textures, DM/FM-based generative IR methods introduce a critical trade-off. 
They require expensive multi-step inference and often sacrifice the pixel-level fidelity in exchange for their remarkable perceptual quality~\cite{luo2023image,liu2023i2sb,albergo2024stochastic}. 
In contrast, high-fidelity restoration and single-step efficiency are precisely where the regression paradigm~\cite{zamir2022restormer,liang2021swinir} excels.
This fundamental dichotomy between distortion and perception~\cite{blau2018perception}, separating regressive methods that prioritize fidelity and efficiency from generative methods that achieve perceptual quality at a high cost, motivates the following question: 
\textit{Can we develop a unified framework to seamlessly bridge regression and generation, thereby enabling fine-grained control over both aspects?}

To answer this question, this paper presents \textbf{DiSI}, a new framework that \textbf{Di}sentangles generation and regression within \textbf{S}tochastic \textbf{I}nterpolants (\cref{fig:compare}c).
In standard SIs (\cref{fig:compare}b)~\cite{albergo2024stochastic,albergo2023stochastic}, the generative capability stems from the stochastic noise term $\gammat\bfz$, while the deterministic path between $\bfx_0$ and $\bfx_1$ inherently constitutes a regression.
DiSI decouples these by introducing two independent time variables:  \textbf{Regression Time} $r$ governing the deterministic transition, and \textbf{Generation Time} $g$ controlling noise injection. 
This explicit separation directly bridges regression and generation, enabling fine-grained control over the fundamental distortion-perception trade-off.
As shown in \cref{fig:compare}d, DiSI can seamlessly transition from a one-step, high-fidelity regression that over-smooths textures, to a multi-step generation process capable of synthesizing rich, realistic details.

%
%
%
%
%
%
%
%
%
%

In the DiSI framework, we design two specific sampling trajectories: 
an elliptical trajectory analogous to DBs~\cite{liu2023i2sb,zhou2024denoising,yue2024image} and a linear trajectory inspired by MRDMs~\cite{luo2023image,luo2023refusion}. 
We also introduce a unified efficient first-order sampler, enabling high-quality, few-step restoration on arbitrary trajectories.
To further improve performance and efficiency, we propose a novel dual-branch U-Net style transformer (DULiT) in pixel space, inspired by \cite{esser2024scaling,xie2025sana,zamir2022restormer}.
It strengthens conditional guidance from degraded images via its dual-branch design, while incorporating efficient modules like linear attention~\cite{katharopoulos2020transformers} to ensure high throughput.

Our technical contributions are: 
1) DiSI, a novel framework that decouples regression and generation within SI for controllable distortion-perception balance.
2) Two sampling trajectories accompanied by an efficient and analytically derived sampler for fast, high-quality inference.
3) A dual-branch pixel-space network that strengthens conditional guidance while maintaining high efficiency.

%% file: sec/2_related_work.tex
\section{Related Work}
\label{sec:related_work}
\noindent{\bf Image Restoration.} 
IR is a long-standing and fundamental problem in 
low-level vision, with tasks such as image deraining~\cite{yang2020single}, deblurring~\cite{zhang2022deep}, and super-resolution~\cite{wang2020deep}.
Deep learning has established neural networks as the prevailing IR approach, commonly implemented via convolutional architectures~\cite{zhang2018residual,zhang2018image,cai2016dehazenet,zhang2017beyond} or vision transformers~\cite{zamir2022restormer,qiu2023mb,liang2021swinir,chen2021pre}, and trained with regression losses. %
While these methods achieve strong distortion metrics like PSNR and SSIM~\cite{wang2004image}, they often yield unrealistic outputs in highly ill-posed scenarios like inpainting~\cite{xiang2023deep,lugmayr2022repaint}.

The advancement of IR has been significantly driven by generative models~\cite{goodfellow2014generative,ho2020denoising}, with DMs and FMs now dominating the field~\cite{li2025diffusion,he2025diffusion}. 
Most DM/FM-based IR methods start from pure noise and condition on degraded images~\cite{lugmayr2022repaint,xia2023diffir,wang2024exploiting,saharia2022palette,saharia2022image,choi2021ilvr}. 
Others, more aligned with our work, modify the initial distribution to incorporate degraded inputs, including MRDMs~\cite{luo2023image,luo2023refusion}, Diffusion Bridges (DBs)~\cite{yue2024image,liu2023i2sb,zhou2024denoising,zheng2025diffusion}, and SI models~\cite{albergo2024stochastic}.
While generative IR methods excel at producing realistic textures, they may compromise pixel-level fidelity.

\noindent{\bf Diffusion Models.} %
DMs~\cite{croitoru2023diffusion} are a prominent class of probabilistic 
generative models~\cite{theis2015note} for image synthesis~\cite{dhariwal2021diffusion}, pioneered by  DDPM~\cite{ho2020denoising} and NCSN~\cite{song2019generative}. 
Since their introduction, the theoretical scope of DMs has been significantly extended by \textit{Score Matching}~\cite{song2020score,karras2022elucidating} and, more recently \textit{Flow Matching} (FM)~\cite{lipman2022flow,liu2023flow} with the concept of differential equations. %
The latter have already become foundational to image generation~\cite{esser2024scaling,rombach2022high}. 
A broader perspective is offered by \textit{Stochastic Interpolant} (SI)~\cite{albergo2023stochastic,ma2024sit}, a generalized framework bridging arbitrary distributions that unifies DMs and FMs, providing our theoretical basis.

The rapid evolution of DMs and FMs has catalyzed significant advancements, with innovations in network architecture and sampling efficiency being particularly relevant to this work. 
Architecturally, backbones have transitioned from standard U-Nets~\cite{nichol2021improved,dhariwal2021diffusion} to scalable Transformers~\cite{vaswani2017attention} (\eg, DiT~\cite{peebles2023scalable} and MM-DiT~\cite{esser2024scaling}), with recent research increasingly prioritizing efficiency~\cite{xie2025sana,crowson2024scalable,chen2024pixartalpha}. 
Concurrently, substantial efforts have focused on accelerating the iterative inference process~\cite{frans2025one,song2023consistency,liu2023flow,geng2025mean,lu2022dpm,lu2025simplifying}, where advanced solvers like DDIM~\cite{song2021denoising} and DPM-Solver~\cite{lu2022dpm,lu2025dpm,zheng2023dpm} reduce sampling steps while maintaining generation quality.

%% file: sec/3_method.tex
\section{Method}
\label{sec:method}
\subsection{Preliminary}
\label{sec:preliminary}
As introduced in~\cite{albergo2024stochastic,albergo2023stochastic}, the stochastic interpolant (SI) process is defined as\sinote:
\begin{equation}
    \bfx_t = \alphat\bfx_0 + \betat\bfx_1 + \gammat\bfz
    \qquad
    t \in [0,1],
    \label{eq:stochastic_interpolant}
\end{equation}
where: (1) $\alphat$, $\betat$, and $\gammat$ are all differentiable time functions satisfying: $\alpha_0=\beta_1=1$, $\alpha_1=\beta_0=\gamma_0=\gamma_1=0$, and $\forall t \in (0,1), \alphat^2+\betat^2+\gammat^2>0$. (2) $\bfx_0$ and $\bfx_1$ are clean and degraded images drawn from a joint distribution $p(\bfx_0,\bfx_1)$ with finite second moments. (3) $\bfz\sgaussian$ is a Gaussian noise with $\sigmad^2$ as its variance. 
Notably, FM~\cite{lipman2022flow,liu2023flow} can be viewed as a special one-sided instance of SI~\cite{albergo2023stochastic}. 
The temporal evolution of the marginal distribution $p_t(\bfx)$ from~\cref{eq:stochastic_interpolant} is governed by the probability flow (PF) Ordinary Differential Equation (ODE)~\cite{lipman2022flow,lipman2024flow,albergo2023stochastic}:
\begin{equation}
\ud\bfx_t = \bfv(\bfx_t,t)\udt,
\end{equation}
where the velocity $\bfv(\bfx_t,t)$ is typically parameterized and learned by a network.

\subsection{DiSI: A Unified Framework}
\label{sec:framework}
%
%
%
Our work is based on a key observation: the generative power of DMs, FMs, and SIs~\cite{song2020score,lipman2024flow,albergo2023stochastic} stems from stochastic noise injection (\ie, $\bfz$ in \cref{fig:compare}). 
We adopt the SI framework because it unifies these models and fits IR's data-to-data nature~\cite{albergo2024stochastic}. 
However, standard SIs have a critical limitation: they fuse the deterministic regression path with the stochastic generative process by coupling both to a single time variable $t$. 
Hence, our core insight is that fine-grained control over the regression-generation balance requires disentangling these processes.

Motivated by this insight, we reformulate the standard SI process in \cref{eq:stochastic_interpolant} by introducing two independent time variables (\cref{fig:compare}c): 
1) a Regression Time $r$ governing the data-to-data transition (\ie, $\bfx_1 \to \bfx_0$), and
2) a Generation Time $g$ controlling the injected noise $\bfz$.
This decoupling yields our DiSI process:
\begin{equation}
    \bfx(r,g) = \lambda_g(\alpha_r\bfx_0 + \beta_r\bfx_1) + \gamma_g\bfz,
    \label{eq:disi}
\end{equation}
where $(\alpha_r, \beta_r)$ and $(\lambda_g, \gamma_g)$ are the regression and generation coefficients, governed by $r$ and $g$, respectively. 
This formulation explicitly disentangles the regression path $\alpha_r\bfx_0 + \beta_r\bfx_1$ from the noise injection, overcoming the single-variable constraint of standard SIs and enabling independent control of the two processes.  

Analogous to standard DMs, FMs, and SIs, the marginal distribution of  $\bfx(r,g)$ evolves according to a PF-ODE~\cite{sarkka2019applied}. 
DiSI generalizes this via two time variables, yielding an exact differential equation with two distinct vector fields:
\begin{equation}
    \ud\bfxrg = \bfvr \ud r + \bfvg \ud g,
    \label{eq:disi_pfode}
\end{equation}
where $\bfvr$ and $\bfvg$ are the velocities along the $r$ and $g$ axes (\cref{fig:compare}c). 
We train a network $\net$, parameterized by $\theta$, to estimate these velocities, as detailed later. 

\noindent{\bf Coefficient Schedule.} 
The coefficient schedule, which defines the formulation of coefficients in \cref{eq:disi}, significantly affects performance.
While early methods often adopt designs like linear schedules~\cite{liu2023flow,lipman2022flow}, recent works~\cite{ma2024sit,lu2025simplifying} highlight the benefits of variance-preserving (VP) schedules, which stabilize training by maintaining constant input variance~\cite{karras2022elucidating,huang2023normalization}. 
Adhering to this principle, following \cite{ma2024sit}, we design our coefficients as a generalized VP (GVP) formulation:
\begin{equation}
    \begin{aligned}
       \alpha_r &= \ar , & \beta_r  &= \br \\
       \lambda_g &= \cos g , &  \gamma_g &= \sin g
    \end{aligned},
    \label{eq:disi_coeffcients}
\end{equation}
where the definitions of $\alpha_r$ and $\beta_r$ incorporate $\rho$, the correlation coefficient between $\bfx_0$ and $\bfx_1$. 
This design is crucial for IR where clean and degraded images are inherently correlated ($\rho \neq 0$). 
The inclusion of $\rho$ ensures that \cref{eq:disi_coeffcients} satisfies the GVP property (see \cref{app:proof_of_gvp}). 
We select the minimal ranges $r \in [-\varphi, \varphi]$ and $g \in [0, \hpi]$ that span the full coefficient evolution, where $\varphi=\rrange$, 
yielding the boundaries $\bfx(-\varphi,0)=\bfx_0$, $\bfx(\varphi,0)=\bfx_1$ and $\bfx(\cdot, \hpi) = \bfz$. 

\noindent{\bf Network Parameterization.} 
The parameterization of the network's prediction goal is a key design in DMs. 
While previous approaches predict the noise $\bfz$~\cite{ho2020denoising,nichol2021improved,dhariwal2021diffusion}, target $\bfx_0$~\cite{karras2022elucidating,karras2024analyzing}, score $\score$~\cite{song2020score,song2019generative,zhou2024denoising}, or velocity $\bfv_t$~\cite{lipman2022flow,liu2023flow,lipman2024flow}, these goals are known to be theoretically interchangeable within the SI framework~\cite{albergo2023stochastic,ma2024sit}. 
But the choice can significantly impact empirical performance. 

In DiSI, the $\bfv_r, \bfv_g \leftrightarrow \bfx_0$ conversion is mathematically inherent under any schedule, as $\bfx_0$ is the only intrinsic unknown. 
Thus, we choose to predict $\bfx_0$, since jointly predicting $(\bfv_r, \bfv_g)$ forces the network to implicitly satisfy their geometric consistency, degrading performance (see \cref{subtab:pred}). 
Letting $\dot\alpha_r$ and $\dot\beta_r$ denote the time derivatives of $\alpha_r$ and $\beta_r$, the predicted decoupled velocities (\ie, $\bfvr$ and $\bfvg$) can be analytically derived by substituting the network prediction $\hat\bfx_0$ into the partial derivatives of \cref{eq:disi} (see \cref{app:derivation_of_vel} for the derivation):
%
%
%
%
\begin{align}
    \bfvg = \cot{g} \bfx(r,g) - \csc{g}\lb\alpha_r\hat\bfx_0 +\beta_r\bfx_1 \rb, \quad
    \bfvr = \cos{g} (\dot\alpha_r\hat\bfx_0+\dot\beta_r\bfx_1), \label{eq:rel_net_and_velrg}
\end{align}

\subsection{Training Strategies}
\label{sec:training}
\input{algos/time_sampler}

Our training procedure builds upon typical DMs and FMs, while introducing three key designs: the time embedding strategy, network normalization, and the optimization goal.
The complete training algorithm is provided in \cref{app:training_algorithm}.

\noindent{\bf Joint Time Embedding.} 
Unlike typical DMs/FMs, which use a single time variable, DiSI conditions the network $\net$ on both the regression time $r$ and the generation time $g$, a mechanism we term joint time embedding.
The key design choice then lies in how $(r, g)$ is sampled, for which we explore two strategies:
\begin{itemize}
    \item \textbf{Generalist Time Sampler}: 
    To train a versatile model, we sample $r$ and $g$ independently from their respective domains.
    This allows the network to be trained over the full 2D space of $(r, g)$, thereby providing maximal flexibility and enabling arbitrary inference trajectories (see \cref{sec:inference}) without retraining.
    A typical example is to uniformly sample $r\in[-\varphi,\varphi]$ and $g\in[0,\hpi]$. 
    \item \textbf{Specialist Time Sampler}: 
    We sample $(r, g)$ pairs directly from the predefined Elliptical (\cref{eq:elliptical_traj}) and Linear (\cref{eq:linear_traj}) trajectories in \cref{sec:inference}. 
    As detailed in \cref{alg:elliptical_sampler,alg:linear_sampler}, practical sampling first selects $\delta$ in \cref{eq:elliptical_traj,eq:linear_traj} to determine the trajectory, then draws a point on it. 
    This trades flexibility for higher performance on dedicated paths, as shown in \cref{sec:experiments}.
\end{itemize}

\noindent{\bf Network Normalization.} 
Although the GVP schedule in \cref{eq:disi_coeffcients} maintains $\bfx(r,g)$ with constant variance $\sigmad^2$, we follow prior works~\cite{karras2022elucidating,huang2023normalization} for enhanced stability by normalizing the network input to unit variance and rescaling its output back to the original scale.
Accordingly, the complete data-prediction process is:
\begin{equation} 
\hat\bfx_0 = \sigmad \net\left(\frac{\bfxrg}{\sigmad}, \bfx_1, r, g\right),
\label{eq:net_prediction}
\end{equation}
where we use the joint time embedding and degraded images $\bfx_1$ as conditions. 

\noindent{\bf Training Objective.}
We adopt the MSE training objective with dynamic adaptive weighting~\cite{karras2024analyzing}, which improves empirical performance and eliminates manual hyperparameter tuning with minimal overhead~\cite{lu2025simplifying,karras2024analyzing}. 
The loss function is:
\begin{equation}
    \mL_{\text{DiSI}}(\theta,\phi) \coloneqq 
    \bbE_{\bfxrg,r,g} \lsb 
    e^{w_{\phi}(r,g)}\cdot{\lL
        \hat\bfx_0
        - \bfx_0
    \rL}_2^2 - w_{\phi}(r,g)
    \rsb,
    \label{eq:disi_loss_func}
\end{equation}
where $\hat\bfx_0$ is given by \cref{eq:net_prediction}, and $w_{\phi}(r,g)$ is the adaptive loss weight~\cite{karras2024analyzing} predicted by an MLP $\phi$ from the  time variables $(r, g)$, following EDM2~\cite{karras2024analyzing}.

\trajectory

\subsection{Inference}
\label{sec:inference}
While training decouples $r$ and $g$, inference demands an explicit path $(r(t),g(t))$ bridging degraded and clean distributions. 
Although DiSI supports any path satisfying $\bfx(-\varphi, 0) = \bfx_0$, we prioritize \textbf{geometric smoothness} for stable ODE integration and generation, as validated in \cref{tab:traj_analysis}.
Thus, we propose two specific choices: \textbf{Elliptical} and \textbf{Linear} (\cref{fig:trajectory}), as the simplest smooth curves to model the behaviors of DBs~\cite{zhou2024denoising} and MRDMs~\cite{luo2023image}, respectively.
Finally, we introduce a universal, analytically derived ODE-based sampler for any trajectory.

\noindent{\bf Elliptical Trajectory.}
To model DBs~\cite{liu2023i2sb,zhou2024denoising,yue2024image} connecting noiseless $\bfx_1$ and $\bfx_0$ with pure intermediate noise, we employ an elliptical path (\cref{fig:trajectory} (a)): 
\begin{equation}
    r= \varphi\sin{t}, g = \delta\cos{t}    
    \qquad t\in \lsb-\frac{\pi}{2},\frac{\pi}{2}\rsb,
    \label{eq:elliptical_traj}
\end{equation}
which satisfies the elliptical equation $\frac{r^2}{\varphi^2} + \frac{g^2}{\delta^2} = 1$, with hyperparameter $\delta \in [0, \hpi]$ controlling the peak noise level. 
It traverses from $t=\hpi$ (\ie, $\bfx(\varphi, 0) = \bfx_1$) to $t=-\hpi$ (\ie, $\bfx(-\varphi, 0) = \bfx_0$). 
As shown in \cref{fig:trajectory}~(a), $\delta=0$ reduces to pure regression, while $\delta=\hpi$ creates a full generative loop $\bfx_1 \to \bfz \to \bfx_0$.
Thus, varying $\delta$ smoothly controls the regression-generation balance.  
Sampling involves discretizing $t$ from $\hpi$ to $-\hpi$ and computing $(r, g)$ via \cref{eq:elliptical_traj} for ODE integration.

\noindent{\bf Linear Trajectory.}
Analogous to MRDMs~\cite{luo2023image} flowing from a noisy initial state $\matN(\bfx_1, \sigma^2\bfI)$ to a deterministic endpoint $\bfx_0$, we propose a linear trajectory:
\begin{equation}
    r = 2\varphi t - \varphi, g = \delta t \qquad t\in[0,1],
    \label{eq:linear_traj}
\end{equation}
which satisfies the linear equation $\frac{r}{-\varphi} + \frac{g}{\delta/2} = 1$. 
The path traverses from $t=1$ (\ie, $\bfx(\varphi, \delta) \sim \matN(\cos\delta\bfx_1, \sin^2{\delta}\bfI)$) to $t=0$ (\ie, $\bfx(-\varphi, 0) = \bfx_0$). 
The hyperparameter $\delta \in [0, \hpi]$ controls this initial noise level (\cref{fig:trajectory} (b)), balancing regression and generation.
Specifically, $\delta=0$ yields pure regression, while $\delta=\hpi$ starts from pure noise. 
Inference proceeds by discretizing $t$ from $1$ to $0$.

\noindent{\bf A Unified ODE-Based Sampler.}
A key property of our DiSI framework is that all inference trajectories stem from the same underlying PF-ODE in \cref{eq:disi_pfode}. 
This allows us to develop a sampler that is not specific to any particular path. 
While basic numerical (\eg, Euler's) solvers~\cite{karras2022elucidating} require many steps to minimize discretization errors, we derive a specific few-step solver for DiSI. 
Inspired by DPM-Solver~\cite{lu2022dpm,zheng2023dpm,lu2025dpm}, our sampler analytically solves the linear portion of the PF-ODE in \cref{eq:disi_pfode} and linearly approximates the isolated non-linear term (see \cref{app:derivation_of_sampler}), supporting much larger step sizes and enabling high-quality generation in very few steps.
The update rule from $(r_1, g_1)$ to $(r_2, g_2)$ is:
\begin{align}
    & \begin{aligned}
     \bfx_{r_2, g_2} = k^{\sqrt{1-\eta^2}}\bfx_{r_1, g_1} +\cos{g_2}(\alpha_{r_2} \hat\bfx_0 + \beta_{r_2}\bfx_1)
     \\ \quad 
     - k^{\sqrt{1-\eta^2}}\cos{g_1} (\alpha_{r_1} \hat\bfx_0 + \beta_{r_1}\bfx_1) + 
    \kappa \bfz, 
    \end{aligned}
    \label{eq:disi_ddim_sampler} \\
    & \kappa = \mathbb{I}(\eta \neq 0) \cdot \frac{\eta(\sin{g_2}-k^{\sqrt{1-\eta^2}}\sin{g_1})}{1 - \sqrt{1-\eta^2}}, \label{eq:kappa}
\end{align}
where $\hat\bfx_0$ is the network prediction at time step $(r_1, g_1)$ from \cref{eq:net_prediction}, $k = \frac{\sin{g_2}}{\sin{g_1}}$, $\bfz\sgaussian$, $\mathbb{I}$ is the indicator function, and $\bfx(r, g)$ is abbreviated as $\bfx_{r,g}$. 
Like DDIM~\cite{song2021denoising}, \cref{eq:disi_ddim_sampler} is a hybrid sampler where $\eta$ controls stochasticity ($\eta=0$ is deterministic, $\eta=1$ is fully stochastic). 
Note that the initial step $g_1=0$ for elliptical path causes a singularity in $k$. 
Thus, we take a small stochastic step ($\eta = 1$) from $t=\hpi$ to $t=\hpi-\epsilon$ ($\epsilon=0.001$) to bypass the singular boundary.
See \cref{app:derivation_of_sampler,app:infer_algorithm} for the detailed derivation and sampling algorithm.

Specifically, for the pure regressive path, the sampling step is defined as:
\begin{equation}
    \bfx_{r_2, g_2} = \bfx_{r_1, g_1} + (\alpha_{r_2} - \alpha_{r_1})\hat\bfx_0 + (\beta_{r_2} - \beta_{r_1}) \bfx_1,
\label{eq:regressive_sampler}
\end{equation}
where $g_1=g_2=0$. 
This sampler contains no stochastic noise terms and thus is non-generative. 
Notably, the special case of one-step sampling (\ie, $r_1 = \varphi \to r_2 = -\varphi$) yields $\bfx_{r_2, g_2} = \hat\bfx_0$, corresponding to an end-to-end regression model.

\subsection{Efficient Network}
\label{sec:network}
Our network $\net$ integrates a hierarchical \textbf{D}ual-branch \textbf{U}-Net style backbone with efficient \textbf{Li}near \textbf{T}ransformer blocks, termed \textbf{DULiT} (see \cref{fig:network}). 
It operates in pixel space, unlike the prevailing latent-space paradigm~\cite{rombach2022high}, to avoid the information loss of compression and preserve fine textures crucial for IR~\cite{yuan2024conditional}.

\network

\noindent{\bf A. DULiT Backbone.} 
We adopt a dual-branch structure from MM-DiT~\cite{esser2024scaling} to enhance conditional guidance, processing the noisy state $\bfxrg$ and the degraded image $\bfx_1$ in parallel. 
However, the constant-resolution pipeline of DiT backbones~\cite{peebles2023scalable,esser2024scaling} is computationally expensive.
To address this, we integrate the dual-branch design into a hierarchical U-Net backbone, motivated by \cite{ronneberger2015u,crowson2024scalable,zamir2022restormer}. 

As shown in \cref{fig:network}, the DULiT backbone features a three-stage encoder with $L_1$, $L_2$, $L_3$ DULiT blocks, a middle layer with $L_4$ blocks, and a symmetric decoder.
The encoder downsamples features via PixelUnshuffle~\cite{shi2016real} and $3\times3$ convolution, while the decoder upsamples using $3\times3$ convolution and PixelShuffle~\cite{shi2016real}.
Dual-Branch Skip Connections are applied per branch at each resolution level. 
Inspired by HDiT~\cite{crowson2024scalable}, we use an efficient learnable linear interpolation (lerp) for image fusions. 
Input and output features are processed with $3\times3$ convolutions, augmented by a long-range residual connection that improves performance.
To accelerate convergence, following \cite{karras2022elucidating,karras2024analyzing}, we zero-initialize all DULiT output projections and replace biases with a channel of ones concatenated to the input. 

\noindent{\bf B. DULiT Block.} 
The core computational unit for all stages is our DULiT Block (see \cref{fig:network} (b)).
We use AdaGroupNorm~\cite{dhariwal2021diffusion} for time conditioning, which, unlike AdaLayerNorm used in DiT~\cite{peebles2023scalable}, avoids costly reshape operations. 
Within each block, the Joint Linear Attention (JLA) layer is the only cross-branch interaction module, followed by FeedForward Networks (FFNs) for both branches.

\noindent{\bf C. JLA Layer.}
The Joint Linear Attention (JLA) layer forms the core of our DULiT Block.
To address the prohibitive quadratic complexity of vanilla attention~\cite{vaswani2017attention,peebles2023scalable} in high-resolution pixel space, following Sana~\cite{xie2025sana}, we employ ReLU linear attention~\cite{katharopoulos2020transformers} (\cref{fig:network} (c)) that eliminates the softmax operator and reorders matrix operations for linear computational complexity.
Our key innovation lies in its joint design (\cref{fig:network} (d)):
features from both noisy and conditional branches pass through efficient Projection blocks, which are composed of $1\times1$ Conv and $3\times3$ DWConv layers inspired by~\cite{howard2017mobilenets,zamir2022restormer}, to generate dual-branch $\bfQ$, $\bfK$, and $\bfV$. 
\ifthenelse{\boolean{showseqatt}}{
These tensors are concatenated along the sequence dimension, a strategy empirically superior to channel-wise fusion as validated in \cref{subtab:concat}.
To ensure efficiency, we accelerate the entire JLA layer using \texttt{torch.compile}.
}{
We then concatenate these tensors channel-wise. 
Unlike the sequence-wise fusion in MM-DiT~\cite{esser2024scaling}, this naturally aligns with our identical branch resolutions and the channel-wise computation of our linear attention.
}

\noindent{\bf D. FFN.} 
The DULiT Block concludes with an efficient FFN (\cref{fig:network} (e)). It replaces the FFN in MM-DiT~\cite{esser2024scaling} with a computationally efficient design, using the same Projection block as in the JLA layer, and a GELU activation.

\noindent{\bf E. Time Embedding and Modulation.}
Time variables $(r, g)$ are individually encoded %
and summed into a token $t$ (\cref{fig:network} (a)). 
As in DiT~\cite{peebles2023scalable}, 
$t$ is projected per block to generate scale/shift parameters for both branches, modulating normalized features before JLA and FFN (\cref{fig:network} (b)). 
Notably, our experiments in \cref{subtab:multiplier} show that the post-modulation steps ($\xi_{\cdot}^1$, $\xi_{\cdot}^2$) are unnecessary in DULiT.

\noindent{\bf F. Positional Embeddings.}
Following Sana~\cite{xie2025sana}, we omit positional embeddings (\eg, RoPE~\cite{su2024roformer}), since convolutions inherently capture positions~\cite{islam2020how}.

%% file: algos/time_sampler.tex
\makeatletter
\newcommand{\removelatexerror}{\let\@latex@error\@gobble}
\makeatother

\begin{figure}[t]
\begin{minipage}[t]{0.495\linewidth}
\begingroup
\small
\removelatexerror
\begin{algorithm}[H]
    \SetAlgoLined
    \DontPrintSemicolon
    \SetNoFillComment
    \KwIn{Hyperparameter $\varphi$}
    \KwOut{Sampled time pair $(r, g)$}
    $\delta \gets \text{Uniform}(0, \pi /2)$, 
    $t \gets \text{Uniform}(-\pi/2, \pi/2)$ \;
    $(r, g) \gets (\varphi\sin{t}, \delta\cos{t})$\textcolor{gray}{\text{// \cref{eq:elliptical_traj}}} \;
    \Return: $(r, g)$ \textcolor{gray}{\text{// for training}} \;
    \caption{Specialist time sampler for Elliptical paths.
    \label{alg:elliptical_sampler}
    }
\end{algorithm}
\endgroup
\end{minipage}%
\hfill
\begin{minipage}[t]{0.495\linewidth}
\begingroup
\small
\removelatexerror
\begin{algorithm}[H]
    \SetAlgoLined
    \DontPrintSemicolon
    \SetNoFillComment
    \KwIn{Hyperparameter $\varphi$}
    \KwOut{Sampled time pair $(r, g)$}
    $\delta \gets \text{Uniform}(0, \pi /2)$, 
    $t \gets \text{Uniform}(0, 1)$ \;
    $(r, g) \gets (2\varphi t - \varphi, \delta t)$\textcolor{gray}{\text{// \cref{eq:linear_traj}}} \;
    \Return: $(r, g)$ \textcolor{gray}{\text{// for training}} \;
    \caption{Specialist time sampler for Linear paths.
    \label{alg:linear_sampler}
    }
\end{algorithm}
\endgroup
\end{minipage}
\end{figure}

%% file: sec/4_experiments.tex
\section{Experiments}
\label{sec:experiments}

We evaluate \name on Rain100H~\cite{yang2017deep} for deraining, GoPro~\cite{nah2017deep} for deblurring, LOL~\cite{chen2018retinex} for low-light enhancement, and Celeba-HQ~\cite{karras2018progressive} for inpainting.
We report both distortion (PSNR, SSIM~\cite{wang2004image}) and perceptual (LPIPS~\cite{zhang2018unreasonable}, FID~\cite{heusel2017gans}) metrics to assess fidelity and generative quality. 
For deraining, PSNR and SSIM are computed on the Y channel (YCbCr space), as in \cite{luo2023image,ren2019progressive,zamir2021multi}.
For efficiency, we report parameter counts (Param, in M) and average per-image inference latency (Lat., in ms) on the test set, measured on an RTX 4090.
Implementation details, dataset information, and additional results are moved to \cref{app:implemtation,app:dataset_information,app:addition_results}.

\input{tables/all_compare_results}
\input{figs/all_compare_results}

\subsection{Comparative Experiments}
\label{sec:comparison}
We compare DiSI against SOTA IR approaches, the results are in \cref{tab:deraining,tab:deblurring,tab:lowlight,tab:inpainting} and visual comparisons are in \cref{fig:deraining,fig:deblur,fig:lowlight,fig:inpainting}.
Best and second-best results are \best{highlighted} and \secondbest{underlined}, with $\uparrow$/$\downarrow$ indicating higher/lower is better performance. 
DM/FM-based methods are marked in \colorbox{baselinecolor}{gray}.
We report DiSI using the proposed elliptical sampler in \cref{alg:elliptical_sampler} and its corresponding elliptical path. 
Results with other training schemes and trajectories are in \cref{app:addition_results}.
To demonstrate flexible trade-offs between regression and generation, we present two variants: 
(1) {\bf DiSI-R} for fidelity via noiseless ($\delta=0$, \ie, regression) sampling, and
(2) {\bf DiSI-G} for perceptual quality via noisy ($\delta>0$) sampling.
Both share the same trained weights and only differ in sampling parameters (see \cref{app:implementation_comparison}).

Based on the results, we observe: 
(1) %
DM/FM-based baselines excel in perceptual metrics (\eg, LPIPS, FID) but lag in distortion metrics (\eg, PSNR, SSIM), and their multi-step inference incurs significant computational overhead.
(2) %
DiSI generally achieves competitive or even SOTA performance across the four  tasks, demonstrating strong restoration capability.
(3) %
DiSI-R excels in distortion metrics with high efficiency, while DiSI-G achieves remarkable perceptual quality at increased computational cost.
Visually, DiSI-R often yields over-smoothed results, whereas DiSI-G produces realistic textures (see \cref{fig:inpainting}).
Crucially, this transition \textbf{requires no retraining} and is achieved solely by \textbf{tuning sampling parameters} (\eg, $\delta$ and sampling steps), enabling a controllable distortion-perception performance balance within a single trained model.
(4) %
DiSI outperforms DM/FM-based baselines in efficiency while maintaining competitive generative performance, owing to our DULiT network and few-step sampler.

\subsection{Ablation Studies}
\label{sec:ablation_studies}
All ablation experiments are conducted on the Rain100H dataset.
We additionally report the Number of Function Evaluations (NFE) to assess inference efficiency.
For more detailed settings of ablation experiments, see \cref{app:implementation_ablation}.

\noindent{\bf A. Effect of Framework.}
To validate our framework, we benchmark DiSI against representative degraded-to-clean interpolant methods, including Standard SI~\cite{albergo2024stochastic}, MRDMs (\eg, IR-SDE~\cite{luo2023image}), and DBs (\eg, DDBM~\cite{zhou2024denoising}). 
For fairness, all methods employ the same DULiT backbone and training protocol, differing only in their interpolant definitions and corresponding samplers. 

As shown in~\cref{tab:interpolant}, DiSI significantly outperforms previous baselines, such as IR-SDE and DDBM, across all metrics under the same network capacity.
Notably, Standard SI achieves performance comparable to our generative mode (DiSI-G), which is theoretically expected as SI functions as a {static special case} of DiSI (see \cref{fig:compare}).
However, unlike SI, DiSI offers inference-time controllability: DiSI-R yields the best distortion metrics and efficiency, while DiSI-G optimizes perceptual quality, unifying regression and generation under a single framework.

\input{tables/combine_interpolant_traj}
\input{tables/network_modules}

\input{tables/time_sampler}

\input{tables/sampler}

\noindent{\bf B. Effects of Trajectories.}
Although DiSI supports arbitrary paths, we further examine the smoothness hypothesis from \cref{sec:inference} that path continuity is critical.
To this end, we design a parameterized ``V-path'' family (\ie, $r=\varphi t$ and $g=\delta(1-|t|^p)$ for $t\in[-1,1]$, detailed in \cref{app:implementation_ablation}).
Varying $p \in \{1, 1.5, 2.5\}$ controls the continuity at the inflection point $t=0$ from $C^0$ to $C^1$ and $C^2$.

Results in \cref{tab:traj_analysis} show that performance improves monotonically with the order of continuity. %
This validates our Elliptical and Linear paths: rather than mere heuristics, they are the simplest $C^\infty$ curves modeling Diffusion Bridge and MRDM behaviors, respectively.
Moreover, the competitive Quadratic (Quad.) Bezier ($C^\infty$) results prove DiSI generalizes to arbitrary smooth paths, with the Elliptical path offering the optimal balance of simplicity and performance. 

\noindent{\bf C. Effects of Network Settings.}
Results are in~ \cref{tab:network_modules}. Concretely, 

\textbf{Loss Weighting.} 
\cref{subtab:logvar} confirms that removing the adaptive loss weighting~\cite{karras2024analyzing} from \cref{eq:disi_loss_func} significantly degrades performance, proving its necessity. 

\ifthenelse{\boolean{showpred}}{
    \textbf{Prediction Objective.} 
    As analyzed in \cref{sec:framework}, predicting $(\bfv_r, \bfv_g)$ forces the network to implicitly satisfy their coupled relationship. 
    This degrades model performance, as validated in \cref{subtab:pred}. 
    Hence, we choose to directly predict $\bfx_0$.
    
    \textbf{DULiT Architecture.} 
    As shown in \cref{subtab:long_range_residual}, the long-range residual connection proves crucial to model performance. 
    Beyond this global architecture setting, we further ablate the key components of DULiT blocks as below:
}{
    \textbf{Architecture.} 
    The long-range residual connection proves crucial to model performance (\cref{subtab:long_range_residual}). 
    In contrast, adding RoPE~\cite{su2024roformer} provides no improvement (\cref{subtab:rope}), consistent with Sana~\cite{xie2025sana}, so we omit positional embeddings.

\textbf{DULiT blocks.} 
We ablate the key components of DULiT blocks as below:
}

(1) JLA design: 
We investigate the effectiveness of the attention mechanism and the projection layer within JLA in \cref{subtab:attn_layer} by comparing three variants:
\begin{itemize}
    \item \textit{Linear}: We combine linear attention with linear projection as the baseline.
    \item \textit{Conv}: We combine linear attention with convolutional projection, which, as illustrated in \cref{fig:network} (d), comprises Conv $1\times1$ and DWConv $3\times3$.
    \item \textit{Vanilla}: We replace linear attention with vanilla attention in the deepest middle layer of DULiT network, while retaining convolutional projection.
\end{itemize}
\ifthenelse{\boolean{showseqatt}}{
The \textit{Conv} design surpasses \textit{Linear}. 
Though \textit{Vanilla} attention offers slight gains, we select linear attention with convolutional projection (\ie, \textit{Conv}) for its linear efficiency.
Furthermore, we ablate the concatenation dimension in \cref{subtab:concat}, where sequence-wise fusion empirically outperforms the channel-wise alternative.
}{
\textit{Conv} design surpasses \textit{Linear}. 
Though \textit{Vanilla} attention brings slight gains, we select linear attention with convolutional projection (\ie, \textit{Conv}) for its linear  efficiency. 
}

(2) FFN design: In \cref{subtab:ffn_layer}, we evaluate three FFN architectures: a standard \textit{Linear} layer; \textit{ConvGate} (a GeGLU-like~\cite{shazeer2020glu} variant with convolutional projections, detailed in \cref{app:implementation_ablation}); and our \textit{Conv} FFN depicted in \cref{fig:network} (e). 
We adopt the \textit{Conv} design as it yields superior performance across most metrics.

(3) Time modulation. 
We ablate modulation parameters in \cref{fig:network} (b), removing post-modulation scales $\xi_{\cdot}^{\cdot}$ and pre-modulation shifts $\varepsilon_{\cdot}^{\cdot}$. 
\cref{subtab:multiplier} shows scales $\xi_{\cdot}^{\cdot}$ are unnecessary while $\varepsilon_{\cdot}^{\cdot}$ removal harms FID. 
We thus retain $\zeta_{\cdot}^{\cdot}$ and $\varepsilon_{\cdot}^{\cdot}$.

\noindent{\bf D. Effects of Time Samplers.} 
We evaluate Generalist and Specialist time samplers in \cref{tab:time_sampler}. 
Generalist samplers include Uniform and two Lognorm~\cite{esser2024scaling} variants, with the latter drawing $(r, g)$ from $[0,1]$ and mapping them to $[-\varphi, \varphi]$ and $[0, \pi/2]$. 
Specialist samplers train the network exclusively on their 1D inference paths. 
All samplers are tested via three inference paths shown in \cref{tab:time_sampler}. 

\cref{tab:time_sampler} shows:
(1) Generalist samplers excel in perception metrics, as broad 2D $(r, g)$ sampling enhances denoising and generation. 
The lognorm2 sampler is most effective on both Elliptical and Linear paths.
(2) Specialist samplers optimize distortion when training and inference paths align. 
For instance, the Elliptical sampler attains peak PSNR/SSIM when $\delta=0$ (\ie, Regression path).
(3) The Regression sampler ($g=0$) performs poorly even on the Regression path, indicating that stochasticity ($g>0$) acts as crucial augmentation for training. 
\noindent{\bf E. Effects of Inference Hyperparameters.}
We ablate $\delta$, $\eta$, and NFE of our sampler in \cref{eq:disi_ddim_sampler} on the Elliptical path in \cref{tab:hyperparameter}. 
For the deterministic regression path ($\delta=0$), $\eta$ is not applicable.
For noisy paths ($\delta > 0$): 
(1) $\text{NFE}\geq 2$: an initial stochastic step ($\eta=1$) from $t=\hpi$ to $t=\hpi-\epsilon$ ($\epsilon=0.001$) is taken to avoid singularity in \cref{eq:disi_ddim_sampler}.
(2) $\text{NFE}=1$: only $\eta=1$ is valid to bypass the singularity boundary, leaving other settings inapplicable (\ie, N/A in \cref{tab:hyperparameter}).

%
%
%
%
%
%
%
%
%
%
%
%
%
%
%
%
%

\cref{tab:hyperparameter} reveals:
(1) Tuning $\delta$ and NFE enables a controllable distortion-perception balance. 
Specifically, pure regression ($\delta=0, \text{NFE}=1$) yields the best fidelity, whereas moderate stochasticity ($\delta=\pi/8, \text{NFE}=15$) maximizes perceptual realism.
However, excessive noise degrades all metrics. 
(2) Meaningful stochastic generation requires sufficient steps ($\text{NFE} \ge 5$), analogous to typical DMs~\cite{song2020score,ho2020denoising}. 
With $\text{NFE}\le 2$, outputs invariably collapse to regression results regardless of $\delta$ or $\eta$. 
(3) $\eta=0$ consistently proves optimal across all settings.

%% file: tables/all_compare_results.tex
\begin{table}[t]
\centering
\begin{minipage}[t]{0.48\linewidth}
\input{tables/deraining}

\end{minipage}\hfill%
\begin{minipage}[t]{0.48\linewidth}
\input{tables/lowlight}

\end{minipage}
\vfill
\begin{minipage}[t]{0.48\linewidth}
\input{tables/deblurring}
\end{minipage}\hfill%
\begin{minipage}[t]{0.48\linewidth}
\input{tables/inpainting}

\end{minipage}
\end{table}

%% file: tables/deraining.tex
\ifthenelse{\boolean{showseqatt}}{ %
\newcommand{\DiSIderaining}{
    \hdashline
    \rowcolor{baselinecolor}
    DiSI-R
    & \best{35.14} & \best{0.947} & 0.053 & 18.84 & 32.49 & 
    170.94 %
    \\
    \rowcolor{baselinecolor}
    DiSI-G
    & \secondbest{34.69} & \secondbest{0.939} & \secondbest{0.043} & \secondbest{16.73} &  
    32.49 &  
    1707.26
    \\
}
}{ %
\newcommand{\DiSIderaining}{
    \hdashline
    \rowcolor{baselinecolor}
    DiSI-R
    & \best{34.93} & \best{0.946} & 0.054 & 18.88 & 32.49 & 
    118.62 %
    \\
    \rowcolor{baselinecolor}
    DiSI-G
    & \secondbest{34.44} & \secondbest{0.936} & \secondbest{0.044} & \secondbest{17.22} &  
    32.49 &  
    1188.44 %
    \\
}
}
\centering
\captionsetup{type=table}
\caption{
    Quantitative comparison for image deraining on the Rain100H test set.
    \label{tab:deraining}
}
\resizebox{\linewidth}{!}{
\begin{tabular}{@{} l *{4}{c} rr @{}}
\toprule
\multirow{2}{*}{Method} &  \multicolumn{2}{c}{Distortion} & \multicolumn{2}{c}{Perceptual}& \multicolumn{2}{c}{Efficiency}  \\ 
\cmidrule(lr){2-3} \cmidrule(lr){4-5} \cmidrule(lr){6-7}
&  PSNR$\uparrow$ & SSIM$\uparrow$ & LPIPS$\downarrow$ & FID$\downarrow$ & Param &    Lat.  \\
\midrule
MPRNet~\cite{zamir2021multi} 
& 30.41 & 0.891 & 0.158 & 61.59 & \s3.64 &  45.82 %
\\
M3SNet-32~\cite{gao2023mountain}
& 30.64 & 0.892 & 0.154 & 60.26 & 16.73 &  47.48   %
\\
MAXIM~\cite{tu2022maxim}
& 30.81 & 0.903 & 0.133 & 58.72 & 14.18& 
101.02
\\
MHNet~\cite{gao2025mixed}
& 31.08 & 0.899 & 0.126 & 57.93 & 67.54 & 117.13 %
\\
\rowcolor{baselinecolor}
IR-SDE~\cite{luo2023image}
& 31.65 & 0.904 & 0.047 &18.64 & 137.15 &  8133.30 
\\
\rowcolor{baselinecolor}
GOUB~\cite{yue2024image}
&31.96& 0.903& 0.046& 18.14 & 137.15 
& 14769.06  %
\\
\rowcolor{baselinecolor}
FoD~\cite{luo2025forward}
& 32.56$^{\retrain{32.78}}$ & 0.925$^{\retrain{0.931}}$ & \best{0.038}$^{\retrain{0.033}}$ & \best{14.10}$^{\retrain{13.74}}$ & 73.45
& 4247.66 
\\
\DiSIderaining
\bottomrule
\end{tabular}
}

%% file: tables/lowlight.tex
\ifthenelse{\boolean{showseqatt}}{ %
\newcommand{\DiSIlowlight}{ 
\hdashline
\rowcolor{baselinecolor}
DiSI-R
&{23.66} & \secondbest{0.857} & 0.100 & 43.95 & 32.49 & 
253.16  %
\\
\rowcolor{baselinecolor}
DiSI-G
&23.51 & 0.845 &\best{0.089}&\best{35.71} & 32.49 & 
2528.87
\\
}
}{ %
\newcommand{\DiSIlowlight}{
\hdashline
\rowcolor{baselinecolor}
DiSI-R
&{23.61} & \secondbest{0.862} & 0.101 &44.58 & 32.49 & 176.01 
\\
\rowcolor{baselinecolor}
DiSI-G
&23.53 & 0.846 &\best{0.088}&\best{38.08} & 32.49 & 1761.25
\\
}
}

\centering
\captionsetup{type=table}
\caption{
    Quantitative low-light enhancement results on the LOL test set.
    \label{tab:lowlight}
}

\resizebox{\linewidth}{!}{
\begin{tabular}{lccccrr}
\toprule
\multirow{2}{*}{Method} &  \multicolumn{2}{c}{Distortion} & \multicolumn{2}{c}{Perceptual}& \multicolumn{2}{c}{Efficiency}  \\ 
\cmidrule(lr){2-3} \cmidrule(lr){4-5} \cmidrule(lr){6-7}
&  PSNR$\uparrow$ & SSIM$\uparrow$ & LPIPS$\downarrow$ & FID$\downarrow$ & Param &    Lat. \\
\midrule
EnlightenGAN~\cite{jiang2021enlightengan}
& 17.61$^{\retrain{17.54}}$  & 0.653$^{\retrain{0.652}}$  & 0.372$^{\retrain{0.322}}$  & 94.71$^{\retrain{105.90}}$  &  8.64
& 24.05 \\
MIRNet~\cite{zamir2020learning}
& \secondbest{24.14}$^{\retrain{24.14}}$& 0.830$^{\retrain{0.845}}$& 0.250$^{\retrain{0.131}}$& 69.18$^{\retrain{71.64}}$ & 31.79
& 171.84
\\
URetinex-Net~\cite{wu2022uretinex}
& 19.84$^{\retrain{19.86}}$ &0.824$^{\retrain{0.825}}$&0.237$^{\retrain{0.128}}$ & 52.38$^{\retrain{57.56}}$ & 0.34 & 27.85
\\
MAXIM~\cite{tu2022maxim}
& 23.43$^{\retrain{23.47}}$ &\best{0.863}$^{\retrain{0.863}}$&\secondbest{0.098}$^{\retrain{0.101}}$&48.59$^{\retrain{48.78}}$ & 14.18 
& 190.27 
\\
Retinexformer~\cite{cai2023retinexformer}
& \best{25.16} & 0.845 & 0.131 & 71.20 
& 1.61  & 27.74 
\\
\rowcolor{baselinecolor}
IR-SDE~\cite{luo2023image}
& 20.45& 0.787&0.129&47.28 & 137.15  & 11830.74 \\ 
\rowcolor{baselinecolor}
GOUB~\cite{yue2024image}
& 19.29$^{\retrain{21.08}}$ & 0.775$^{\retrain{0.811}}$ & 0.148$^{\retrain{0.099}}$ & 50.44$^{\retrain{37.60}}$ & 137.15 & 27289.98 
\\
\rowcolor{baselinecolor}
FoD~\cite{luo2025forward}
&21.61$^{\retrain{20.44}}$ & 0.819$^{\retrain{0.790}}$ & 0.105$^{\retrain{0.113}}$ & \secondbest{41.31}$^{\retrain{45.82}}$  & 73.45  & 5391.53 
\\
\DiSIlowlight
\bottomrule
\end{tabular}
}

%% file: tables/deblurring.tex
\ifthenelse{\boolean{showseqatt}}{ %
\newcommand{\DiSIdeblur}{
    \hdashline
    \rowcolor{baselinecolor}
    DiSI-R
    & \secondbest{31.23} & 0.917 & 0.113 & 12.85 & 32.49 & %
    976.93 %
    \\
    \rowcolor{baselinecolor}
    DiSI-G
    & 30.06 & 0.886 & 0.092 & \secondbest{10.21} & 32.49 & 9757.27 %
    \\
}
}{ %
\newcommand{\DiSIdeblur}{
    \hdashline
    \rowcolor{baselinecolor}
    DiSI-R
    & 31.00 &0.913 &0.115 & 12.53 & 32.49 & %
    682.38  %
    \\
    \rowcolor{baselinecolor}
    DiSI-G
    & 29.82 & 0.883 & 0.099 & \secondbest{10.79} & 32.49 & 6824.33 %
    \\
}
}

\centering
\captionsetup{type=table}
\caption{
    Quantitative comparison for image deblurring on the GoPro test set.
    \label{tab:deblurring}
}
\resizebox{\linewidth}{!}{
\begin{tabular}{lccccrr}
\toprule
\multirow{2}{*}{Method} &  \multicolumn{2}{c}{Distortion} & \multicolumn{2}{c}{Perceptual}& \multicolumn{2}{c}{Efficiency}  \\ 
\cmidrule(lr){2-3} \cmidrule(lr){4-5} \cmidrule(lr){6-7}
&  PSNR$\uparrow$ & SSIM$\uparrow$ & LPIPS$\downarrow$ & FID$\downarrow$ & Param &    Lat. \\
\midrule
DeepDeblur~\cite{nah2017deep} &
30.42 & 0.902 & 0.133 & 15.90  &  11.72
& 113.33
\\
DeblurGAN~\cite{kupyn2018deblurgan} &
28.70$^{\retrain{24.89}}$ & 0.858$^{\retrain{0.7736}}$  & 0.178$^{\retrain{0.232}}$  & 27.02$^{\retrain{38.11}}$  & 6.07 &
33.43
\\
DeblurGAN-v2~\cite{kupyn2019deblurgan} &
29.55$^{\retrain{20.80}}$ & \secondbest{0.934}$^{\retrain{0.8368}}$ & 0.117$^{\retrain{0.240}}$ & 13.40$^{\retrain{16.24}}$ & 5.09 &
68.14
\\
DBGAN~\cite{zhang2020deblurring} &
\ifthenelse{\boolean{showseqatt}}{
31.18$^{\retrain{31.18}}$
}{
\secondbest{31.18}$^{\retrain{31.18}}$
}
& 0.916$^{\retrain{0.9164}}$ & 0.112$^{\retrain{0.111}}$ & 12.65$^{\retrain{12.66}}$ & 11.59 & 949.66 %
\\
MAXIM~\cite{tu2022maxim} & 
\best{32.86}$^{\retrain{32.90}}$ & \best{0.940}$^{\retrain{0.9405}}$ & \secondbest{0.089}$^{\retrain{0.088}}$ & 11.57$^{\retrain{11.59}}$ &
 22.21 & 584.68\\
\rowcolor{baselinecolor}
IR-SDE~\cite{luo2023image} &
30.70 & 0.901 & \best{0.064} & \s\best{6.32} & 137.15
&71573.85 \\
\DiSIdeblur
\bottomrule
\end{tabular}
}

%% file: tables/inpainting.tex
\ifthenelse{\boolean{showseqatt}}{ %
\newcommand{\DiSIinpainting}{
    \hdashline
    \rowcolor{baselinecolor}
    DiSI-R
    & \best{33.85} & \best{0.955} & 0.036 & 18.74  & 32.49 & 
    65.85 %
    \\ %
    \rowcolor{baselinecolor}
    DiSI-G
    & \secondbest{32.84} & \secondbest{0.943} & \secondbest{0.024} & \secondbest{12.82} & 32.49 &
    649.31
    \\
}
}{ %
\newcommand{\DiSIinpainting}{
    \hdashline
    \rowcolor{baselinecolor}
    DiSI-R
    & \best{33.85} & \best{0.955} & 0.037 & 19.09  & 32.49 &  46.20 %
    \\ %
    \rowcolor{baselinecolor}
    DiSI-G
    & \secondbest{32.80} & 0.942 & \secondbest{0.027} & \secondbest{14.20} & 32.49 & 449.27 %
    \\
}
}

\centering
\captionsetup{type=table}
\caption{
    Quantitative image inpainting results on the Celeba-HQ test set.
    \label{tab:inpainting}
}

\resizebox{\linewidth}{!}{
\begin{tabular}{lccccrr}
\toprule
\multirow{2}{*}{Method} &  \multicolumn{2}{c}{Distortion} & \multicolumn{2}{c}{Perceptual}& \multicolumn{2}{c}{Efficiency}  \\ 
\cmidrule(lr){2-3} \cmidrule(lr){4-5} \cmidrule(lr){6-7}
&  PSNR$\uparrow$ & SSIM$\uparrow$ & LPIPS$\downarrow$ & FID$\downarrow$ & Param &    Lat. \\
\midrule
PromptIR~\cite{potlapalli2306promptir}
&30.22& 0.918& 0.068& 32.69 & 35.59 & 66.63  \\ %
\rowcolor{baselinecolor}
DDRM~\cite{kawar2022denoising}
&27.16& 0.899& 0.089& 37.02 & 113.67 &  742.24 \\ %
\rowcolor{baselinecolor}
IR-SDE~\cite{luo2023image}
&28.37$^{\retrain{28.35}}$& 0.917$^{\retrain{0.9161}}$& 0.046$^{\retrain{0.042}}$& 25.13$^{\retrain{24.44}}$& 34.80 & 5852.14 \\
\rowcolor{baselinecolor}
GOUB~\cite{yue2024image}
&30.28 & 0.930 & 0.030 & 14.98 & 137.15 & 6731.15 \\
\rowcolor{baselinecolor}
FoD~\cite{luo2025forward} &
32.02& \secondbest{0.943} & \best{0.022} & \best{11.90}  & 73.45 & 1681.86 \\ %
\DiSIinpainting
\bottomrule
\end{tabular}
}

%% file: figs/all_compare_results.tex
\ifthenelse{\boolean{showseqatt}}{
\newcommand{\derainingvis}{
    \includegraphics[width=\linewidth]{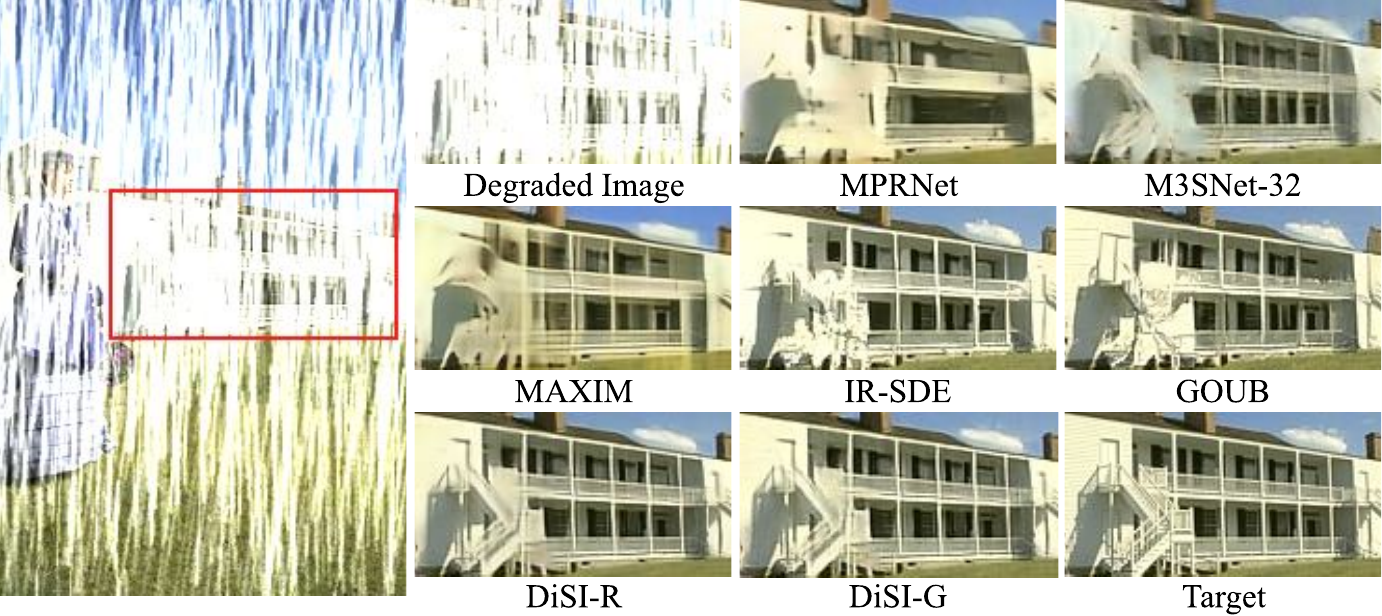}
}
}{
\newcommand{\derainingvis}{
    \includegraphics[width=\linewidth]{figs/deraining.pdf}
}
}
\ifthenelse{\boolean{showseqatt}}{
\newcommand{\lowlightvis}{
    \includegraphics[width=\linewidth]{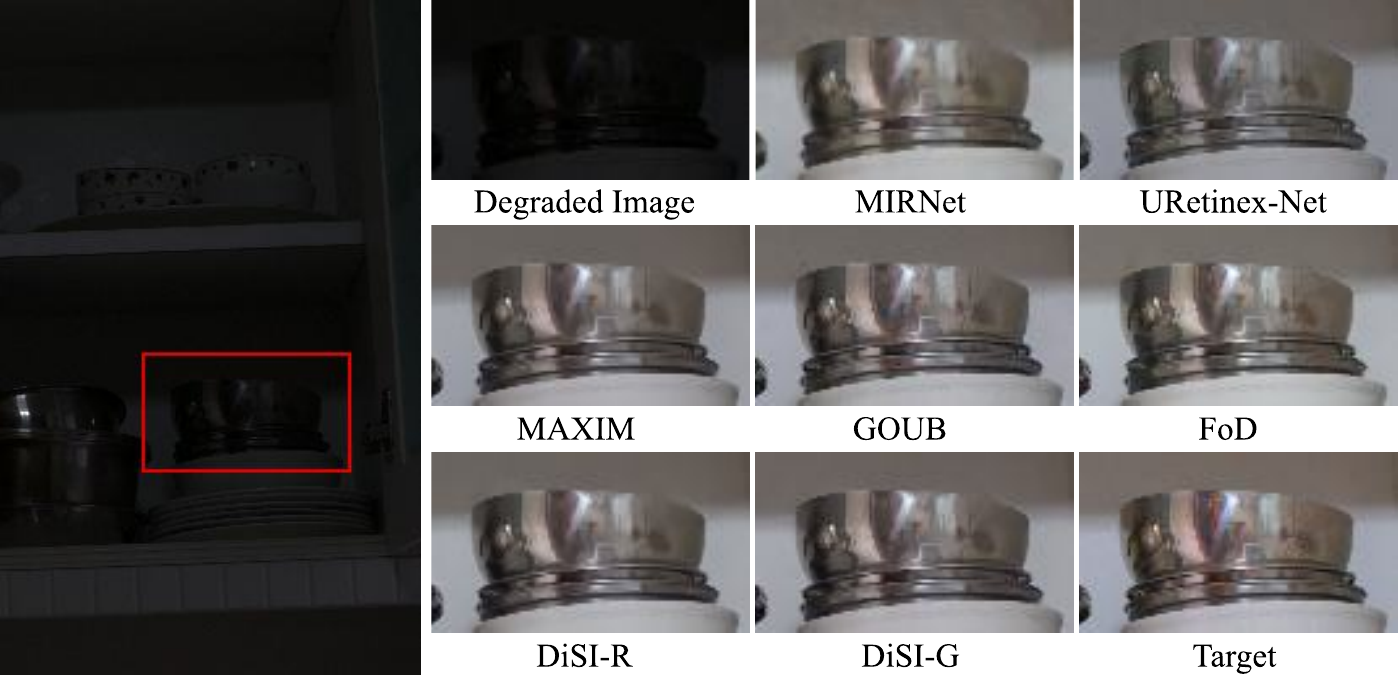}
}
}{
\newcommand{\lowlightvis}{
    \includegraphics[width=\linewidth]{figs/lowlight.pdf}
}
}
\ifthenelse{\boolean{showseqatt}}{
\newcommand{\deblurvis}{
    \includegraphics[width=\linewidth]{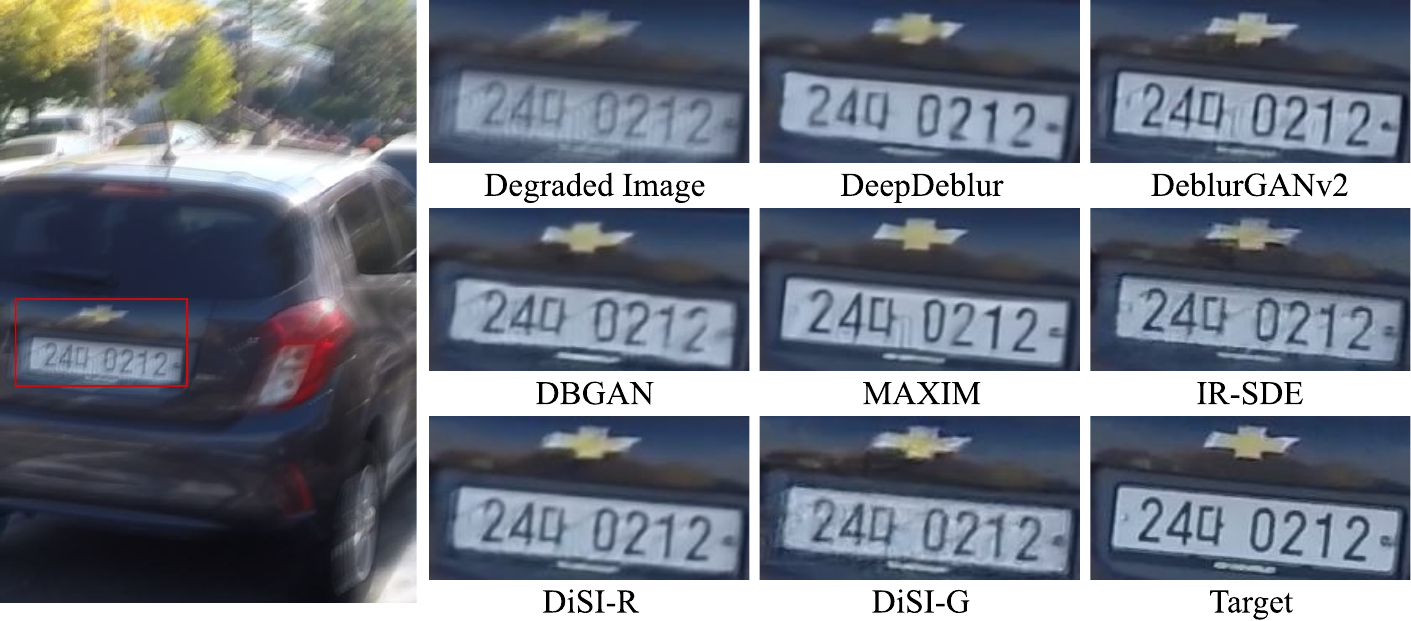}
}
}{
\newcommand{\deblurvis}{
    \includegraphics[width=\linewidth]{figs/deblur.pdf}
}
}
\ifthenelse{\boolean{showseqatt}}{
\newcommand{\inpaintingvis}{
    \includegraphics[width=\linewidth]{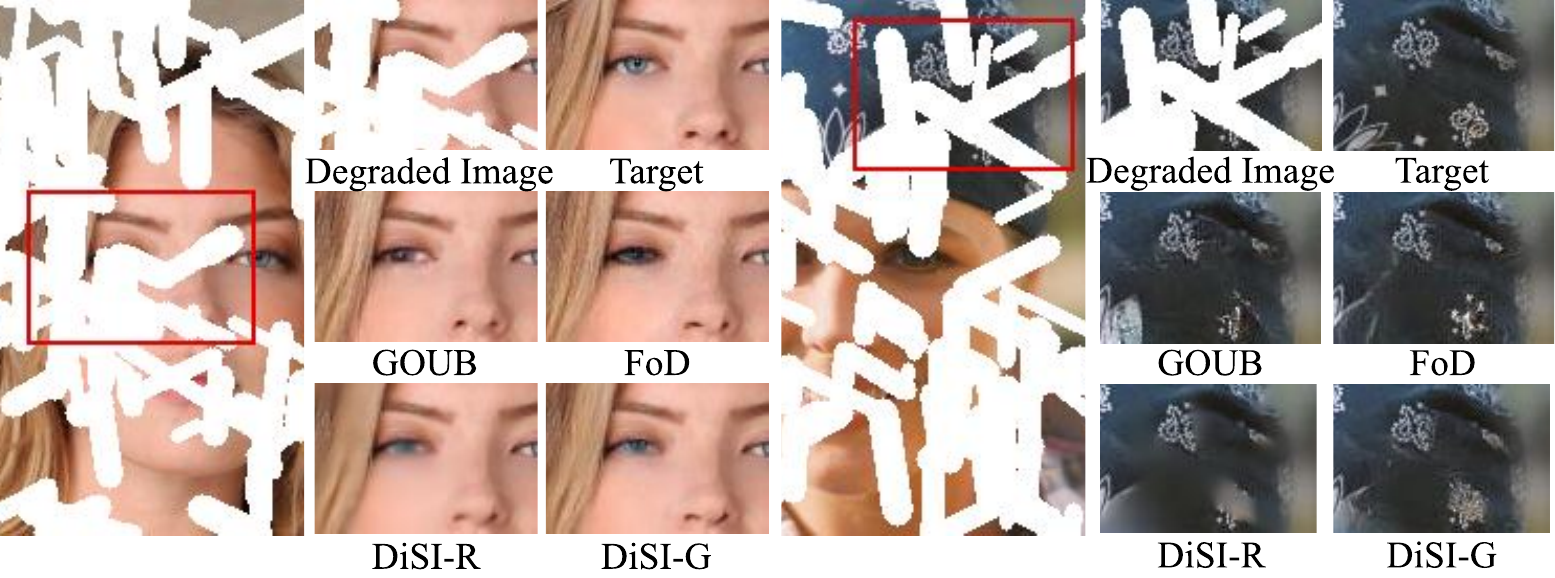}
}
}{
\newcommand{\inpaintingvis}{
    \includegraphics[width=\linewidth]{figs/inpainting.pdf}
}
}

\begin{figure}[t]
\centering
\begin{minipage}[t]{0.51\linewidth}
    \vfill
    \derainingvis
    \vfill
    \captionof{figure}{
        Visual results on Rain100H test set.
        \label{fig:deraining}
    }
\end{minipage}\hfill%
\begin{minipage}[t]{0.48\linewidth}
    \centering
    \vfill
    \lowlightvis
    \vfill
    \caption{
        Visual results on LOL test set.
        \label{fig:lowlight}
    }
\end{minipage}

\begin{minipage}[c]{0.48\linewidth}
    \centering
    \vfill
    \deblurvis
    \vfill
    \captionof{figure}{
        Visual results on GoPro test set.
        \label{fig:deblur}
    }
\end{minipage}\hfill%
\begin{minipage}[c]{0.51\linewidth}
    \centering
    \vfill
    \inpaintingvis
    \vfill
    \caption{
       Visual results on Celeba-HQ test set.
       \label{fig:inpainting}
    }
\end{minipage}
\end{figure}

%% file: tables/combine_interpolant_traj.tex
\begin{table}[t]
    \centering
    \scriptsize 
    \setlength{\tabcolsep}{1.5pt}

    \begin{minipage}[t]{0.49\linewidth}
        \centering
        \captionof{table}{
            \textbf{Framework ablation.}
            Benchmarking DiSI against prior interpolants.
            \label{tab:interpolant}
        }
        \renewcommand{\arraystretch}{1.1} 
        \begin{tabularx}{\linewidth}{@{} l ccccc @{}} %
            \toprule
            Method & PSNR$\uparrow$ & SSIM$\uparrow$ & LPIPS$\downarrow$ & FID$\downarrow$ & NFE \\
            \midrule
            DiSI-R      & \textbf{34.89} & \textbf{0.944} & 0.056 & 19.53 & 1 \\
            DiSI-G      & 34.34 & 0.934 & \textbf{0.047} & 18.62 & 10 \\
            SI & 33.91 & 0.937 & 0.048 & \textbf{17.62} & 10 \\
            IR-SDE    & 20.94 & 0.783 & 0.209 & 89.24 & 100 \\
            DDBM-VP    & 23.54 & 0.876 & 0.139 & 43.55 & 10 \\
            DDBM-VE    & 22.64   & 0.841  & 0.172  & 52.45   & 10 \\
            \bottomrule
        \end{tabularx}
    \end{minipage}
    \hfill %
    \begin{minipage}[t]{0.49\linewidth}
        \centering
        \captionof{table}{
            \textbf{Path Sensitivity.} 
            V $(p)$ denotes the V-path family (defined in \cref{sec:ablation_studies} B) \label{tab:traj_analysis}.
        }
        \renewcommand{\arraystretch}{1.1}
        \begin{tabularx}{\linewidth}{@{} l ccccc @{}}
            \toprule
            Path & PSNR$\uparrow$ & SSIM$\uparrow$ & LPIPS$\downarrow$ & FID$\downarrow$ & Cont. \\
            \midrule
            V ($p$=1)   & 28.95 & 0.911 & 0.061 & 31.36 & $C^0$ \\
            V ($p$=1.5) & 29.57 & 0.918 & 0.056 & 26.41 & $C^1$ \\
            V ($p$=2.5) & 30.40 & 0.926 & 0.053 & 22.37 & $C^2$ \\
            Quad. Bezier     & 30.05 & 0.923 & 0.054 & 23.61 & $C^{\infty}$ \\
            Elliptical      & \textbf{33.61} & \textbf{0.933} & 0.048 & \textbf{18.64} & $C^{\infty}$ \\
            Linear    & 32.94   & 0.917 & \textbf{0.047} & 23.59   &  $C^{\infty}$ \\
            \bottomrule
        \end{tabularx}
    \end{minipage}
\end{table}

%% file: tables/network_modules.tex
\ifthenelse{\boolean{showpred}}{
\newcommand{\netpred}{
    \multicolumn{5}{@{}l}{\refstepcounter{subtable}\textbf{(\thesubtable) Prediction objective \label{subtab:pred}}} \\
    \rowcolor{baselinecolor}
    \quad $\bfx_0$ 
    & \textbf{34.44} & \textbf{0.9364} & \textbf{0.044} & \textbf{17.22} \\
    \quad $(\bfv_r,\bfv_g)$
    & 32.32 & 0.9247 & 0.068 & 28.14 \\
    \midrule
}
}{
\newcommand{\netpred}{}
}

\ifthenelse{\boolean{showpred}}{
\newcommand{\rope}{}
}{
    \multicolumn{5}{@{}l}{\refstepcounter{subtable}\textbf{(\thesubtable) RoPE \cite{su2024roformer}}\label{subtab:rope}} \\
    \rowcolor{baselinecolor}
    \quad \ding{55} & \textbf{34.39} & \textbf{0.9396} & \textbf{0.054} & \textbf{20.43} \\
    \quad \ding{52} & 33.04 & 0.9310 & 0.057 & 24.10 \\
    \midrule
}

\ifthenelse{\boolean{showseqatt}}{ %
\begin{table}[t]
\centering
\caption{Ablation on DULiT modules. Selected configurations are marked in \colorbox{baselinecolor}{gray}.}
\label{tab:network_modules}

\begin{minipage}[t]{0.49\linewidth}
    \centering
    \small 
    \begin{tabularx}{\linewidth}{@{} l *{4}{Y} @{}}
    \toprule
    Config. & PSNR$\uparrow$ & SSIM$\uparrow$ & LPIPS$\downarrow$ & FID$\downarrow$ \\
    \midrule    
    \netpred

    \multicolumn{5}{@{}l}{\refstepcounter{subtable}\textbf{(\thesubtable) Loss weighting}\label{subtab:logvar}} \\
    \quad \ding{55} & 34.38 & 0.9380 & \textbf{0.052} & 19.94 \\
    \rowcolor{baselinecolor}
    \quad \ding{52} & \textbf{34.50} & \textbf{0.9405} & 0.053 & \textbf{19.27} \\
    \midrule
    \multicolumn{5}{@{}l}{\refstepcounter{subtable}\textbf{(\thesubtable) Long-range residual connection}\label{subtab:long_range_residual}} \\
    \quad \ding{55} & 34.39 & 0.9396 & \textbf{0.053} & 19.92 \\
    \rowcolor{baselinecolor}
    \quad \ding{52} & \textbf{34.50} & \textbf{0.9405} & \textbf{0.053} & \textbf{19.27} \\
    \midrule
    \rope
    
    \multicolumn{5}{@{}l}{\refstepcounter{subtable}\textbf{(\thesubtable)  Concatenation} \label{subtab:concat}}\\
    \rowcolor{baselinecolor}
    \quad Sequence & \best{35.00} & \best{0.9456} & \best{0.049} & \best{18.16}   \\
    \quad Channel  & 34.80 & 0.9440 & 0.051 & 18.37  \\
    \bottomrule
    \end{tabularx}
\end{minipage}\hfill%
\begin{minipage}[t]{0.49\linewidth}
    \centering
    \small
    \begin{tabularx}{\linewidth}{@{} l *{4}{Y} @{}}
    \toprule
    Config. & PSNR$\uparrow$ & SSIM$\uparrow$ & LPIPS$\downarrow$ & FID$\downarrow$ \\
    \midrule

    \multicolumn{5}{@{}l}{\refstepcounter{subtable}\textbf{(\thesubtable) The design of JLA layer}\label{subtab:attn_layer}} \\
    \quad Linear & 34.20 & 0.9383 & 0.056 & 20.77 \\
    \rowcolor{baselinecolor}
    \quad Conv & 34.39 & 0.9397 & 0.054 & 20.59 \\
    \quad Vanilla & \textbf{34.42} & \textbf{0.9400} & \textbf{0.053} & \textbf{20.20} \\
    \midrule

    \multicolumn{5}{@{}l}{\refstepcounter{subtable}\textbf{(\thesubtable) The design of FFN layer}\label{subtab:ffn_layer}} \\
    \quad Linear & \textbf{34.39} & \textbf{0.9397} & \textbf{0.054} & 20.59 \\
    \rowcolor{baselinecolor}
    \quad Conv & \textbf{34.39} & 0.9396 & \textbf{0.054} & \textbf{20.43} \\
    \quad ConvGate & 33.94 & 0.9351 & 0.058 & 21.83 \\
    \midrule

    \multicolumn{5}{@{}l}{\refstepcounter{subtable}\textbf{(\thesubtable) Time modulation}\label{subtab:multiplier}} \\
    \quad $(\zeta_{\cdot}^{\cdot})$ & \textbf{34.40} & \textbf{0.9399} & \textbf{0.054} & 21.04 \\
    \rowcolor{baselinecolor}
    \quad $(\zeta_{\cdot}^{\cdot},\varepsilon_{\cdot}^{\cdot})$ & 34.39 & 0.9397 & \textbf{0.054} & \textbf{20.59} \\
    \quad $(\zeta_{\cdot}^{\cdot},\varepsilon_{\cdot}^{\cdot},\xi_{\cdot}^{\cdot})$ & 34.37 & 0.9386 & 0.058 & 20.85 \\
    \bottomrule
    \end{tabularx}
\end{minipage}

\end{table}

}{ %

\begin{table}[t]
\centering
\caption{Ablation on DULiT modules. Selected configurations are marked in \colorbox{baselinecolor}{gray}.}
\label{tab:network_modules}

\begin{minipage}[t]{0.49\linewidth}
    \centering
    \small 
    \begin{tabularx}{\linewidth}{@{} l *{4}{Y} @{}}
    \toprule
    Config. & PSNR$\uparrow$ & SSIM$\uparrow$ & LPIPS$\downarrow$ & FID$\downarrow$ \\
    \midrule

    \multicolumn{5}{@{}l}{\refstepcounter{subtable}\textbf{(\thesubtable) Loss weighting}\label{subtab:logvar}} \\
    \quad \ding{55} & 34.38 & 0.9380 & \textbf{0.052} & 19.94 \\
    \rowcolor{baselinecolor}
    \quad \ding{52} & \textbf{34.50} & \textbf{0.9405} & 0.053 & \textbf{19.27} \\
    \midrule
    \addlinespace[2.5ex] 

    \multicolumn{5}{@{}l}{\refstepcounter{subtable}\textbf{(\thesubtable) Long-range residual connection}\label{subtab:long_range_residual}} \\
    \quad \ding{55} & 34.39 & 0.9396 & \textbf{0.053} & 19.92 \\
    \rowcolor{baselinecolor}
    \quad \ding{52} & \textbf{34.50} & \textbf{0.9405} & \textbf{0.053} & \textbf{19.27} \\
    \midrule
    \addlinespace[2.5ex] 

    \multicolumn{5}{@{}l}{\refstepcounter{subtable}\textbf{(\thesubtable) RoPE \cite{su2024roformer}}\label{subtab:rope}} \\
    \rowcolor{baselinecolor}
    \quad \ding{55} & \textbf{34.39} & \textbf{0.9396} & \textbf{0.054} & \textbf{20.43} \\
    \quad \ding{52} & 33.04 & 0.9310 & 0.057 & 24.10 \\
    
    \bottomrule
    \end{tabularx}
\end{minipage}\hfill%
\begin{minipage}[t]{0.49\linewidth}
    \centering
    \small
    \begin{tabularx}{\linewidth}{@{} l *{4}{Y} @{}}
    \toprule
    Config. & PSNR$\uparrow$ & SSIM$\uparrow$ & LPIPS$\downarrow$ & FID$\downarrow$ \\
    \midrule

    \multicolumn{5}{@{}l}{\refstepcounter{subtable}\textbf{(\thesubtable) The design of JLA layer}\label{subtab:attn_layer}} \\
    \quad Linear & 34.20 & 0.9383 & 0.056 & 20.77 \\
    \rowcolor{baselinecolor}
    \quad Conv & 34.39 & 0.9397 & 0.054 & 20.59 \\
    \quad Vanilla & \textbf{34.42} & \textbf{0.9400} & \textbf{0.053} & \textbf{20.20} \\
    \midrule

    \multicolumn{5}{@{}l}{\refstepcounter{subtable}\textbf{(\thesubtable) The design of FFN layer}\label{subtab:ffn_layer}} \\
    \quad Linear & \textbf{34.39} & \textbf{0.9397} & \textbf{0.054} & 20.59 \\
    \rowcolor{baselinecolor}
    \quad Conv & \textbf{34.39} & 0.9396 & \textbf{0.054} & \textbf{20.43} \\
    \quad ConvGate & 33.94 & 0.9351 & 0.058 & 21.83 \\
    \midrule

    \multicolumn{5}{@{}l}{\refstepcounter{subtable}\textbf{(\thesubtable) Time modulation}\label{subtab:multiplier}} \\
    \quad $(\zeta_{\cdot}^{\cdot})$ & \textbf{34.40} & \textbf{0.9399} & \textbf{0.054} & 21.04 \\
    \rowcolor{baselinecolor}
    \quad $(\zeta_{\cdot}^{\cdot},\varepsilon_{\cdot}^{\cdot})$ & 34.39 & 0.9397 & \textbf{0.054} & \textbf{20.59} \\
    \quad $(\zeta_{\cdot}^{\cdot},\varepsilon_{\cdot}^{\cdot},\xi_{\cdot}^{\cdot})$ & 34.37 & 0.9386 & 0.058 & 20.85 \\
    \bottomrule
    \end{tabularx}
\end{minipage}

\end{table}

}

%% file: tables/time_sampler.tex
\begin{table}[t]
    \centering
    \caption{
    Effect of training time samplers.
    Models are trained via these time samplers and evaluated on three deterministic ($\eta=0$) trajectory:
    \textbf{Regression} ($\delta=0$, 1 step), \textbf{Elliptical} ($\delta=\pi/8$, 5 steps), and \textbf{Linear} ($\delta=\pi/8$, 5 steps).
    Lognorm1 samples $r,g \sim \mathsf{lognorm}(0,1)$; Lognorm2 samples $r \sim \mathsf{lognorm}(0,1)$ and $g \sim \mathsf{lognorm}(-0.5,1)$.
    \label{tab:time_sampler}
    }
    
    \resizebox{\textwidth}{!}{
    \begin{tabular}{c c cccc  cccc cccc}
    \toprule
    \multicolumn{2}{c}{\multirow{2}*{\shortstack{Time Samplers \\ of $(r, g)$}}} & \multicolumn{4}{c}{Regression Trajectory}&\multicolumn{4}{c}{Elliptical Trajectory} & \multicolumn{4}{c}{Linear Trajectory} \\
\cmidrule(lr){3-6} \cmidrule(lr){7-10} \cmidrule(lr){11-14} 
&& PSNR$\uparrow$ & SSIM$\uparrow$ & LPIPS$\downarrow$ & FID$\downarrow$ & PSNR$\uparrow$ & SSIM$\uparrow$ & LPIPS$\downarrow$ & FID$\downarrow$ & PSNR$\uparrow$ & SSIM$\uparrow$ & LPIPS$\downarrow$ & FID$\downarrow$ \\
    \midrule
        \multirow{3}*{\shortstack{Generalist \\ Samplers}}
        & Uniform 
        &34.16&0.9375&\best{0.057}&\best{20.70}
        &33.46&0.9283&0.051&21.23
        &32.61&0.9083&0.058&26.55\\
        & lognorm1
        &33.54&0.9378&0.058&23.75
        &\best{33.75}&0.9320&0.049&19.28 
        &32.80&0.9122&0.055&26.21\\
        & lognorm2
        &32.73&0.9301&0.061&27.31
        &33.61&\best{0.9379}&\best{0.048}&\best{18.63}
        &\best{32.94}&\best{0.9169}&\best{0.047}&\best{23.59}\\
        \hdashline
        \multirow{3}*{\shortstack{Specialist \\ Samplers}}
        & Regression 
        &31.91&0.9113&0.084&36.10
        &14.92&0.1163&1.302&309.53
        &9.57&-0.0397&1.440&443.87\\
        & Elliptical  
        &\best{34.54}&\best{0.9417}&0.058&20.84
        &33.51&0.9294&0.051&20.61
        &32.71&0.9110&0.056&26.44\\
        & Linear
        &33.41&0.9307&0.063&24.70
        &32.77&0.9221&0.059&23.44
        &32.04&0.9036&0.059&29.69\\
    \bottomrule
    \end{tabular}
    }
    
\end{table}

%% file: tables/sampler.tex
\begin{table}[t]
    \centering
    \caption{
        Ablation of $\delta$, $\eta$, and NFE on Elliptical path.
        Regression mode is in \colorbox{gray!15}{gray}.
        \label{tab:hyperparameter}
    }
    \begingroup
    \footnotesize 
    \renewcommand{\arraystretch}{0.95} 
    \setlength{\tabcolsep}{2pt} 

    \begin{tabularx}{0.95\linewidth}{@{} c c *{5}{YY} @{}}
        \toprule
        \multirow{2}{*}{\textbf{$\delta$}} & \multirow{2}{*}{\textbf{$\eta$}} & \multicolumn{2}{c}{\textbf{NFE = 1}} & \multicolumn{2}{c}{\textbf{NFE = 2}} & \multicolumn{2}{c}{\textbf{NFE = 5}} & \multicolumn{2}{c}{\textbf{NFE = 15}} & \multicolumn{2}{c}{\textbf{NFE = 50}} \\
        \cmidrule(lr){3-4} \cmidrule(lr){5-6} \cmidrule(lr){7-8} \cmidrule(lr){9-10} \cmidrule(l){11-12}
        & & PSNR$\uparrow$ & LPIPS$\downarrow$ & PSNR$\uparrow$ & LPIPS$\downarrow$ & PSNR$\uparrow$ & LPIPS$\downarrow$ & PSNR$\uparrow$ & LPIPS$\downarrow$ & PSNR$\uparrow$ & LPIPS$\downarrow$ \\
        \midrule

        \rowcolor{baselinecolor}
        0 & \na & \textbf{34.92} & 0.055 & 34.91 & 0.055 & 34.79 & 0.052 & 34.62 & \textbf{0.050} & 34.52 & \textbf{0.050} \\
        \addlinespace %

        \multirow{4}{*}{$\cfrac{\pi}{8}$} 
        & 0.0 & \na & \na & \textbf{34.92} & 0.055 & 33.85 & 0.049 & 32.85 & \textbf{0.047} & 32.37 & \textbf{0.047} \\
        & 0.2 & \na & \na & \textbf{34.92} & 0.055 & 31.61 & 0.092 & 30.95 & 0.094 & 31.34 & 0.071 \\
        & 0.5 & \na & \na & \textbf{34.92} & 0.055 & 26.94 & 0.289 & 27.86 & 0.241 & 30.30 & 0.125 \\
        & 1.0 & \textbf{34.92} & 0.055 & \textbf{34.92} & 0.055 & 24.27 & 0.537 & 28.53 & 0.280 & 31.82 & 0.138 \\
        \midrule

        \multirow{4}{*}{$\cfrac{\pi}{4}$} %
        & 0.0 & \na & \na & \textbf{34.92} & 0.055 & 34.29 & 0.052 & 33.12 & 0.050 & 32.41 & 0.049 \\
        & 0.2 & \na & \na & \textbf{34.92} & 0.055 & 29.86 & 0.169 & 30.29 & 0.127 & 30.97 & 0.082 \\
        & 0.5 & \na & \na & \textbf{34.92} & 0.055 & 22.73 & 0.545 & 25.74 & 0.336 & 29.07 & 0.180 \\
        & 1.0 & \textbf{34.92} & 0.055 & \textbf{34.92} & 0.055 & 19.58 & 0.917 & 24.20 & 0.558 & 28.54 & 0.289 \\
        \midrule

        \multirow{4}{*}{$\cfrac{\pi}{2}$} 
        & 0.0 & \na & \na & \textbf{34.92} & 0.055 & 34.78 & 0.053 & 33.67 & 0.052 & 32.65 & 0.051 \\
        & 0.2 & \na & \na & \textbf{34.92} & 0.055 & 27.07 & 0.319 & 29.61 & 0.169 & 30.63 & 0.104 \\
        & 0.5 & \na & \na & \textbf{34.92} & 0.055 & 18.69 & 0.919 & 22.71 & 0.530 & 27.35 & 0.264 \\
        & 1.0 & \textbf{34.92} & 0.055 & \textbf{34.92} & 0.055 & 16.28 & 1.177 & 20.45 & 0.853 & 25.14 & 0.496 \\
        
        \bottomrule
    \end{tabularx}
    \endgroup
\end{table}

%% file: sec/5_conclusion.tex
\section{Conclusion}
\label{sec:conclusion}
We introduce DiSI, a unified SI-based framework for controllable IR that decouples regression and generation, enabling a single model to flexibly balance distortion and perception.
Our implementation integrates two sampling trajectories, an analytically derived sampler for fast, high-quality inference, and a dual-branch pixel-space network that strengthens conditional guidance with high efficiency.
Extensive experiments validate the effectiveness and efficiency of DiSI for IR.

%% file: sec/X_suppl.tex
\clearpage
\setcounter{page}{1}

{
    \newpage
    \centering
    \Large
    \textbf{Disentangling Generation and Regression in Stochastic Interpolants for Controllable\\Image Restoration}\\
    \vspace{2mm}
    \textbf{Appendix} \label{app:appendix}\\
}
\appendix
\section{Proofs and Derivations}
\label{app:detrivations}

Our proofs and derivations are based on the frameworks of Stochastic Interpolants~\cite{albergo2023stochastic} and SiT~\cite{ma2024sit}, with specific adaptations for our DiSI framework.

\subsection{Proof of the GVP Property in Eq. (\ref{eq:disi_coeffcients})}
\label{app:proof_of_gvp}
\begin{proof}
Substituting \cref{eq:disi_coeffcients} into \cref{eq:disi} yields the noisy state $\bfxrg$:
\begin{equation}
    \begin{split}
        \bfxrg &= \cos{g} \lsb\ar \bfx_0 \rdot\\
            &\ldot \quad + \br \bfx_1\rsb + \sin{g}\bfz.
    \end{split}
    \label{eq:detailed_disi}
\end{equation}
The variance of $\bfxrg$ is given by:
\begin{align}
& \begin{aligned}
     &\var\lb\bfx(r,g)\rb = \sin^2{g}\sigmad^2 + \cos^2{g} \lsb
        \frac{1}{2}\lb
            \frac{\cos^2{r}}{\rrho}+\frac{\sin^2r}{\rmrho} - \frac{2\cos{r}\sin{r}}{\rsqrhos}
        \rb\sigmad^2 \rdot\\
        &\ldot\quad+ 
        \frac{1}{2}\lb
            \frac{\cos^2{r}}{\rrho}+\frac{\sin^2r}{\rmrho} + \frac{2\cos{r}\sin{r}}{\rsqrhos}
        \rb\sigmad^2+
    \lb
        \frac{\cos^2{r}}{\rrho} - \frac{\sin^2{r}}{\rmrho}
    \rb\sigmad^2\rho
    \rsb 
\end{aligned}\\
    &\qquad= \sin^2{g}\sigmad^2 + \cos^2{g} \lsb \lb
            \frac{\cos^2r}{\rrho}+\frac{\sin^2r}{\rmrho}
        \rb\sigmad^2  +
    \lb
        \frac{\cos^2{r}}{\rrho} - \frac{\sin^2{r}}{\rmrho}
    \rb\sigmad^2\rho
    \rsb \\
    &\qquad= \sin^2{g}\sigmad^2  +\cos^2{g}\lb
        \cos^2r+\sin^2r
    \rb \sigmad^2 = \sigmad^2 \label{eq:gvp_final_results}, 
\end{align}
where $\sigmad$ is the standard deviation of $\bfx_0$, $\bfx_1$, and $\bfz$, and $\rho$ is the correlation coefficient between $\bfx_0$ and $\bfx_1$ (\ie, $\cov(\bfx_0, \bfx_1) = \rho\sigmad^2$).
In practice, we pre-normalize the clean image $\bfx_0$ and the degraded image $\bfx_1$ to share this standard deviation (typically $\sigmad = 1$). 
The noise $\bfz \sim \matN(0, \sigmad^2\bfI)$ is sampled to match the standard deviation $\sigmad$, as defined in \cref{sec:preliminary}. 
Consequently, \cref{eq:gvp_final_results} demonstrates that $\var(\bfxrg)$ remains constant at $\sigmad^2$ for all values of $r$ and $g$.
This completes the proof of the Generalized Variance Preserving (GVP) property of \cref{eq:disi_coeffcients}.
\end{proof}

\subsection{Derivation of the Velocity Forms in Eq. (\ref{eq:rel_net_and_velrg})}
\label{app:derivation_of_vel}
The marginal distribution $p_{r,g}(\bfx)$ of the interpolated state $\bfxrg$ is governed by the following probability flow ordinary differential equation (PF-ODE):
\begin{equation}
    \ud\bfx_{r,g} = \bfv(\bfx_{r,g}, r)\ud t + \bfv(\bfx_{r,g}, g)\ud g,
    \label{eq:disi_pfode_detail}
\end{equation}
where $\bfv(\bfx_{r,g}, r)$ and $\bfv(\bfx_{r,g}, g)$ represent the velocity fields associated with the temporal variables $r$ and $g$, respectively. 
For brevity, we adopt the shorthand notation $\bfx_{r,g}$ for the state $\bfxrg$. 
Based on the proof of the transport equation in \cref{app:proof_of_transport_equation} detailed later, these two velocity fields are defined as:
\begin{flalign}
    &\bfv(\bfx_{r,g}, r) = \bbE\lsb \frac{\partial \bfx_{r,g}}{\partial r} \Bigg|\bfx_{r,g} = \bfx \rsb 
    = \bbE\lsb \lambda_g\lb
        \dot\alpha_r\bfx_0 + \dot\beta_r\bfx_1
    \rb\Big| \bfx_{r,g} = \bfx \rsb,  \label{eq:vr_expectation} \\
    &\bfv(\bfx_{r,g}, g) = \bbE\lsb \frac{\partial \bfx_{r,g}}{\partial g} \Bigg|\bfx_{r,g} = \bfx \rsb = \bbE\lsb
        \dot\lambda_g\lb
        \alpha_r\bfx_0 + \beta_r\bfx_1
        \rb +\dot\gamma_g \bfz \Big|\bfx_{r,g} = \bfx
    \rsb,  \label{eq:vg_expectation}
\end{flalign}
where the dot notation denotes derivatives \wrt the corresponding time variables.
Given the GVP schedule in \cref{eq:disi_coeffcients}, the explicit derivatives are:
\begin{equation}
    \begin{aligned}
        & \dot\alpha_r = \dotar, %
        &\dot\beta_r &= \dotbr, %
        \\
        &\dot\lambda_g =-\sin{g}, %
        &\dot\gamma_g &= \cos{g}. %
    \end{aligned}
    \label{eq:disi_coeffcients_derivatives}
\end{equation}

While Diffusion Models (DMs)~\cite{song2020score,ho2020denoising} and Flow Matching (FM) methods~\cite{lipman2022flow,liu2023flow} typically use a network to predict the velocity field, we instead predict the clean image $\bfx_0$, as discussed in \cref{sec:framework}.
Let $\hat\bfx_0$ denote the network prediction of $\bfx_0$.
We then derive the explicit mappings from $\hat\bfx_0$ to the two velocity fields.

\paragraph{Derivation of $\bfv(\bfx_{r,g}, r)$}
Expanding the expectation in \cref{eq:vr_expectation} yields:
\begin{align}
    \bfv(\bfx_{r,g}, r) &= \bbE\Big[ \lambda_g
        \dot\alpha_r\bfx_0\bcondxrg\Big] + \bbE\lsb\lambda_g\dot\beta_r\bfx_1 \bcondxrg 
    \rsb  \label{eq:der_rel_vr_and_x0_1} \\
    &= \lambda_g \dot\alpha_r\bbE\lsb\bfx_0\condxrg\rsb + \lambda_g\dot\beta_r\bfx_1, \label{eq:der_rel_vr_and_x0_2}
\end{align}
where \cref{eq:der_rel_vr_and_x0_2} holds because $\bfx_1$ (\ie, the degraded input) is deterministic and can be factored out of the expectation.
The term $\bbE\lsb\bfx_0\condxrg\rsb$ in \cref{eq:der_rel_vr_and_x0_2} is the minimum mean squared error (MMSE) estimator of the clean image given the noisy state.
In practice, we approximate $\bbE\lsb\bfx_0\condxrg\rsb$ with the network prediction $\hat{\bfx}_0$, as detailed in \cref{eq:net_prediction}.
Replacing $\bbE\lsb\bfx_0\condxrg\rsb$ with $\hat\bfx_0$ in \cref{eq:der_rel_vr_and_x0_2} and recalling that $\lambda_g=\cos g$, we obtain the parameterized velocity:
\begin{equation}
   \bfv(\bfx_{r,g}, r) = \lambda_g \dot\alpha_r\hat\bfx_0+ \lambda_g\dot\beta_r\bfx_1 = \cos{g}\lb\dot\alpha_r \hat\bfx_0+\dot\beta_r\bfx_1\rb,
   \label{eq:vr_rel_and_coeff}
\end{equation}
which corresponds to \cref{eq:rel_net_and_velrg}. 
We denote $\bfv(\bfx_{r,g}, r)$ by $\bfv_r^\theta$ in \cref{eq:rel_net_and_velrg} to emphasize its dependence on the network parameters $\theta$.
See \cref{eq:disi_coeffcients_derivatives} for $\dot\alpha_r$ and $\dot\beta_r$.

\paragraph{Derivation of $\bfv(\bfx_{r,g}, g)$}
Factoring the deterministic $\bfx_1$ (\ie, the degraded input) out of the conditional expectation, we expand \cref{eq:vg_expectation} to obtain:
\begin{align}
    \bfv(\bfx_{r,g}, g) 
    =\dot\lambda_g \alpha_r \bbE\lsb
        \bfx_0 \condxrg
    \rsb + 
        \dot\lambda_g\beta_r\bfx_1 
    + \dot\gamma_g \bbE\lsb 
        \bfz  \condxrg
    \rsb, \label{eq:vr_to_x0_2}
\end{align}
From the definition of DiSI process in \cref{eq:disi}, we have $\bfz = \frac{\bfx_{r,g}- \lambda_g\lb\alpha_r\bfx_0 + \beta_r\bfx_1\rb}{\gamma_g}$.
Substituting this into \cref{eq:vr_to_x0_2} and regrouping the terms, we obtain:
\begin{equation}
    \begin{split}
    \bfv(\bfx_{r,g}, g) &= \dot\lambda_g \alpha_r \bbE\lsb
    \bfx_0 \condxrg
    \rsb + 
        \dot\lambda_g\beta_r\bfx_1 
    \\
    & + \dot\gamma_g \bbE\lsb 
        \frac{\bfx_{r,g}- \lambda_g\lb\alpha_r\bfx_0 + \beta_r\bfx_1\rb}{\gamma_g}  \bbcondxrg
    \rsb.
    \end{split}
    \label{eq:derivation_of_velg_2}
\end{equation}
Expanding the second expectation in \cref{eq:derivation_of_velg_2}, we obtain:
\begin{equation}
    \begin{split}
    \bfv(\bfx_{r,g}, g) &= \dot\lambda_g \alpha_r \bbE\lsb
        \bfx_0 \condxrg
    \rsb + 
        \dot\lambda_g\beta_r\bfx_1 
    + \frac{\dot\gamma_g}{\gamma_g} \bbE\lsb 
        \bfx_{r,g}\condxrg
    \rsb \\
    &- \frac{\dot\gamma_g\lambda_g\alpha_r}{\gamma_g}\bbE\lsb 
        \bfx_0 \condxrg
    \rsb- \frac{\dot\gamma_g\lambda_g\beta_r}{\gamma_g}\bbE\lsb 
        \bfx_1 \condxrg
    \rsb.
    \end{split}
    \label{eq:derivation_of_velg_3}
\end{equation}
Replacing $\bbE\lsb
        \bfx_0 \condxrg
    \rsb$ with $\hat\bfx_0$ and simplifying \cref{eq:derivation_of_velg_3} yields
\begin{align}
    \bfv(\bfx_{r,g}, g)  = \frac{\dot\lambda_g\gamma_g - \lambda_g\dot\gamma_g}{\gamma_g}\alpha_r\hat\bfx_0 +
    \frac{\dot\lambda_g\gamma_g -\lambda_g\dot\gamma_g}{\gamma_g}\beta_r\bfx_1 
    + \frac{\dot\gamma_g}{\gamma_g}\bfx.
    \label{eq:derivation_of_velg_6}
\end{align}
Applying the coefficients and their derivatives from \cref{eq:disi_coeffcients,eq:disi_coeffcients_derivatives} gives:
\begin{align}
    \bfv(\bfx_{r,g}, g) 
    & = \frac{-\sin^2{g} - \cos^2{g}}{\sin{g}}\alpha_r\hat\bfx_0 +
    \frac{-\sin^2{g} -\cos^2{g}}{\sin{g}}\beta_r\bfx_1 
    + \frac{\cos{g}}{\sin{g}}\bfx \\
    & = \cot{g}\bfx - \csc{g}\lb
        \alpha_r \hat\bfx_0 + \beta_r\bfx_1
    \rb.
\end{align}
This matches \cref{eq:rel_net_and_velrg}, where we denote the velocity $\bfv(\bfx_{r,g}, g)$ as $\bfv_{g}^{\theta}$ to emphasize its dependence on the network parameters $\theta$. 
Note that $\bfxrg$ is used interchangeably with $\bfx$ to explicitly highlight its temporal dependency.

\subsection{Derivation of the Sampler in Eqs. (\ref{eq:disi_ddim_sampler}) and (\ref{eq:regressive_sampler})}
\label{app:derivation_of_sampler}
This section derives the underlying mathematical formulations of DiSI's sampler.
We first formulate the general hybrid update rule in \cref{eq:disi_ddim_sampler}, and subsequently deduce the deterministic regression sampler in \cref{eq:regressive_sampler} as a specialized case.

\subsubsection{General Hybrid Sampler in \cref{eq:disi_ddim_sampler}.}
\label{app:general_hybrid_sampler}
Recall that the generation velocity $\bfvgf$ derived in \cref{eq:derivation_of_velg_6} explicitly involves division by $\gamma_g$. This induces a singularity at $g=0$ (since $\gamma_0 = 0$ under our GVP schedule), which is typically encountered at the start of the Elliptical path. 
This stems from the implicit estimation of $\bfz$ in \cref{eq:derivation_of_velg_2}.
To circumvent this, inspired by prior stochastic samplers~\cite{song2021denoising,zheng2025diffusion}, we inject a controlled stochastic term into $\bfvgf$ by replacing a portion of the estimated noise with fresh noise $\bfz \sim \matN(\boldsymbol{0}, \sigmad^2\bfI)$:
\begin{equation}
    \begin{split}
    \bftvgf 
    &= \dot\lambda_g \alpha_r \bbE\lsb
        \bfx_0 \condxrg
    \rsb + 
        \dot\lambda_g\beta_r\bfx_1 
    \\
    &+ \retas \dot\gamma_g \bbE\lsb 
        \frac{\bfx_{r,g}- \lambda_g\lb\alpha_r\bfx_0 + \beta_r\bfx_1\rb}{\gamma_g}  \bbcondxrg
    \rsb +  \eta\dot\gamma_g \bfz,
    \end{split}
    \label{eq:ddim_vel_g}
\end{equation}
where $\eta \in [0, 1]$ controls the level of stochasticity, with $\eta=0$ corresponding to the deterministic case and $\eta=1$ to the fully stochastic case.
Setting $\eta=1$ removes the $\gamma_g^{-1}$ term, thereby avoiding the singularity at $g=0$.
Following a derivation analogous to \cref{eq:derivation_of_velg_2,eq:derivation_of_velg_3}, the modified velocity simplifies to:
\begin{equation}
    \begin{split}
    \bftvgf 
    & = \frac{\dot\lambda_g\gamma_g- \retas\lambda_g\dot\gamma_g}{\gamma_g} \alpha_r\hat\bfx_0 \\
    &+
    \frac{\dot\lambda_g\gamma_g - \retas\lambda_g\dot\gamma_g}{\gamma_g}\beta_r\bfx_1 +
    \retas \frac{\dot\gamma_g}{\gamma_g} \bfx_{r,g} + \eta\dot\gamma_g\bfz.
    \end{split}
    \label{eq:ddim_vel_g_detail}
\end{equation}
Substituting it and $\bfvrf$ in \cref{eq:vr_rel_and_coeff} into the PF-ODE of DiSI in \cref{eq:disi_pfode_detail} and reorganizing the terms gives:
\begin{equation}
    \begin{aligned}
    \ud\bfx_{r,g} &=
    \underbrace{\retas \frac{\dot\gamma_g}{\gamma_g} \bfx_{r,g} \ud g}_{(1)\text{ linear term}} 
    + 
    \underbrace{\lsb
        \lambda_g \dot\alpha_r\ud r +  \frac{\dot\lambda_g\gamma_g- \retas\lambda_g\dot\gamma_g}{\gamma_g} \alpha_r \ud g
    \rsb \hat\bfx_0}_{(2) \text{ non-linear term}} \\
    &+
    \underbrace{\lsb
        \lambda_g \dot\beta_r\ud r + \frac{\dot\lambda_g\gamma_g - \retas\lambda_g\dot\gamma_g}{\gamma_g}\beta_r\ud g
    \rsb\bfx_1 +
    \eta\dot\gamma_g\bfz\ud g}_{(3) \text{ bias term}}.
    \end{aligned}
    \label{eq:disi_ddim_pfode_with_vel_reorg} 
\end{equation}
This formulation decomposes the ODE into a semi-linear structure comprising three components:
(1) a linear term proportional to $\bfx_{r,g}$; 
(2) a non-linear term driven by the network prediction $\hat\bfx_0$, which implicitly depends on $\bfx_{r,g}$; and 
(3) a bias term involving $\bfx_1$ and $\bfz$, independent of $\bfx_{r,g}$.
This form enables the use of exponential integrators~\cite{lu2022dpm} via \textit{variation of constants}. 
We parameterize the sampling trajectory as $(r(t), g(t))$ with independent variable $t$, and integrate from $t_1$ to $t_2$.
For brevity, we write $r_i = r(t_i)$ and $g_i = g(t_i)$ ($i=1,2$):
%
%
%
\begin{equation}
    \begin{split}
    & \bfx_{r_2,g_2} 
    = \underbrace{K(t_1) \bfx_{r_1,g_1}}_{(1)}   + \underbrace{\int_{t_1}^{t_2}
        K(\tau) 
            \lambda_{g(\tau)}\drt 
            \lb
                \dot\alpha_{r(\tau)}\hat\bfx_0 + \dot\beta_{r(\tau)}\bfx_1
            \rb
            \ud \tau
        }_{(2)}\\
    & + \underbrace{\int_{t_1}^{t_2}
        K(\tau)
            \frac{\dot\lambda_{g(\tau)}\gamma_{g(\tau)}- \retas\lambda_{g(\tau)}\dot\gamma_{g(\tau)}}{\gamma_{g(\tau)}} \dgt
            \lb
                \alpha_{r(\tau)}\hat\bfx_0 + \beta_{r(\tau)}\bfx_1
            \rb
            \ud \tau
        }_{(3)}\\    
    & + \underbrace{\int_{t_1}^{t_2}
        K(\tau)
            \eta\dot\gamma_g\bfz\dgt\ud\tau
        }_{(4)}\\    
    \end{split},
    \label{eq:variation_of_constants_1}
\end{equation}
where 
\begin{equation}
    K(t) =  \exp\lb{\int_{t}^{t_2} \retas\frac{\dot\gamma_{g(\tau)}}{\gamma_{g(\tau)}}\dot g(\tau)\ud\tau}\rb.
\end{equation}
Applying the chain rule, the integrating factor $K(t)$ evaluates to:
\begin{equation}
    K(t)= \exp\lb{
        \retas \int_t^{t_2}
        \frac{d}{d\tau}\log \gamma_{g(\tau)} \,\ud\tau
    }\rb 
    =\frac{\gamma_{g_2}^{\retas}}{\gamma_{g(t)}^{\retas}}.
    \label{eq:exp_varaition_of_constants}
\end{equation}
Substituting the closed-form expression of $K(\tau)$ into term (2) of \cref{eq:variation_of_constants_1} and using $\ud r(\tau)=\dot r(\tau)\,\ud\tau$, we obtain:
\begin{equation}
     \begin{split}
     (2)
        =
        \int_{t_1}^{t_2}\frac{\gamma_{g_2}^{\retas}}{\gamma_{g(\tau)}^{\retas}}
        \lambda_{g(\tau)}
        \lb
            \dot\alpha_{r(\tau)}\hat\bfx_0 + \dot\beta_{r(\tau)}\bfx_1
        \rb  
        \ud r(\tau),
    \end{split}
    \label{eq:variation_of_constants_2}
\end{equation}
where the network prediction $\hat\bfx_0$ depends implicitly on $\tau$ via $r(\tau)$ and $g(\tau)$, as detailed in \cref{eq:net_prediction}.
Following DPM-Solver~\cite{lu2022dpm}, we analytically approximate this temporal dependence using a Taylor expansion around $t_1$:
\begin{equation}
    \begin{split}
    \hat\bfx_0 &= \sigmad \net\lb\frac{\bfx_{r,g}}{\sigmad}, \bfx_1, r(\tau), g(\tau)\rb
    \\ 
    &= \sigmad
    \lsb \sum_{k=0}^n  
        \frac{{(\tau - t_1)}^k}{k!}  \net^{(k)}\lb\frac{\bfx_{r,g}}{\sigmad}, \bfx_1, r({t_1}), g({t_1})\rb
        + \mathcal{O}(h^{n+1})
    \rsb,
    \end{split}
    \label{eq:taylor_expand_net}
\end{equation}
where $\net^{(k)}$ denotes the $k$-th derivative of $\net$ \wrt $\tau$.
While higher-order solvers ($n \ge 1$) are available, the first-order approximation ($n=0$), corresponding to DDIM~\cite{song2021denoising}, is sufficient for few-step IR.
Accordingly, the network prediction $\hat\bfx_0\approx\sigmad \net\lb\frac{\bfx_{r,g}}{\sigmad}, \bfx_1, r({t_1}), g({t_1})\rb$ is fixed at $t_1$ and treated as constant \wrt $r(\tau)$. 
Consequently, \cref{eq:variation_of_constants_2} simplifies to:
\begin{equation}
    (2)
    =
    \int_{t_1}^{t_2}\frac{\gamma_{g_2}^{\retas}}{\gamma_{g(\tau)}^{\retas}}
        \lambda_{g(\tau)}
        \ud \lb\alpha_{r(\tau)}\hat\bfx_0 + \beta_{r(\tau)}\bfx_1 \rb.
\end{equation}
Applying integration by parts yields:
\begin{equation}
    \begin{split}
    (2) &= \frac{\gamma_{g_2}^{\retas}}{\gamma_{g(\tau)}^{\retas}}
        \lambda_{g(\tau)}
        \lb\alpha_{r(\tau)}\hat\bfx_0 + \beta_{r(\tau)}\bfx_1 \rb\Bigg|_{t_1}^{t_2}\\
    &- 
    \int_{t_1}^{t_2}\ud\lb\frac{\gamma_{g_2}^{\retas}}{\gamma_{g(\tau)}^{\retas}}
        \lambda_{g(\tau)} \rb
         \lb\alpha_{r(\tau)}\hat\bfx_0 + \beta_{r(\tau)}\bfx_1 \rb.
    \end{split}
    \label{eq:variation_of_constants_3}
\end{equation}
The remaining integral term can be written as:
\begin{equation}
    \begin{split}
    &\int_{t_1}^{t_2}\ud\lb\frac{\gamma_{g_2}^{\retas}}{\gamma_{g(\tau)}^{\retas}}
        \lambda_{g(\tau)} \rb
         \lb\alpha_{r(\tau)}\hat\bfx_0 + \beta_{r(\tau)}\bfx_1 \rb 
    =\int_{t_1}^{t_2}   \lsb \frac{\gamma_{g_2}^{\retas}}{\gamma_{g(\tau)}^{\retas}} \rdot\\
    &\qquad  \ldot        
        \lb
            \frac{\dot\lambda_{g(\tau)} \gamma_{g(\tau)} -  \retas\lambda_{g(\tau)}\dot\gamma_{g(\tau)}}
            {\gamma_{g(\tau)}}
        \rb \lb\alpha_{r(\tau)}\hat\bfx_0 + \beta_{r(\tau)}\bfx_1 \rb  
    \dgt\rsb\ud\tau.
    \end{split}
    \label{eq:variation_of_constants_4}
\end{equation}
This exactly cancels term (3) in \cref{eq:variation_of_constants_1}, leaving only the boundary term:
\begin{equation}
    \begin{split}
    & \frac{\gamma_{g_2}^{\retas}}{\gamma_{g(\tau)}^{\retas}}
        \lambda_{g(\tau)}
        \lb\alpha_{r(\tau)}\hat\bfx_0 + \beta_{r(\tau)}\bfx_1 \rb\Bigg|_{t_1}^{t_2} \\
    & \qquad = \lambda_{g_2} \lb
        \alpha_{r_2}\hat\bfx_0 + \beta_{r_2}\bfx_1
    \rb -  \lambda_{g_1}\frac{\gamma_{g_2}^{\retas}}{\gamma_{g_1}^{\retas}}\lb
        \alpha_{r_1}\hat\bfx_0 + \beta_{r_1}\bfx_1
    \rb.
    \end{split}
    \label{eq:variation_of_constants_5}
\end{equation}
Finally, term (4) evaluates to
\begin{equation}
    (4) =  
    \int_{g_1}^{g_2}
    \eta\bfz\frac{\gamma_{g_2}^{\retas}}{\gamma_{g(\tau)}^{\retas}}\ud \gamma_g
    = \lb
        \frac{\gamma_{g_1}^{\retas}\gamma_{g_2} - \gamma_{g_1}\gamma_{g_2}^{\retas}}{\gamma_{g_1}^{\retas}}
    \rb\frac{\eta\bfz}{1-\retas}.
    \label{eq:variation_of_constants_6}
\end{equation}
Combining the results from \cref{eq:variation_of_constants_2,eq:variation_of_constants_3,eq:variation_of_constants_4,eq:variation_of_constants_5,eq:variation_of_constants_6}, \cref{eq:variation_of_constants_1} simplifies to:
\begin{equation}
    \begin{split}
    \bfx_{r_2,g_2} &= \frac{\gamma_{g_2}^{\retas}}{\gamma_{g_1}^{\retas}} \bfx_{r_1, g_1}  \\
    & + \lambda_{g_2} \lb
        \alpha_{r_2}\hat\bfx_0 + \beta_{r_2}\bfx_1
    \rb
    -  \lambda_{g_1}\frac{\gamma_{g_2}^{\retas}}{\gamma_{g_1}^{\retas}}\lb
        \alpha_{r_1}\hat\bfx_0 + \beta_{r_1}\bfx_1
    \rb + \kappa \bfz,
    \end{split}
    \label{eq:unified_disi_sampling_step}
\end{equation}
where the noise coefficient $\kappa$ is defined as:
\begin{equation}
    \kappa = \mathbb{I}(\eta \neq0)\cdot \lb
        \frac{\gamma_{g_1}^{\retas}\gamma_{g_2} - \gamma_{g_1}\gamma_{g_2}^{\retas}}{\gamma_{g_1}^{\retas}}
    \rb\frac{\eta}{1-\retas}.
    \label{eq:kappa_unified_form}
\end{equation}
The indicator function $\mathbb{I}(\eta \neq 0)$ is explicitly introduced to ensure that the stochastic bias term naturally vanishes in deterministic sampling ($\eta=0$).

Substituting the GVP schedules from \cref{eq:disi_coeffcients} into \cref{eq:unified_disi_sampling_step,eq:kappa_unified_form} yields
\begin{equation}
    \begin{split}
    \bfx_{r_2,g_2} &= k^{\retas} \bfx_{r_1, g_1}  \\
    &+ \cos{g_2} \lb
        \alpha_{r_2}\hat\bfx_0 + \beta_{r_2}\bfx_1
    \rb -  k^{\retas}\cos{g_1}\lb
        \alpha_{r_1}\hat\bfx_0 + \beta_{r_1}\bfx_1
    \rb + \kappa \bfz,
    \end{split}
    \label{eq:unified_disi_sampling_step_gvp}
\end{equation}
where $k \coloneqq \frac{\sin g_2}{\sin g_1}$, and $\kappa$ evaluates to:
\begin{equation}
    \kappa = \mathbb{I}(\eta \neq0)\cdot 
    \frac{\eta
    \lb
        \sin{g_2} - k^{\retas}\sin{g_1}
    \rb
    }{1-\retas}.
    \label{eq:kappa_unified_form_gvp} 
\end{equation}
These derived results perfectly recover the sampling formulas in \cref{eq:disi_ddim_sampler,eq:kappa}.

\subsubsection{Regression Sampler in \cref{eq:regressive_sampler}.}
\label{sec:regression_sampler}
In the pure regression case where $g=0$ (and thus $\lambda_g = \lambda_0$), $\bfvgf$ vanishes, simplifying the PF-ODE in \cref{eq:disi_pfode_detail} to:
\begin{equation}
    \ud\bfx_{r,g} =
        \lb
            \lambda_0\dot\alpha_r\hat\bfx_0 + \lambda_0\dot\beta_r\bfx_1
        \rb
    \ud r.
    \label{eq:disi_regression_pfode}
\end{equation}
Integrating \cref{eq:disi_regression_pfode} over the interval $[t_1,t_2]$ gives:
\begin{equation}
    \bfx_{r_2,g_2} = \bfx_{r_1,g_1} + \int_{t_1}^{t_2} \lb \lambda_{0}\dot\alpha_{r(t)}\hat\bfx_0 + \lambda_{0}\dot\beta_{r(t)}\bfx_1
    \rb \dr \udt.
\end{equation}
As in the general hybrid sampler, we retain the first-order Taylor term in \cref{eq:taylor_expand_net}.
Accordingly, $\hat{\bfx}_0$ is assumed to be constant \wrt $t$ during integration, yielding:
\begin{equation}
\begin{split}
    \bfx_{r_2,g_2} &= \bfx_{r_1,g_1} + 
    \lambda_0 \hat\bfx_0\int_{r_1}^{r_2}\dot\alpha_{r(t)} \ud r(t) + 
    \lambda_0 \bfx_1\int_{r_1}^{r_2}\dot\beta_{r(t)}\ud r(t)\\
    &= \bfx_{r_1, g_1} + \lambda_0 (\alpha_{r_2} - \alpha_{r_1}) \hat\bfx_0 + \lambda_0 (\beta_{r_2} - \beta_{r_1}) \bfx_1.
\end{split}
    \label{eq:disi_universal_regression_sampler}
\end{equation}
Under the GVP schedule where $\lambda_0 = \cos(0) = 1$, \cref{eq:disi_universal_regression_sampler} simplifies to
\begin{equation}
    \bfx_{r_2,g_2} = \bfx_{r_1, g_1} + (\alpha_{r_2} - \alpha_{r_1}) \hat\bfx_0 +  (\beta_{r_2} - \beta_{r_1}) \bfx_1,
    \label{eq:disi_gvp_regression_sampler}
\end{equation}
which recovers the deterministic regression sampler in \cref{eq:regressive_sampler}.

\subsection{Proof of the PF-ODE in Eq. (\ref{eq:disi_pfode_detail}) and the Velocity Fields in Eqs. (\ref{eq:vr_expectation}) and (\ref{eq:vg_expectation}).}
\label{app:proof_of_transport_equation}
\begin{proof}
Let $\ptx$ be the time-dependent probability density function (PDF) of $\bfxrg$ in \cref{eq:disi}. 
Its characteristic function $\hptx$ is given by:
\begin{equation}
    \hptk = \int_{\bbR^d} e^{i\bfk\cdot\bfx}\ptx\ud\bfx = \bbE\lsb
        e^{i\bfk\cdot\bfx_{r,g}}
    \rsb,
    \label{eq:char_func_of_xrg}
\end{equation}
where $i$ is the imaginary unit, and $\bfxrg$ is simplified to $\bfx_{r,g}$ for notational brevity. 
The expectation is taken over $\bfx_0$, $\bfx_1$ and $\bfz$ in \cref{eq:disi}. 
Unlike standard SIs, $\ptx$ depends on two temporal variables: the regression time $r$ and the generation time $g$.
Accordingly, we then derive the corresponding velocity fields along the two time axes separately.

\paragraph{Derivation of $\bfv(\bfx_{r,g},r)$}
Taking the partial derivative of \cref{eq:char_func_of_xrg} \wrt $r$:
\begin{align}
    \partial_r\hptk 
    &= i\bfk\cdot \bbE\lsb
        \partial_r\bfx_{r,g} e^{i\bfk\cdot\bfx_{r,g}}
    \rsb \\
    &= i\bfk\cdot \bbE_{\bfx\sim \ptx}\lsb
         \bbE\lsb\partial_r\bfx_{r,g} e^{i\bfk\cdot\bfx_{r,g}} \condxrg\rsb
    \rsb \label{eq:tower_property_of_expectation}\\
    &= i\bfk\cdot \bbE_{\bfx\sim \ptx}\lsb
         \bbE\lsb\partial_r\bfx_{r,g} \condxrg\rsb e^{i\bfk\cdot\bfx}
    \rsb \\
    &= i\bfk\cdot \bbE_{\bfx\sim \ptx}\lsb
         \bfv\lb\bfx_{r,g}, r\rb e^{i\bfk\cdot\bfx}
    \rsb \label{eq:partial_r_char_func},
\end{align}
where \cref{eq:tower_property_of_expectation} applies the tower property of conditional expectation, and the velocity field $\bfvrf$ is defined as
\begin{equation}
    \bfv\lb\bfx_{r,g}, r\rb \coloneqq \bbE\lsb\partial_r\bfx_{r,g} \condxrg\rsb.
    \label{eq:vel_r_detail}
\end{equation}
Expanding the expectations on both sides of \cref{eq:partial_r_char_func}, we obtain
\begin{equation}
    \partial_r \int_{\bbR^d} {e^{i\bfk\cdot\bfx}\ptx\ud\bfx} = 
    i\bfk\cdot \intr{
         \bfvrf e^{i\bfk\cdot\bfx}\ptx
    \ud\bfx},  \label{eq:expanding_expectation}
\end{equation}
from which we can deduce:
\begin{align}
    &\int_{\bbR^d} {e^{i\bfk\cdot\bfx}\partial_r\ptx\ud\bfx} = 
    \intr{
         \bfvrf \nabla_{\bfx}\lb{e^{i\bfk\cdot\bfx}}\rb\ptx
    \ud\bfx}, \label{eq:expanding_expectation_1} \\
    &\int_{\bbR^d} {e^{i\bfk\cdot\bfx}\partial_r\ptx\ud\bfx} = 
    -\intr{
        e^{i\bfk\cdot\bfx}\nabla_{\bfx}\cdot\lsb\bfvrf\ptx\rsb
    \ud\bfx}, \label{eq:part_integrate}\\
    &\int_{\bbR^d} {
        e^{i\bfk\cdot\bfx}\lsb
        \partial_r\ptx
         + \nabla_{\bfx}\cdot\lsb\bfvrf\ptx\rsb
        \rsb
     }
     = 0,\label{eq:tranport_equation_fourier}
\end{align}
where $\nabla_{\bfx}\cdot\lsb\bfv p_{r,g}\rsb = \sum_{i=1}^d\frac{\partial}{\partial\bfx_i}\lsb v_i p_{r,g}\rsb$ denotes the divergence operator.
\Cref{eq:part_integrate} follows from integration by parts. 
By the uniqueness of the Fourier transform, \cref{eq:tranport_equation_fourier} implies that $\ptx$ satisfies the transport equation:
\begin{equation}
    \partial_r\ptx
         + \nabla_{\bfx}\cdot\lsb\bfvrf\ptx\rsb = 0.
    \label{eq:transport_equation_vr}
\end{equation}
Based on the theory of continuous normalizing flows (CNFs)~\cite{chen2018neural} and Flow Matching~\cite{lipman2022flow}, the probability flow associated with \cref{eq:transport_equation_vr} is governed by:
\begin{equation}
    \frac{\partial \bfx_{r,g}}{\partial r} = \bfvrf.
    \label{eq:vel_r_ode}
\end{equation}

\paragraph{Derivation of $\bfv(\bfx_{r,g},g)$}
Following the same derivation \wrt the generation time $g$, we obtain the corresponding transport equation for the $g$-axis:
\begin{equation}
    \partial_g\ptx
         + \nabla_{\bfx}\cdot\lsb\bfvgf\ptx\rsb = 0.
    \label{eq:transport_equation_vg}
\end{equation}
The associated probability flow is therefore governed by
\begin{equation}
    \frac{\partial \bfx_{r,g}}{\partial g} = \bfvgf, \quad \text{where} \quad \bfvgf \coloneqq \bbE\lsb\partial_g\bfx_{r,g} \condxrg\rsb.
    \label{eq:vel_g_ode}
\end{equation}

Combining \cref{eq:vel_r_ode,eq:vel_g_ode} yields the total differential:
\begin{equation}
    \ud\bfx_{r,g} = \bfvrf\ud r + \bfvgf \ud g,
    \label{eq:exact_equation_detail}
\end{equation}
which matches the PF-ODE in \cref{eq:disi_pfode_detail}. 
Moreover, $\bfvrf$ and $\bfvgf$ in \cref{eq:vel_r_detail,eq:vel_g_ode} exactly match \cref{eq:vr_expectation,eq:vg_expectation}.
This completes the proof.
\end{proof}

\section{Connection between DiSI and Stochastic Interpolants}
\label{app:theoretical_discussions}
Fundamentally, DiSI serves as a generalized extension of the standard SI framework~\cite{albergo2023stochastic}.
As discussed in \cref{sec:framework}, DiSI differs from standard SIs by decoupling the single time variable into regression time $r$ and generation time $g$.
This decoupling enables independent control over the transition between $\bfx_0$ and $\bfx_1$ and the generative process governed by noise injection of $\bfz$.
Once an explicit functional relationship between $r$ and $g$ is specified, DiSI reduces to a conventional one-dimensional SI process.
The Elliptical and Linear trajectories exemplify this reduction.
Below, we analyze the connection between DiSI and SIs under these trajectories and discuss the Regression Trajectory as a special case in detail.

\paragraph{DiSI with Elliptical Trajectory.}
Applying the relation between $r$ and $g$ in \cref{eq:elliptical_traj}, the DiSI process in \cref{eq:disi} can be reparameterized via a single time variable $t$:
\begin{equation}
    \begin{aligned}
    \bfx_t &= \underbrace{\rqftwo\lb
        \frac{\cos{(\ovalr)}}{\sqrt{1+\rho}}-
        \frac{\sin{(\ovalr)}}{\sqrt{1-\rho}}
    \rb\cos(\ovalg)}_{\alpha_t}\bfx_0 \\ &+  \underbrace{\rqftwo\lb
        \frac{\cos{(\ovalr)}}{\sqrt{1+\rho}}+
        \frac{\sin{(\ovalr)}}{\sqrt{1-\rho}}
    \rb\cos(\ovalg)}_{\beta_t}\bfx_1 + 
    \underbrace{\sin(\ovalg)}_{\gamma_t}\bfz,
    \end{aligned}
    \label{eq:disi_elliptical_traj}
\end{equation}
which exactly takes the standard SI form in \cref{eq:stochastic_interpolant}.
It is straightforward to verify that the coefficients $\alpha_t$, $\beta_t$, and $\gamma_t$ satisfy the boundary conditions $\alpha_{-\hpi} = \beta_{\hpi} = 1$, $\alpha_{\hpi}=\beta_{-\hpi} = \gamma_{-\hpi}=\gamma_{\hpi} = 0$ and $\forall t\in[-\hpi,\hpi], \alpha_t^2+\beta_t^2+\gamma_t^2 > 0$. 
Furthermore, $\alpha_t$, $\beta_t$, and $\gamma_t$ are all differentiable functions \wrt $t$ over $[-\hpi,\hpi]$.
Consequently, \cref{eq:disi_elliptical_traj} satisfies all SI conditions except for the time domain. 
This discrepancy is inessential, since we can linearly map the time interval to the standard $[0,1]$ range by simply replacing $t$ with $\pi \tilde t - \hpi$ in \cref{eq:disi_elliptical_traj}.

\paragraph{DiSI with Linear Trajectory.}
Substituting \cref{eq:linear_traj} into \cref{eq:disi} yields: %
\begin{equation}
    \begin{aligned}
    \bfx_t &= \underbrace{\rqftwo\lb
        \frac{\cos{(\linr)}}{\sqrt{1+\rho}}-
        \frac{\sin{(\linr)}}{\sqrt{1-\rho}}
    \rb\cos(\ling)}_{\alpha_t}\bfx_0 \\ & +  \underbrace{\rqftwo\lb
        \frac{\cos{(\linr)}}{\sqrt{1+\rho}}+
        \frac{\sin{(\linr)}}{\sqrt{1-\rho}}
    \rb\cos(\ling)}_{\beta_t}\bfx_1 + 
    \underbrace{\sin(\ling)}_{\gamma_t}\bfz,
    \end{aligned}
    \label{eq:disi_linear_traj}
\end{equation}
where $\alpha_t$, $\beta_t$, and $\gamma_t$ are differentiable in $[0,1]$ and satisfy the standard SI conditions $\alpha_{0} = \beta_{1} = 1$, $\alpha_{1}=\beta_{0} = \gamma_{0} = 0$, and $\alpha_t^2+\beta_t^2+\gamma_t^2 > 0$. 
The key difference is that $\gamma_1 = \sin(\delta) \neq 0$ for $\delta > 0$, keeping the terminal state stochastic.
Thus, the Linear path preserves the standard SI structure except relaxing the endpoint constraint $\gamma_1 = 0$.
We view this as a mild extension of the standard SI rather than a departure from it, as justified by its strong empirical performance. %

\paragraph{DiSI with Regression Trajectory.}
For the Regression Trajectory, setting $g=0$ in \cref{eq:disi_coeffcients} and substituting into \cref{eq:disi} yields a process governed solely by $r$:
\begin{equation}
    \bfx_r = \underbrace{\ar}_{\alpha_r} \bfx_0 + \underbrace{\br}_{\beta_r} \bfx_1,
    \label{eq:disi_regression_traj}
\end{equation}
where $\alpha_r$ and $\beta_r$ retain their forms from \cref{eq:disi_coeffcients}, but $\gamma_r$ strictly vanishes ($\gamma_r = 0$).
These coefficients are differentiable in $[-\hpi,\hpi]$ and satisfy all standard SI conditions.
Any mismatch in the time interval can again be removed by a linear reparameterization.
Most importantly, the condition $\forall r\in[-\hpi,\hpi], \gamma_r = 0$ makes the process entirely noiseless.
In this sense, the Regression Trajectory is theoretically aligned with the concept of latent-free SIs, as formulated in \cite{albergo2023stochastic}.

\input{algos/training}
\input{algos/inference}

\section{Algorithms}
\label{app:algorithms}
This section presents pseudo-code for training (\cref{sec:training}) and inference (\cref{sec:inference}).

\subsection{Training Algorithm}
\label{app:training_algorithm}
\cref{alg:train} summarizes the training procedure of DiSI.
As described in \cref{sec:training}, training takes as input a clean image $\bfx_0$, its degraded counterpart $\bfx_1$, the denoising network $\net$, and the adaptive weighting MLP $w_{\phi}$.
In each iteration, we first draw a pair of time variables $(r, g)$ using the chosen Time Sampler (Generalist or Specialist). 
We then draw noise $\bfz \sim \mathcal{N}(0,\sigmad^2\bfI)$ and construct the noisy intermediate state $\bfx(r,g)$ according to the DiSI process in \cref{eq:disi}.
At the same time, following EDM2~\cite{karras2024analyzing}, the adaptive loss weight $w$ is produced by an MLP $w_{\phi}$ conditioned on $(r, g)$, with $\phi$ as its parameters. 
The network $\net$ then takes the normalized noisy state $\frac{\bfx(r,g)}{\sigmad}$, the degraded image $\bfx_1$, and the joint time embeddings of $(r, g)$ as input to predict the clean image $\hat\bfx_0$ (see \cref{eq:net_prediction}).
Finally, the model parameters $\theta$ and the weighting parameters $\phi$ are jointly updated using the objective in \cref{eq:disi_loss_func}.
This procedure is repeated until convergence.

\subsection{Inference Algorithm}
\label{app:infer_algorithm}
\paragraph{Inference Algorithm for Generation Trajectory.}
\cref{alg:inference} summarizes DiSI inference for the Elliptical and Linear trajectories using the hybrid sampler in \cref{eq:disi_ddim_sampler}.
The time variable $t$ is discretized into $N$ steps along the chosen trajectory, from $\hpi$ to $-\hpi$ for the Elliptical path and from $1$ to $0$ for the Linear path.
The initial state $\bfx_{r_N,g_N}$ is then constructed from the degraded image $\bfx_1$ and sampled noise $\bfz$ according to the corresponding trajectory definition.
At each denoising step $i$, the network $\net$ predicts the clean image $\hat\bfx_0$ based on the current state $\bfx_{r_i,g_i}$, the degraded image $\bfx_1$, and the joint time embedding $(r_i, g_i)$. 
For the Elliptical trajectory, the first step ($i=N$) uses a fully stochastic update ($\eta=1$) to avoid the singularity at $g=0$, as mentioned in \cref{sec:inference}.
All remaining steps, as well as the entire Linear trajectory, use the standard hybrid sampler, where $\eta$ controls the injected stochasticity via the noise coefficient $\kappa$.
The process iteratively refines the image until the final restored output $\bfx_{r_0, g_0}$ is obtained.

\input{algos/regression_inference}
\paragraph{Inference Algorithm for Regression Trajectory.}
\Cref{alg:regresion_inference} describes the inference procedure for the regression path ($\delta=0$ in both the Elliptical and Linear cases).
Unlike the general sampler, the generation time is explicitly fixed at $g=0$, while the regression time $r$ is discretized from $\varphi$ to $-\varphi$.
Accordingly, the initial state $\bfx_{r_N,g_N}$ is set to the degraded image $\bfx_1$ without noise injection.
At each denoising step, the network predicts the clean image $\hat\bfx_0$. 
Since $g$ is fixed at $0$, the update rule simplifies to the deterministic form derived in \cref{eq:regressive_sampler}.
This yields a multi-step regression process that progressively refines $\bfx_1$ toward $\bfx_0$.
When $N=1$, the sampler reduces to standard one-step regression.

\section{Implementation Details}
\label{app:implemtation}
To ensure transparency and reproducibility, this section details our method configuration (\cref{app:implementation_comparison}), baseline reproduction (\cref{app:baseline_reproduction}), evaluation metrics (\cref{app:metric_details}), and ablation setups (\cref{app:implementation_ablation}).

\subsection{Implementation Details of Our Method}
\label{app:implementation_comparison}

\input{tables/implementation_details}
Optimal configurations for our main results (\cref{tab:deraining,tab:deblurring,tab:lowlight,tab:inpainting}) are in \cref{tab:training_details}.

\paragraph{Correlation Coefficient.}
We compute $\rho$ as the average correlation coefficient across all degraded-clean pairs in the training set, and apply it during testing.

\paragraph{Model Complexity.}
The computational complexity and parameter count remain consistent across all tasks due to the same architecture.
We report a computational cost of 128.67 GMACs (Giga-Multiply-ACcumulate operations) and 32.49M parameters, measured via \texttt{ptflops} with a $1 \times 3 \times 256 \times 256$ input.

\paragraph{Training Protocol.}
For a fair comparison, following IR-SDE~\cite{luo2023image}, we train all models for 500k steps with a global batch size of 16 (4 per GPU). 
To handle varying image resolutions, we retain the IR-SDE cropping strategy, but leverage the inherent efficiency of our model to enlarge the training patch size from $128 \times 128$ to $256 \times 256$. 
Finally, we use the Elliptical Specialist Time Sampler om \cref{alg:elliptical_sampler} for training. 
Since our inference variants, DiSI-R and DiSI-G, are both based on the Elliptical path (with the former being a special case), this strategy ensures the model is optimized for the corresponding sampling path.

\paragraph{Network Configuration.}
The network architecture is fixed across all tasks, adopting the optimal configuration from our ablation studies in \cref{sec:ablation_studies}. %
Specifically, we omit positional embeddings and post-modulation scales ($\xi_{\cdot}^{\cdot}$), retaining only pre-modulation shifts ($\varepsilon_{\cdot}^{\cdot}$) and scales ($\zeta_{\cdot}^{\cdot}$). 
We also use adaptive loss weighting and Conv projections ($1\times 1$ Conv + $3\times 3$ DWConv) for JLA and FFN.

\paragraph{Optimization.}
We train the model with AdamW at a learning rate of $10^{-4}$. 
We further apply EMA with a decay rate of 0.9999 to stabilize training.

\paragraph{Inference Configuration.}
We employ two inference variants, DiSI-R and DiSI-G, sharing identical model weights.
(1) DiSI-R maximizes efficiency and fidelity. It uses a single-step Regression Trajectory ($\delta=0, N=1$), functioning as a deterministic end-to-end regression model.
(2) DiSI-G targets high perceptual quality with moderate efficiency. 
We use the Elliptical path with $N=10$ since it introduces generative capability without the severe fidelity degradation of the Linear path (see \cref{tab:time_sampler}).
To resolve DiSI-G's initial singularity, we apply a stochastic booting step ($\eta=1$, $\epsilon=10^{-3}$), with $\eta=0$ for subsequent steps.

\subsection{Baseline Reproduction Details}
\label{app:baseline_reproduction}
\paragraph{Image Deraining.}
We categorize the metrics in \cref{tab:deraining} into two types:
(1) \textbf{Performance Metrics} (\eg, PSNR, SSIM): 
Most baseline scores are adopted directly from GOUB~\cite{yue2024image}, except for FoD~\cite{luo2025forward}, which is sourced from its original paper.
(2) \textbf{Efficiency Metrics} (Params, Lat.): 
To ensure a consistent benchmark, we re-evaluate these across all methods using their official codebases.
A notable exception for latency is the JAX-based MAXIM~\cite{tu2022maxim}, which we optimize with \texttt{jax.compile} to accurately reflect its full efficiency.

\paragraph{Image Deblurring.} 
For the deblurring comparison results in \cref{tab:deblurring}, we adopt performance metrics for most baselines directly from IR-SDE~\cite{luo2023image}. 
We re-evaluate efficiency metrics for all methods to ensure consistency, including the \texttt{jax.compile} optimization for MAXIM~\cite{tu2022maxim}.
Notably, for DeepDeblur~\cite{nah2017deep}, all the metrics reported derive from a modern \texttt{pytorch} reimplementation, since this version yields slight performance gains over the original \texttt{lua} codebase.

\paragraph{Low-Light Enhancement.}
Most performance metrics in \cref{tab:lowlight} are sourced from DA-CLIP~\cite{luo2024controlling}. 
However, we exclude DA-CLIP, as its all-in-one paradigm is incompatible with our task-specific setup. 
For other baselines, we report FoD~\cite{luo2025forward} metrics from its original paper and re-train GOUB~\cite{yue2024image}. 
Efficiency metrics are re-evaluated as above, retaining the \texttt{jax.compile} optimization for MAXIM~\cite{tu2022maxim}.

\paragraph{Image Inpainting.} 
Metrics for PromptIR~\cite{potlapalli2306promptir}, DDRM~\cite{kawar2022denoising}, and IR-SDE~\cite{luo2023image} are adopted from the IR-SDE paper. 
Moreover, we re-train GOUB~\cite{yue2024image} and FoD~\cite{luo2025forward} for inpainting to report the remaining results in \cref{tab:inpainting}. 
As in previous tasks, we evaluate efficiency across all methods using their official implementations.

\subsection{Implementation Details for Evaluation Metrics}
\label{app:metric_details}
We adhere to the evaluation protocol of IR-SDE~\cite{luo2023image} for all metric calculations.
\begin{itemize}
    \item Distortion Metrics (PSNR, SSIM): We rescale and clamp pixel values to $[0, 255]$. Computations are performed on RGB channels, with the exception of image deraining, which utilizes the Y channel (YCbCr space). %
    \item Perception Metrics: We compute LPIPS~\cite{zhang2018unreasonable} using its official implementation with the AlexNet~\cite{krizhevsky2012imagenet} backbone, scaling inputs to $[-1, 1]$. FID~\cite{heusel2017gans} is calculated via \texttt{pytorch-fid} with InceptionV3~\cite{szegedy2016rethinking} backbone, comparing the restored images against the ground truth of the test set.
    \item Efficiency Metrics: We compute parameter counts using the \texttt{ptflops} library. For latency, we ensure precise GPU timing via \texttt{torch.cuda.Event} and \texttt{torch.cuda.synchronize}, reporting the average across the test set.
\end{itemize}

\subsection{Implementation Details for Ablation Studies}
\label{app:implementation_ablation}
This section outlines the setup of ablation studies in \cref{sec:ablation_studies,app:addition_ablation}.

\subsubsection{Ablation on Optimization Settings.}
\label{app:ablation_optimization}
Our deraining ablation baseline adopts the default configurations in \cref{tab:training_details} under DiSI-R sampling, except we disable adaptive loss weighting and use \textit{Linear} FFN projections.
Using this baseline, we investigate the impact of the Exponential Moving Average (EMA) decay and the learning rate (LR) schedule (\cref{tab:optimization}). 
Specifically, the LR schedule follows IR-SDE~\cite{luo2023image}, halving the LR every 200k steps.
All other settings remain identical. 

\subsubsection{Ablation on Framework.}
\label{app:ablation_on_framework}
Building upon the preceding optimization ablation, we strictly fix the EMA decay to 0.9999 and disable the LR schedule. 
Under this established baseline, we benchmark DiSI against representative degraded-to-clean interpolants: Standard SI~\cite{albergo2024stochastic}, IR-SDE~\cite{luo2023image}, and DDBM~\cite{zhou2024denoising}. 
For fairness, all methods share the identical DULiT backbone and training protocol, differing solely in interpolants and corresponding samplers, with results in \cref{tab:interpolant}.

\subsubsection{Ablation on Regression Overhead.}
\label{app:ablation_regression_overhead}
To isolate the extra architectural cost of DiSI-R compared to pure regression networks, we adopt the preceding optimization baseline with EMA decay fixed at 0.9999 and a constant LR of 1e-4 and remove the Conditional Branch and Time Embedding/Modulation modules in \cref{tab:regression_analysis}.
The {w/o T\&C} variant is trained without noise perturbation to learn the deterministic LQ-to-GT mapping, mirroring an end-to-end regression network.

\subsubsection{Ablation on Network Settings.}
\label{app:ablation_network_componets}
Orthogonal to the progressive structural ablations below, we evaluate the prediction objective using the fully optimized baseline from \cref{tab:training_details}, predicting coupled velocities $(\bfv_r, \bfv_g)$ instead of the target $\bfx_0$ in \cref{subtab:pred}.
For subsequent component ablations, we initialize a baseline via an intermediate optimization setup (EMA decay 0.999, retaining the LR schedule).

\paragraph{Loss Weighting.} Under the aforementioned specific baseline setup, we first ablate the dynamic adaptive loss weighting, with results summarized in \cref{subtab:logvar}.

\paragraph{Long-Range Residual Connection.}
Enabling loss weighting, we proceed to ablate the long-range residual connection.
We replace the standard additive connection (see \cref{fig:network}) with a dynamic variant, scaling the noisy branch's residual feature via a time-dependent weight linearly predicted from joint time embeddings in \cref{subtab:longgate}.
Additionally, we evaluate removing this connection entirely in \cref{subtab:long_range_residual}.

\paragraph{Time Modulation.}
Retaining the dynamic residual, we introduce single-branch modulation in \cref{subtab:stime} by removing time modulation from the degraded image branch, as the degraded input remains constant across time steps.

Adopting the single-branch design, we integrate a level-shared adaptation of PixArt-$\alpha$ modulation~\cite{chen2024pixartalpha} in \cref{subtab:smod}.
While PixArt-$\alpha$ relies on a single global time projection with block-specific learnable offsets, DULiT's varying channel dimensions prohibit this global approach.
Consequently, we propose a level-shared approach: deploying one shared time projection per resolution level, where blocks within each level share base coefficients but maintain unique learnable offsets.

Adopting the level-shared PixArt-$\alpha$ modulation, we systematically prune its parameters to reduce redundancy in \cref{subtab:multiplier}. 
The full mechanism (see \cref{fig:network}b) comprises pre-norm scales $\zeta_{\cdot}^{\cdot}$, pre-norm shifts $\varepsilon_{\cdot}^{\cdot}$, and post-module scales $\xi_{\cdot}^{\cdot}$, collectively denoted as $(\zeta_{\cdot}^{\cdot}, \varepsilon_{\cdot}^{\cdot}, \xi_{\cdot}^{\cdot})$. 
We sequentially ablate the post-module scales $\xi_{\cdot}^{\cdot}$ (yielding  the $(\zeta_{\cdot}^{\cdot}, \varepsilon_{\cdot}^{\cdot})$ setting) and the pre-norm shifts $\varepsilon_{\cdot}^{\cdot}$ (retaining only $\zeta_{\cdot}^{\cdot}$).

\paragraph{JLA Deisgn.}
Adopting the $(\zeta_{\cdot}^{\cdot}, \varepsilon_{\cdot}^{\cdot})$ modulation, we first ablate the JLA module in \cref{subtab:attn_layer}. 
We compare \textit{Linear} and \textit{Conv} projections, defining the latter as a pointwise $1\times1$ followed by a depthwise $3\times3$ convolution. 
Additionally, we evaluate replacing the linear attention in DULiT's middle block with standard attention (\ie, \textit{Vanilla}) implemented by flash-attention~\cite{dao2022flashattention}.

\paragraph{FFN Design.}
Parallel to the JLA experiments, we independently ablate the FFN structure on the same $(\zeta_{\cdot}^{\cdot}, \varepsilon_{\cdot}^{\cdot})$ baseline (\cref{subtab:ffn_layer}). 
Beyond comparing \textit{Linear} and \textit{Conv} projections, we investigate \textit{ConvGate}, a convolutional adaptation of GeGLU~\cite{shazeer2020glu}. 
While standard GeGLU splits linearly projected inputs channel-wise to multiply a GeLU-activated half with the other, our \textit{ConvGate} replaces all linear projections with the aforementioned \textit{Conv} layers.

\paragraph{Positional Embedding.}
Fixing both JLA and FFN to use \textit{Conv} projections while retaining linear attention for JLA and the standard non-GeGLU structure for FFN, we evaluate applying RoPE~\cite{su2024roformer} to the JLA mechanism in \cref{subtab:rope}.

\subsubsection{Ablation on Time Samplers.}
\label{app:ablation_on_time_samplers}
To evaluate time samplers, we adopt a baseline incorporating all preceding architectural modification, specifically configuring \textit{Conv} JLA and \textit{Linear} FFN projections without positional embeddings.
Using this fixed setup, we vary the training time samplers in \cref{tab:time_sampler,tab:additonal_time_sampler}.

\subsubsection{Ablation on Trajectories.}
\label{app:ablation_on_trajectories}
\vpath %
Evaluating trajectory continuity requires no retraining.
We directly reuse the model trained via the \textit{lognorm2} generalist time sampler in \cref{tab:time_sampler}.
During inference, we deploy a parameterized V-path family (see \cref{fig:v_path}), defined as $r=\varphi t$ and $g=\delta(1-|t|^p)$ for $t\in[-1,1]$.
We vary $p \in \{1, 1.5, 2.5\}$ to control inflection point continuity ($C^0$ to $C^2$), alongside testing a Quadratic Bezier curve in \cref{tab:traj_analysis}.

\subsubsection{Ablation on Inference Hyperparameters.}
\label{app:ablation_on_inference_hyper_parameters}
Decoupled from training, ablating inference hyperparameters ($\delta$, NFE, $\eta$) requires no retraining. 
Hence, we evaluate these parameters on the fully optimized baseline in \cref{tab:training_details} via the default Elliptical path.
For the Linear path, we retain this optimal configuration but train a separate model via the \textit{Linear Specialist Time Sampler} (\cref{alg:linear_sampler}) for parallel inference ablations in \cref{tab:linear_hyperparameter}.

\section{Datsets Information}
\label{app:dataset_information}
\input{tables/datasets}
We evaluate DiSI on four IR benchmarks, which are detailed in \cref{tab:datasets}. 
For inpainting, we follow IR-SDE~\cite{luo2023image} by partitioning CelebA-HQ~\cite{karras2018progressive} into 29,900 training and 100 test images.
Corruptions are introduced using 100 thin masks from RePaint~\cite{lugmayr2022repaint}, dynamically sampled during training but fixed for testing.

\section{Additional Experimental Results}
\label{app:addition_results}

\subsection{Additional Ablation Studies}
\label{app:addition_ablation}
We put other ablation studies, which are less important compared to ablation experiments in \cref{tab:network_modules,tab:time_sampler,tab:hyperparameter}, in this section. 
All of the implementation details of ablation experiments in this section
are also discussed in \cref{app:implementation_ablation}.

\input{tables/concat_additional_modules_optimization}

\paragraph{Analysis of Optimization Settings.}
Ablating the EMA decay and LR schedule (\cref{tab:optimization}) reveals that higher EMA decay enhances training stability, yielding optimal performance at 0.9999.
Conversely, adopting IR-SDE's~\cite{luo2023image} LR schedule noticeably hampers performance compared to a constant learning rate.
Consequently, we adopt the $(0.9999, \text{w/o schedule})$ configuration in \cref{tab:optimization}. %

\paragraph{Analysis of More Network Settings.}
While our main ablation studies (\cref{tab:network_modules}) validate the core DULiT design, we also evaluate exploratory variants detailed in \cref{app:implementation_ablation}. 
These complex designs fail to improve our baseline (\cref{tab:additional_network_modules}):
\begin{itemize}
    \item Dynamic Long-Range Residuals (\cref{subtab:longgate}): 
    Introducing this mechanism causes a slight performance drop, indicating that simple additive residuals are sufficient for stable model performance. 
    \item Positional Embedding (\cref{subtab:rope}): 
    Applying positional embeddings, such as RoPE~\cite{su2024roformer}, yields no gains, aligning with recent findings in Sana~\cite{xie2025sana}. 
    \item Single-branch Time Modulation (\cref{subtab:stime}): 
    This configuration degrades most metrics. We hypothesize that this performance degradation stems from breaking the architectural symmetry between the dual branches.
    \item PixArt-$\alpha$ Style Modulation (\cref{subtab:smod}): 
    While reducing parameters, adopting the modulation scheme from PixArt-$\alpha$~\cite{chen2024pixartalpha} severely worsens FID.
\end{itemize}
Consequently, we exclude these modules to preserve optimal restoration quality.

\input{tables/regression_analysis}
\paragraph{Analysis of Regression Overhead.}
In the pure regression mode (DiSI-R), the model functions as a one-step predictor but retains components crucial for generation: Time Modulation and the Conditional Branch. 
To evaluate their architectural overhead, we strip these modules to establish a pure regression network in \cref{tab:regression_analysis}. 
While retaining these components introduces a modest cost ($\sim$13M parameters, 11ms latency), removing them yields negligible PSNR gains at the expense of noticeably degraded generative performance. 
This confirms that the overhead is an acceptable trade-off to preserve optimal generative quality. %

\input{tables/additional_time_sampler}
\paragraph{Analysis of Additional Time Samplers.}
\Cref{tab:additonal_time_sampler} extends our analysis to Lognorm variants~\cite{esser2024scaling}, showing that the optimal generalist sampler is highly trajectory-dependent. 
Specifically, the Uniform sampler excels in regression, whereas Lognorm2 and Lognorm4 prove superior for Elliptical and Linear stochastic paths, respectively. 
Crucially, shifting the mean of $r$ away from zero (\eg, Lognorm3) yields sub-optimal results, confirming that a centered prior for $r$ is essential for training stability, while the optimal distribution for $g$ varies by trajectory.

\input{tables/linear_sampler}

\paragraph{Analysis of Inference Hyperparameters for Linear Trajectory.}
Ablating the Linear path (\cref{tab:linear_hyperparameter}) reveals:
(1) 
Like the Elliptical path, generative capacity requires $\text{NFE} > 2$, with $\eta=0$ remaining optimal.
(2) %
While enabling a regression-generation transition, the Linear path exhibits sharper fidelity drops under increased stochasticity than its Elliptical counterpart. 
(3) %
The regression baseline ($\delta=0$) is inferior, likely due to the asymmetric sampling of $r$ around zero in the Linear Specialist Time Sampler (\cref{alg:linear_sampler}).
Such asymmetry creates a less favorable condition for the model to learn a stable, deterministic regression mapping compared to the symmetric Elliptical case.
These findings justify employing the Elliptical path for DiSI-G to maximize the fidelity-perception envelope.

\subsection{Additional Visual Experimental Results}
\label{app:addition_visual}
We provide qualitative comparisons across four IR tasks: full-image results are in \cref{fig:additional_deraining,fig:additional_deblurring,fig:additional_lowlight,fig:additional_inpainting}, with corresponding zoomed-in details in \cref{fig:additional_deraining_details,fig:additional_deblurring_details,fig:additional_lowlight_details,fig:additional_inpainting_details}.
\aderaining
\adeblurring
\alowlight
\ainpainting
\clearpage
\adderaining
\addebluring
\addlowlight
\addinpainting

\clearpage


%% file: algos/training.tex
\begin{algorithm}[t]
    \small
    \SetAlgoLined
    \DontPrintSemicolon
    \SetNoFillComment
    \KwIn{clean image $\bfx_0$, degraded image $\bfx_1$, network $\netred$, MLP $w_{\textcolor{red}{\phi}}$} 
    \Repeat{$\mathrm{converged}$}{          %
        $r, g \gets $ TimeSampler() 
        \textcolor{gray}{\text{// draw time variables}} \;
        $\bfz \sim \matN(0,\sigmad^2\bfI)$ 
        \textcolor{gray}{\text{// sample the noise from Gaussian Distribution}} \;
        $\bfx(r,g) = \lambda_g(\alpha_r\bfx_0 + \beta_r\bfx_1) + \gamma_g\bfz,$
        \textcolor{gray}{\text{// get noisy states $\bfxrg$ using \cref{eq:disi}}} \;
        $w \gets w_{\textcolor{red}{\phi}}(r, g)$ 
        \textcolor{gray}{\text{// calculate the loss weight}} \;
        $\hat\bfx_0 \gets \sigmad \netred(\frac{\bfx(r,g)}{\sigmad}, \bfx_1, r, g)$
        \textcolor{gray}{\text{// predict the clean image}} \;
        Take gradient descent step on 
            $\nabla_{\textcolor{red}{\theta,\phi}}\lsb e^{w} {\lL\hat\bfx_0 - \bfx_0 \rL}_2^2 - w\rsb$
        \textcolor{gray}{\text{// \cref{eq:disi_loss_func}}} \;
    }
    \caption{\small{\name Training Procedure.}
        \label{alg:train}
    }
\end{algorithm}

%% file: algos/inference.tex
\begin{algorithm}[t]
    \small
    \SetAlgoLined
    \SetAlgoNoEnd
    \DontPrintSemicolon
    \SetNoFillComment
    \SetKwIF{If}{ElseIf}{Else}{if}{}{else if}{else}{}
    \SetKwFor{ForEach}{foreach}{}{}
    \KwIn{
        network $\net$, 
        degraded image $\bfx_1$, 
        sampling step $N$, 
        trajectory type $\mathsf{Type}$, 
        hyperparameters $\eta, \sigma_d$
    } 
    \KwOut{
        restored image $\bfx_{r_0, g_0}$  %
    }
    \BlankLine
    \If{$\mathsf{Type}$ == \textnormal{"Elliptical"}}{
        $\{t_i\}_{i=0}^N \gets$ TimeDiscretization($N$, $\hpi \to -\hpi$) 
        \textcolor{gray}{\text{// discretize $t$ from $\hpi$ to $-\hpi$}} \;
    }
    \ElseIf{$\mathsf{Type}$ == \textnormal{"Linear"}}{
        $\{t_i\}_{i=0}^N \gets$ TimeDiscretization($N$, $1 \to 0$)
        \textcolor{gray}{\text{// discretize $t$ from $1$ to $0$}}\;
    }
    $r_N, g_N \gets $ TrajEquation($t_N$) 
    \textcolor{gray}{\text{// compute initial $(r, g)$ via \cref{eq:elliptical_traj} or \cref{eq:linear_traj}}}\;
    $\bfz \sgaussian$ 
    \textcolor{gray}{\text{// sample Gaussian noise for initialization}}\;
    $\bfx_{r_N,g_N} \gets \cos{g_N}\beta_{r_N}\bfx_1 + \sin{g_N}\bfz$
    \textcolor{gray}{\text{// initialize the starting state via \cref{eq:disi}}}\;
    \ForEach{$i$ {\textnormal{\bf in}} $\{N, N-1, \cdots, 1\}$}
    { 
        $r_i, g_i \gets$ TrajEquation($t_i$)   
        \textcolor{gray}{\text{// get current $(r, g)$ via \cref{eq:elliptical_traj} or \cref{eq:linear_traj}}}\;
        $r_{i-1}, g_{i-1} \gets$ TrajEquation($t_{i-1}$) 
        \textcolor{gray}{\text{// get next $(r, g)$ via \cref{eq:elliptical_traj} or \cref{eq:linear_traj}}}\;
        $\hat\bfx_0\gets \sigmad \net\lb \bfx_{r_i,g_i}/\sigmad, \bfx_1, r_i, g_i\rb$
        \textcolor{gray}{\text{// predict clean image $\hat\bfx_0$ (\cref{eq:net_prediction})}}\;
        $\bfz\sgaussian$
        \textcolor{gray}{\text{// sample noise for the stochastic step}}\;
        \If{$i$ == $N$ \textnormal{and} $\mathsf{Type}$ == \textnormal{"Elliptical"}}{  
            \textcolor{gray}{\text{// force fully stochastic step ($\eta=1$) to resolve singularity}}\;
            $\kappa \gets \sin{g_{i-1}}-\sin{g_i}$ 
            \textcolor{gray}{\text{// compute noise coefficient $\kappa$}}\;
            $\bfx_{r_{i-1}, g_{i-1}} \gets \bfx_{r_{i}, g_{i}}  + \cos{g_{i-1}}(\alpha_{r_{i-1}} \hat\bfx_0 + \beta_{r_{i-1}}\bfx_1)
              - \cos{g_i} (\alpha_{r_i} \hat\bfx_0 + \beta_{r_i}\bfx_1) $ $+ 
            \kappa \bfz$    
            \textcolor{gray}{\text{// update state using \cref{eq:disi_ddim_sampler} with $\eta=1$}}\;
        }
        \Else{
            \textcolor{gray}{\text{// standard hybrid sampling step}}\;
            $k \gets \sin{g_{i-1}} / \sin{g_i}$ \;
            $\kappa \gets \mathbb{I}(\eta \neq 0) \cdot \frac{\eta(\sin{g_{i-1}}-k^{\sqrt{1-\eta^2}}\sin{g_i})}{1 - \sqrt{1-\eta^2}}$ 
            \textcolor{gray}{\text{// compute noise coefficient $\kappa$}}\;
            $\bfx_{r_{i-1}, g_{i-1}} \gets k^{\sqrt{1-\eta^2}}\bfx_{r_{i}, g_{i}} +\cos{g_{i-1}}(\alpha_{r_{i-1}} \hat\bfx_0 + \beta_{r_{i-1}}\bfx_1)
              - k^{\sqrt{1-\eta^2}}\cos{g_i} (\alpha_{r_i} \hat\bfx_0 + \beta_{r_i}\bfx_1) + 
            \kappa \bfz$    
            \textcolor{gray}{\text{// update state using \cref{eq:disi_ddim_sampler}}}\;
        }
    }
    \Return{$\bfx_{r_0, g_0}$} 
    \caption{\small{\name Inference Procedure via the Elliptical or Linear Path.}
        \label{alg:inference}
    }
\end{algorithm}

%% file: algos/regression_inference.tex
\begin{algorithm}[t]
    \small
    \SetAlgoLined
    \DontPrintSemicolon
    \SetNoFillComment
    \SetKwIF{If}{ElseIf}{Else}{if}{}{else if}{else}{}
    \SetKwFor{ForEach}{foreach}{}{}
    \KwIn{
        network $\net$, 
        degraded image $\bfx_1$, 
        sampling step $N$, 
        hyperparameter $\sigma_d$
    } 
    \KwOut{
        restored image $\bfx_{r_0, g_0}$
    }
    \BlankLine  
    $\{r_i\}_{i=0}^N \gets$ TimeDiscretization($N$, $\varphi \to -\varphi$) 
    \textcolor{gray}{\text{// discretize $r$ from $\varphi$ to $-\varphi$}} \;
    $\{g_i\}_{i=0}^N \gets 0$
    \textcolor{gray}{\text{// fix generation time $g=0$ for the pure regression path}} \;
    $\bfx_{r_N,g_N} \gets \bfx_1$
    \textcolor{gray}{\text{// initialize state directly with the degraded image $\bfx_1$}}\;
    \ForEach{$i$ {\textnormal{\bf in}} $\{N, N-1, \cdots, 1\}$}
    { 
        $\hat\bfx_0\gets \sigmad \net\lb \bfx_{r_i,g_i}/\sigmad, \bfx_1, r_i, g_i\rb$
        \textcolor{gray}{\text{// predict clean image $\hat\bfx_0$ (\cref{eq:net_prediction})}}\;
        $\bfx_{r_{i-1}, g_{i-1}} \gets \bfx_{r_i, g_i} + (\alpha_{r_{i-1}} - \alpha_{r_i})\hat\bfx_0 + (\beta_{r_{i-1}} - \beta_{r_i}) \bfx_1$     
        \textcolor{gray}{\text{// \cref{eq:regressive_sampler}}}\;
    }
    \Return{$\bfx_{r_0, g_0}$} 
    \caption{\small{\name Inference Procedure via the Regression Path.}
        \label{alg:regresion_inference}
    }
\end{algorithm}

%% file: tables/implementation_details.tex
\begin{table}[t]
    \centering
    \caption{
    Details of our best training and testing configurations.
    \label{tab:training_details}
    }
    \definecolor{graybg}{gray}{0.9}
    \small
    \begin{tabularx}{0.98\linewidth}{@{}lYYYY}
        \toprule
        \textbf{Dataset} & \textbf{Rain100H}& \textbf{GoPro}& \textbf{LOL}& \textbf{Celeba-HQ} \\
        \midrule
        \rowcolor{graybg} \multicolumn{5}{l}{\textit{\textbf{Dataset-Specific Settings}}} \\
        Training Epochs & 4,425 & 3,788 & 16,130 & 268 \\
        Correlation Coeff. $\rho$ & 0.7482 & 0.9678 & 0.9678 & 0.5777 \\
        
        \midrule
        \rowcolor{graybg} \multicolumn{5}{l}{\textit{\textbf{Common Training and Hardware Configurations}}} \\
        Model Size & \multicolumn{4}{c}{32.49 M Parameters / 128.67 GMACs} \\
        Hardware & \multicolumn{4}{c}{4$\times$RTX 4090 (Global Batch Size: 16)} \\
        Training Setup & \multicolumn{4}{c}{500k Steps, $256^2$ Resolution, TF32 Precision} \\
        Time Sampler & \multicolumn{4}{c}{Elliptical (Specialist), Time Emb: 128 (Sinusoidal)} \\
        Time Modulation & \multicolumn{4}{c}{
        using $(\zeta_{\cdot}^{\cdot},\varepsilon_{\cdot}^{\cdot})$, omitting $\xi_{\cdot}^{\cdot}$
        }\\
        
        \midrule
        \rowcolor{graybg} \multicolumn{5}{l}{\textit{\textbf{Network Architecture}}} \\
        Levels \& Depth & \multicolumn{4}{c}{Levels: 4, Depths: [2, 2, 2, 2]} \\
        Channel Widths & \multicolumn{4}{c}{[64, 128, 256, 512]} \\
        Attention & \multicolumn{4}{c}{Conv Proj. Heads: [4, 4, 4, 4], Dim: [16, 32, 64, 128]} \\
        FFN Structure & \multicolumn{4}{c}{Conv Proj., Widths: [128, 256, 512, 1024]} \\
        Components & \multicolumn{4}{c}{Adaptive Loss Weighting (\ding{52}), AdaGroupNorm (32 groups)} \\
        Positional Emb. & \multicolumn{4}{c}{\ding{55} (None)} \\

        \midrule
        \rowcolor{graybg} \multicolumn{5}{l}{\textit{\textbf{Optimizer (AdamW)}}} \\
        Hyperparameters & \multicolumn{4}{c}{Lr: 1e-4, $\beta$: (0.9, 0.999), $\epsilon$: 1e-8, W. Decay: 1e-2} \\
        EMA Decay & \multicolumn{4}{c}{0.9999} \\
        
        \midrule
        \rowcolor{graybg} \multicolumn{5}{l}{\textit{\textbf{Inference Strategy: DiSI-R}}} \\
        Config & \multicolumn{4}{c}{Regression Trajectory, Steps: 1, $\delta$: 0, No Booting} \\

        \midrule
        \rowcolor{graybg} \multicolumn{5}{l}{\textit{\textbf{Inference Strategy: DiSI-G}}} \\
        Common Config & \multicolumn{4}{c}{Elliptical Trajectory, Steps: 10, Booting: \ding{52} ($\epsilon$=1e-3)} \\
        $\delta$ (Peak Noise Level) & 0.02 & 0.1 & 0.05 & 0.1 \\
        
        \bottomrule
    \end{tabularx}
\end{table}

%% file: tables/datasets.tex
\begin{table}[t]
    \centering
    \caption{Summary of the public image restoration datasets used in our work.\label{tab:datasets}}
    \renewcommand{\arraystretch}{1.2}
    \begin{tabularx}{\linewidth}{llYYY}
        \toprule
        \textbf{Task} & \textbf{Dataset} & \textbf{Type} & \textbf{Train Set} & \textbf{Test Set} \\
        \midrule
        Image Deraining & Rain100H~\cite{yang2017deep} & Synthetic & 1,800 & 100 \\
        Image Deblurring & GoPro~\cite{nah2017deep} & Synthetic & 2,103 & 1,111 \\
        Low-Light Enhancement & LoL~\cite{chen2018retinex} & Real & 485 & 15 \\
        Image Inpainting & CelebaHQ~\cite{karras2018progressive} & Synthetic & \multicolumn{2}{c}{30,000} \\
        \bottomrule
    \end{tabularx}
\end{table}

%% file: tables/concat_additional_modules_optimization.tex
\ifthenelse{\boolean{showpred}}{
\newcommand{\addrope}{
    \multicolumn{5}{@{}l}{\refstepcounter{subtable}\textbf{(\thesubtable) RoPE~\cite{su2024roformer}}\label{subtab:rope}} \\
    \rowcolor{baselinecolor}
    \quad \ding{55} & \textbf{34.39} & \textbf{0.9396} & \textbf{0.054} & \textbf{20.43} \\
    \quad \ding{52} & 33.04 & 0.9310 & 0.057 & 24.10 \\
}}{\newcommand{\addrope}{}}

\begin{table}[t]
    \centering
    \small
    \setlength{\tabcolsep}{2pt} 

    \begin{minipage}[t]{0.5\linewidth} %
        \centering
        \captionof{table}{
            Impact of optimization. Configurations are formatted as (EMA decay rate, LR schedule usage). The selected configuration is marked in \colorbox{baselinecolor}{gray}.
            \label{tab:optimization}
        }
        \renewcommand{\arraystretch}{1.55} 
        \begin{tabularx}{\linewidth}{@{} c *{4}{Y} @{}}
            \toprule
            Config. & PSNR$\uparrow$ & SSIM$\uparrow$ & LPIPS$\downarrow$ & FID$\downarrow$ \\
            \midrule
            (0.95, \s\s\ding{55})    &34.28&0.9396&0.052&19.94\\
            (0.99,  \s\s\ding{55})   &33.96&0.9344&0.058&22.36\\
            (0.999, \s\ding{55})     &34.56&0.9402&0.050&19.89\\
            (0.999, \s\ding{52})     &34.38&0.9380&0.052&19.94\\
            \rowcolor{baselinecolor}
            (0.9999,  \ding{55})     &\best{34.69}&\best{0.9417}&\best{0.049}&\best{18.88}\\
            \bottomrule
        \end{tabularx}

    \end{minipage}
    \hfill
    \begin{minipage}[t]{0.45\linewidth}
        \centering
        \captionof{table}{
            \textbf{Additional DULiT Ablations.} 
            Selected settings in \colorbox{baselinecolor}{gray}.
            \label{tab:additional_network_modules}
        }
        \renewcommand{\arraystretch}{0.92} 
        \begin{tabularx}{\linewidth}{@{} l *{4}{Y} @{}}
            \toprule
            Config. & PSNR$\uparrow$ & SSIM$\uparrow$ & LPIPS$\downarrow$ & FID$\downarrow$ \\
            \midrule

            \multicolumn{5}{@{}l}{ \refstepcounter{subtable}\textbf{(\thesubtable) Dynamic long-range residual}\label{subtab:longgate}} \\
            \rowcolor{baselinecolor}
            \quad \ding{55} & \textbf{34.50} & \textbf{0.9405} & 0.053 & \textbf{19.27} \\
            \quad \ding{52} & 34.46 & 0.9401 & \textbf{0.052} & 19.60 \\

            \addrope

            \multicolumn{5}{@{}l}{ \refstepcounter{subtable}\textbf{(\thesubtable) Single-branch modulation}\label{subtab:stime}} \\
            \rowcolor{baselinecolor}
            \quad \ding{55} & \textbf{34.46} & \textbf{0.9401} & \textbf{0.052} & \textbf{19.60} \\
            \quad \ding{52} & 34.40 & \textbf{0.9401} & 0.053 & 19.80 \\

            \multicolumn{5}{@{}l}{ \refstepcounter{subtable}\textbf{(\thesubtable) PixArt-$\alpha$ style modulation}\label{subtab:smod}} \\
            \rowcolor{baselinecolor}
            \quad \ding{55} & 34.40 & \textbf{0.9401} & \textbf{0.053} & \textbf{19.80} \\
            \quad \ding{52} & \textbf{34.41} & 0.9399 & 0.054 & 20.15 \\
            \bottomrule
        \end{tabularx}
    \end{minipage}
\end{table}

%% file: tables/regression_analysis.tex
\begin{table}[t]
    \centering
    \caption{
        Architectural overhead and quantitative performance of DiSI-R and its stripped variants. 
        $T$: Time Embedding and Modulation, $C$: Conditional Branch.
    }
    \label{tab:regression_analysis}
    \renewcommand{\arraystretch}{1.2}

    \begin{tabular}{l cc cccc}
        \toprule
        \multirow{2}{*}{\textbf{Model}} & \multicolumn{2}{c}{\textbf{(a) Overhead}} & \multicolumn{4}{c}{\textbf{(b) Comparison}} \\
        \cmidrule(lr){2-3} \cmidrule(lr){4-7}
        & Params (M) & Latency (ms) & PSNR $\uparrow$ & SSIM $\uparrow$ & LPIPS $\downarrow$ & FID $\downarrow$ \\
        \midrule
        DiSI-R & 32.49 & 45.37 $\pm$ 0.26 & 34.93 & \textbf{0.946} & \textbf{0.054} & \textbf{18.88} \\
        w/o $T$ & 26.46 & 43.13 $\pm$ 0.09 & -- & -- & -- & -- \\
        w/o $T$\&$C$ & 19.42 & 33.94 $\pm$ 0.07 & \textbf{35.11} & \textbf{0.946} & 0.055 & 20.30 \\
        \bottomrule
    \end{tabular}
\end{table}

%% file: tables/additional_time_sampler.tex
\begin{table*}[t]
    \centering
    \caption{
    Ablation on generalist time samplers. 
    Models trained with these samplers are evaluated on three deterministic ($\eta=0$) trajectories:
    \textbf{Regression Trajectory} ($\delta=0$, 1 step), \textbf{Elliptical Trajectory} ($\delta=\pi/8$, 5 steps), and \textbf{Linear Trajectory} ($\delta=\pi/8$, 5 steps).
    Lognorm1 uses $\mathsf{lognorm}(0.0,1.0)$ for both $r,g$. 
    Lognorm2 uses $\mathsf{lognorm}(0.0,1.0)$ for $r$ and $\mathsf{lognorm}(-0.5,1.0)$ for $g$.
    Lognorm3 uses $\mathsf{lognorm}(-0.5,1.0)$ for both $r,g$.
    Lognorm4 uses $\mathsf{lognorm}(0.0,1.0)$ for $r$ and $\mathsf{lognorm}(-1.0,1.0)$ for $g$.
    \label{tab:additonal_time_sampler}
    }
    
    \resizebox{\textwidth}{!}{
    \begin{tabular}{c c cccc  cccc cccc}
    \toprule
    \multicolumn{2}{c}{\multirow{2}*{\shortstack{Time Samplers \\ of $(r, g)$}}} & \multicolumn{4}{c}{Regression Trajectory}&\multicolumn{4}{c}{Elliptical Trajectory} & \multicolumn{4}{c}{Linear Trajectory} \\
\cmidrule(lr){3-6} \cmidrule(lr){7-10} \cmidrule(lr){11-14} 
&& PSNR$\uparrow$ & SSIM$\uparrow$ & LPIPS$\downarrow$ & FID$\downarrow$ & PSNR$\uparrow$ & SSIM$\uparrow$ & LPIPS$\downarrow$ & FID$\downarrow$ & PSNR$\uparrow$ & SSIM$\uparrow$ & LPIPS$\downarrow$ & FID$\downarrow$ \\
    \midrule
        \multicolumn{2}{c}{Uniform} 
        &\best{34.16}&0.9375&\best{0.057}&\best{20.70}
        &33.46&0.9283&0.051&21.23
        &32.61&0.9083&0.058&26.55\\
        \multicolumn{2}{c}{lognorm1} %
        &33.54&\best{0.9378}&0.058&23.75
        &\best{33.75}&0.9320&0.049&19.28 
        &32.80&0.9122&0.055&26.21\\
        \multicolumn{2}{c}{lognorm2} %
        &32.73&0.9301&0.061&27.31
        &33.61&\best{0.9379}&\best{0.048}&\best{18.63}
        &\best{32.94}&0.9169&0.047&23.59\\
        \multicolumn{2}{c}{lognorm3} %
        &32.00&0.9217&0.076&34.30 
        &33.43&0.9325&0.049&18.78
        &32.87&0.9165&0.048&23.81\\
        \multicolumn{2}{c}{lognorm4} %
        &32.05&0.9161&0.092&40.57
        &33.34&0.9316&0.050&18.99
        &32.84&\best{0.9174}&\best{0.046}&\best{22.54}\\
    \bottomrule
    \end{tabular}
    }
    
\end{table*}

%% file: tables/linear_sampler.tex
\begin{table}[t]
    \centering
    \caption{
        \textbf{Effects of $\delta$, $\eta$, and NFE on Linear path.}
        Regression mode in \colorbox{gray!15}{gray}.
        \label{tab:linear_hyperparameter}
    }
    \begingroup
    \footnotesize 
    \renewcommand{\arraystretch}{0.95} 
    \setlength{\tabcolsep}{2pt} 

    \begin{tabularx}{0.95\linewidth}{@{} c c *{5}{YY} @{}}
        \toprule
        \multirow{2}{*}{\textbf{$\delta$}} & \multirow{2}{*}{\textbf{$\eta$}} & \multicolumn{2}{c}{\textbf{NFE = 1}} & \multicolumn{2}{c}{\textbf{NFE = 2}} & \multicolumn{2}{c}{\textbf{NFE = 5}} & \multicolumn{2}{c}{\textbf{NFE = 15}} & \multicolumn{2}{c}{\textbf{NFE = 50}} \\
        \cmidrule(lr){3-4} \cmidrule(lr){5-6} \cmidrule(lr){7-8} \cmidrule(lr){9-10} \cmidrule(l){11-12}
        & & PSNR$\uparrow$ & LPIPS$\downarrow$ & PSNR$\uparrow$ & LPIPS$\downarrow$ & PSNR$\uparrow$ & LPIPS$\downarrow$ & PSNR$\uparrow$ & LPIPS$\downarrow$ & PSNR$\uparrow$ & LPIPS$\downarrow$ \\
        \midrule

        \rowcolor{gray!15}
        0 &  N/A  & 33.77 & 0.058 & \textbf{33.78} & 0.058 & 33.84 & 0.056 & \textbf{33.78} & \textbf{0.055} & 33.72 & \textbf{0.055} \\
        \addlinespace %

        \multirow{4}{*}{$\cfrac{\pi}{8}$} 
        & 0.0 &  N/A  &  N/A  & 33.77 & 0.058 & 32.96 & 0.056 & 32.02 & \textbf{0.055} & 31.56 & \textbf{0.055} \\
        & 0.2 &  N/A  &  N/A  & 33.77 & 0.058 & 30.24 & 0.119 & 29.71 & 0.116 & 30.32 & 0.085 \\
        & 0.5 &  N/A  &  N/A  & 33.77 & 0.058 & 25.26 & 0.340 & 26.12 & 0.295 & 29.11 & 0.148 \\
        & 1.0 & 33.77 & 0.058 & 33.77 & 0.058 & 24.14 & 0.543 & 28.34 & 0.286 & 31.42 & 0.147 \\
        \midrule

        \multirow{4}{*}{$\cfrac{\pi}{4}$} 
        & 0.0 &  N/A  &  N/A  & \textbf{33.78} & 0.058 & 33.08 & 0.059 & 32.00 & 0.061 & 31.39 & 0.061 \\
        & 0.2 &  N/A &  N/A & \textbf{33.78} & 0.058 & 28.08 & 0.212 & 28.29 & 0.183 & 29.56 & 0.114 \\
        & 0.5 &  N/A  &  N/A  & \textbf{33.78} & 0.058 & 21.99 & 0.549 & 23.59 & 0.419 & 27.26 & 0.234 \\
        & 1.0 & 33.77 & 0.058 & \textbf{33.78} & 0.058 & 19.52 & 0.922 & 24.19 & 0.565 & 28.35 & 0.300 \\
        \midrule

        \multirow{4}{*}{$\cfrac{\pi}{2}$} 
        & 0.0 &  N/A  &  N/A  & \textbf{33.78} & 0.058 & 31.02 & 0.075 & 30.18 & 0.079 & 29.56 & 0.078 \\
        & 0.2 &  N/A  &  N/A  & \textbf{33.78} & 0.058 & 26.49 & 0.286 & 26.48 & 0.245 & 27.38 & 0.190 \\
        & 0.5 & N/A &  N/A  & \textbf{33.78} & 0.058 & 19.52 & 0.828 & 21.70 & 0.558 & 24.56 & 0.352 \\
        & 1.0 & 33.77 & 0.058 & \textbf{33.78} & 0.058 & 16.07 & 1.179 & 20.15 & 0.861 & 24.43 & 0.502 \\
        
        \bottomrule
    \end{tabularx}
    \endgroup
\end{table}